\documentclass[10pt,twocolumn,letterpaper]{article}

\usepackage{iccv}
\usepackage{times}
\usepackage{epsfig}
\usepackage{graphicx}
\usepackage{amsmath}
\usepackage{amssymb}
\usepackage{bm}
\usepackage{multirow}
\usepackage{color}
\usepackage{overpic}

\def\ie{\emph{i.e}.,~}
\def\eg{\emph{e.g}.,~}

\usepackage[breaklinks=true,bookmarks=false]{hyperref}

\iccvfinalcopy

\def\iccvPaperID{2101}

\ificcvfinal\fi

\begin{document}

\title{Pro-Cam SSfM: Projector-Camera System\\for Structure and Spectral Reflectance from Motion}

\author{Chunyu Li, Yusuke Monno, Hironori Hidaka, and Masatoshi Okutomi\\
Tokyo Institute of Technology, Tokyo, Japan}

\maketitle
\ificcvfinal\thispagestyle{empty}\fi

\begin{abstract}
In this paper, we propose a novel projector-camera system for practical and low-cost acquisition of a dense object 3D model with the spectral reflectance property. In our system, we use a standard RGB camera and leverage an off-the-shelf projector as active illumination for both the 3D reconstruction and the spectral reflectance estimation. We first reconstruct the 3D points while estimating the poses of the camera and the projector, which are alternately moved around the object, by combining multi-view structured light and structure-from-motion (SfM) techniques. We then exploit the projector for multispectral imaging and estimate the spectral reflectance of each 3D point based on a novel spectral reflectance estimation model considering the geometric relationship between the reconstructed 3D points and the estimated projector positions. Experimental results on several real objects demonstrate that our system can precisely acquire a dense 3D model with the full spectral reflectance property using off-the-shelf devices.
\end{abstract}

\section{Introduction}
\begin{figure}[!tbp]
  \centering
 \includegraphics[width=\hsize]{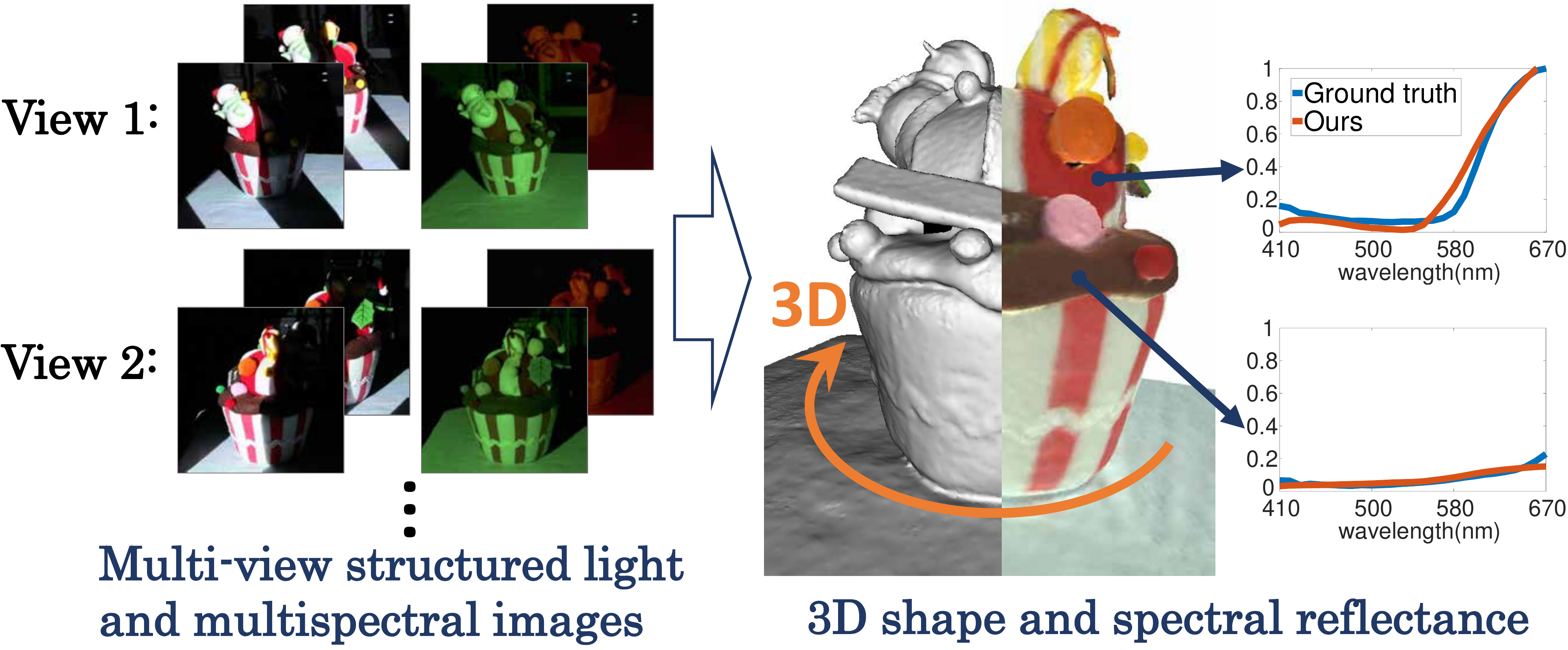}\\ \vspace{1mm}
  \caption{From multi-view structured light and multispectral images captured using an alternately moved projector and camera, our system can reconstruct a dense object 3D model having the spectral reflectance property for each 3D point.}
  \label{fig:introduction}
\end{figure}

The nature of an object is typically represented by two properties: geometric and photometric properties. The geometric property is determined by the 3D structure of the object, while the photometric property is determined by how the incident light is reflected at each 3D point of the object surface. Among various photometric parameters, spectral reflectance is one of the most fundamental physical quantities, which defines the amount of reflected light over that of the incident light at each wavelength. In this work, our aim is to acquire the spectral 3D information of an object using low-cost off-the-shelf devices (see Fig.~\ref{fig:introduction}).
Practical and low-cost acquisition of the spectral 3D information has many potential applications in fields such as cultural heritage~\cite{Chane,Kim3}, plant modeling~\cite{Behmann,Liang}, spectral rendering~\cite{Devlin}, and multimedia~\cite{Mansouri}.

3D reconstruction is a very active research area in computer vision. Structure from motion (SfM)~\cite{Schoenberger,Snavely}, multi-view stereo~\cite{Furukawa,Seitz}, and structured light~\cite{Furukawa2,Garrido,Geng,Li7,Salvi} are common approaches for the 3D shape acquisition. However, they usually focus on the geometric reconstruction. Although some recent methods combine the geometric and the photometric reconstruction~\cite{Kim4,Maurer,Melou}, they still focus on the estimation of RGB albedo, which is dependent on the camera RGB sensitivity and not an inherent property of the object unlike the spectral reflectance.

Multispectral imaging is another active research area. Various hardware-based systems~\cite{Baek,Cao,Cui,Han2,Monno,Park,Oh} and software-based methods~\cite{Aeschbacher,Arad,Fu,Jia4,Nguyen2,Shi} have been proposed for recovering scene's spectral reflectance. However, they usually assume a single-viewpoint input image and do not consider the geometric relationship between the object surface and the light source, only achieving scene and viewpoint-dependent spectral recovery, where shading or shadow is ``baked in" the recovered spectral reflectance.

Some systems have also been proposed for spectral 3D acquisition~\cite{Hirai,Ito,Kim,Kitahara,Ozawa,Zia}. However, they rely on a dedicated setup using a multispectral camera~\cite{Kim,Ozawa,Zia} or a multispectral light source~\cite{Hirai,Ito,Kitahara}, which makes the system impractical and expensive for most users.

In this paper, we propose a novel projector-camera system, named Pro-Cam SSfM, for structure and spectral reflectance from motion. In Pro-Cam SSfM, we use a standard RGB camera and leverage an off-the-shelf projector for two roles: structured light and multispectral imaging. For the data acquisition, structured light patterns (for geometric observations) and uniform color illuminations (for multispectral observations) are sequentially projected onto the object surface while alternately moving the camera and the projector positions around the object. Using the multi-view structured light data, we first reconstruct the 3D points while estimating the poses of all moved cameras and projectors. Using the multi-view multispectral data, we then estimate the spectral reflectance of each 3D point considering the geometric relationship between the reconstructed 3D points and the estimated projector positions.

\vspace{2mm}
\noindent
{\bf Technical contributions} of this work are listed as below.
\begin{enumerate}
  \vspace{-1mm}
  \item We propose an extended self-calibrating multi-view structured light method, where we include the moved projectors for feature correspondences and pose estimation to realize denser 3D reconstruction. The estimated projector positions are further exploited for the following spectral reflectance estimation.
  \vspace{-1mm}
  \item We propose a novel spectral reflectance estimation model by incorporating the geometric relationship between the reconstructed 3D points and the estimated projector positions into the cost optimization. Our model leads accurate estimation of the inherent spectral reflectance of each 3D point while eliminating the baked-in effect of the shading and the shadow.
  \vspace{-1mm}
  \item By integrating the above key techniques into one system, we propose Pro-Cam SSfM, a novel projector-camera system for practical and low-cost spectral 3D acquisition. We experimentally demonstrate that Pro-Cam SSfM can precisely reconstruct a dense object 3D model with the spectral reflectance property. To the best of our knowledge, Pro-Cam SSfM is the first spectral 3D acquisition system using off-the-shelf devices.
\end{enumerate}

\section{Related Work}

\noindent {\bf Structured light systems:}
Structured light is a well-adopted technique to accurately reconstruct the 3D points irrespective of surface textures by projecting structured light patterns~\cite{Aliaga,Geng,Salvi,Michael}. While structured light methods are generally based on a pre-calibrated projector-camera system, some multi-view structured light methods~\cite{Furukawa2,Garrido,Li7} have realized self-calibrating reconstruction of the 3D points. The key of these methods is that the structured light patterns projected by one fixed projector is captured from more than two camera viewpoints having viewing angle overlaps, so that feature matching and tracking can be made to connect all projectors and cameras. This setup can be realized by simultaneously using multiple projectors and cameras~\cite{Furukawa2,Garrido} or alternately moving a projector and a camera~\cite{Li7}.

In Pro-Cam SSfM, we extend the method~\cite{Li7} for realizing denser 3D reconstruction, as detailed in Section~\ref{ssec:3d}. We also exploit the estimated projector positions for modeling the spectral reflectance, while existing methods~\cite{Furukawa2,Garrido,Li7} only focus on the geometric reconstruction.

\vspace{2mm}
\noindent {\bf Multispectral imaging systems:}
Existing hardware-based~\cite{Baek,Cao,Cui,Han2,Monno,Oh,Park} or software-based~\cite{Aeschbacher,Arad,Fu,Jia4,Nguyen2,Shi} multispectral imaging systems commonly apply a single-viewpoint image-based spectral reflectance estimation method ignoring scene's or object's geometric information. This means that they only achieve scene-dependent spectral reflectance estimation, where the viewpoint and scene-dependent shading or shadow is baked in the estimated spectral reflectance. 

In Pro-Cam SSfM, we use an off-the-shelf RGB camera and projector for multispectral imaging. Although this setup is the same as~\cite{Han2}, we propose a novel spectral reflectance estimation model for recovering the object's inherent spectral reflectance considering the geometric information that can be obtained at the 3D reconstruction step.

\vspace{2mm}
\noindent {\bf Spectral 3D acquisition systems:}
Existing systems for spectral 3D acquisition are roughly classified into photometric stereo-based~\cite{Kitahara,Ozawa}, SfM and multi-view stereo-based~\cite{Ito,Zia}, and active lighting or scanner-based~\cite{Hirai,Kim} systems.
The photometric stereo-based systems~\cite{Kitahara,Ozawa} can acquire dense surface normals for every pixels of the single-view image. However, they have a main limitation that the light positions should be calibrated.
The SfM and multi-view stereo-based systems~\cite{Ito,Zia} enable self-calibrating 3D reconstruction using multi-view images. However, they only can provide a sparse point cloud especially for a texture-less object.
The active lighting or scanner-based systems~\cite{Hirai,Kim} can provide an accurate and dense 3D model based on the active sensing. However, they require burdensome calibration of the entire system.
Furthermore, all of the above-mentioned systems rely on a dedicated setup using a multispectral camera~\cite{Kim,Ozawa,Zia} or a multispectral light source~\cite{Hirai,Ito,Kitahara} for achieving multispectral imaging capability.
Those limitations make the existing system impractical and expensive for most users, narrowing the range of applications.

Pro-Cam SSfM overcomes those limitations because (i) it uses a low-cost off-the-shelf camera and projector, (ii) it does not require geometric calibration, and (iii) it generates a dense 3D model based on the structured light.

\section{Proposed Pro-Cam SSfM}
\label{sec:proposed}

Pro-Cam SSfM consists of three parts: data acquisition, self-calibrating 3D reconstruction, and spectral reflectance estimation. Each part is detailed below.

\subsection{Data acquisition}
\label{ssec:data}

Figure~\ref{fig:configuration} illustrates the data acquisition procedure of Pro-Cam SSfM. We use an off-the-shelf projector as active illumination and a standard RGB camera as the imaging device to capture the object illuminated by the projector. As shown in Fig.~\ref{fig:configuration}(a), the projector is used to project a sequence of structured light patterns and uniform color illuminations to acquire geometric and photometric observations. As the structured light patterns, we use the binary gray code~\cite{Geng,Salvi}. As the uniform color illuminations, we use the seven illuminations: red, green, blue, cyan, magenta, yellow, and white illuminations, which are generated using the binary combinations of the RGB primaries as (R,G,B) = (1,0,0), (0,1,0), (0,0,1), (0,1,1), (1,0,1), (1,1,0), (1,1,1), respectively. The sequence of active projections are effectively exploited in the 3D reconstruction and the spectral reflectance estimation.

To scan the whole object, we follow the data acquisition procedure of~\cite{Li7}. The data acquisition starts with initial projector and camera positions (\eg position~1 in~Fig.~\ref{fig:configuration}(b)). Then, the camera and the projector are alternately moved around the object (\eg motion~1, motion~2, and so on, in Fig.~\ref{fig:configuration}(b)). This acquisition procedure enables to connect the structured light codes (\ie feature points) between successive projectors and cameras. All connected feature points using all projector and camera positions are used as correspondences for the SfM pipeline, which enables self-calibrating reconstruction of the 3D points while estimating the poses of all moved projectors and cameras.

\subsection{Self-calibrating 3D reconstruction}
\label{ssec:3d}

Given the structured light encoded images, we first perform self-calibrating reconstruction of the 3D points while estimating the poses of all moved cameras and projectors. This is performed by extending~\cite{Li7} as below.

\subsubsection{Feature correspondence}

\begin{figure}[!tbp]
  \centering
   \includegraphics[width=\hsize]{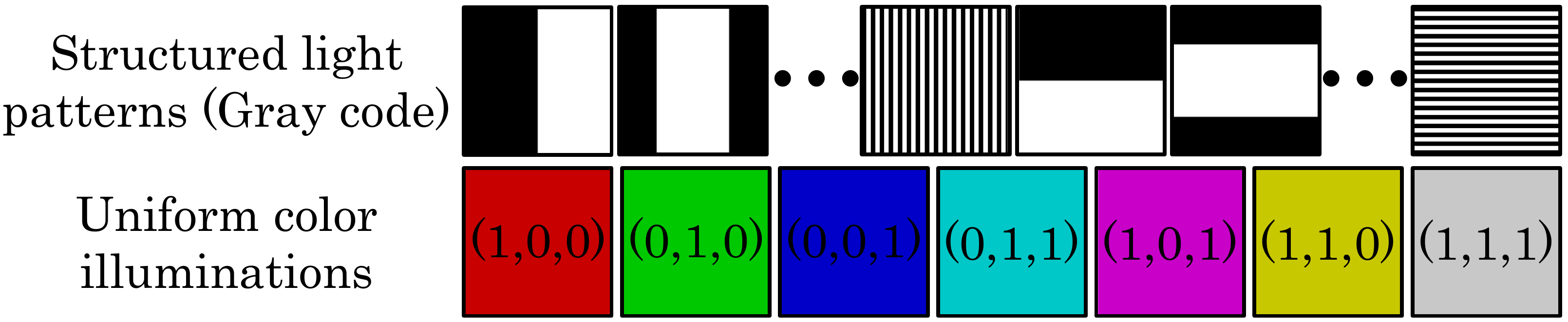}
   \vspace{3mm}   
   \small{(a) Projected illuminations}
   \includegraphics[width=0.7\hsize]{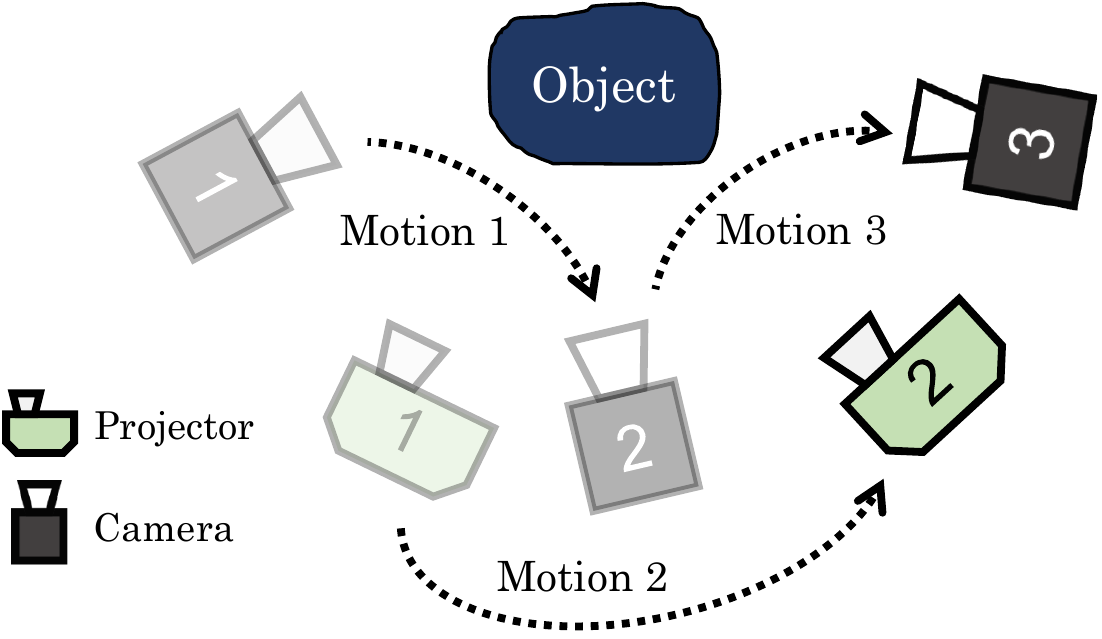}\\
   \small{(b) Data acquisition procedure}
   \vspace{2mm}
  \caption{Data acquisition procedure of Pro-Cam SSfM. (a) The projector projects a sequence of structured light patterns (gray code) and uniform color illuminations to acquire geometric and photometric observations. (b) The camera and the projector are alternately moved around the object, so that the structured light codes can be connected among all camera and projector positions.}
  \label{fig:configuration}
\end{figure}

By projecting gray code patterns as shown in Fig.~\ref{fig:configuration}(a), the projector can add “features” on object surfaces. Those features have different codes whose number is the same as the projector resolution. By decoding the code for each pixel and calculating the center position of the pixels having the same code, we can obtain “features” at sub-pixel accuracy positions in each image.

The feature correspondences for camera-projector pairs (\eg camera1-projector1 and camera2-projector1 pairs in Fig.~\ref{fig:configuration}(b)) and camera-camera pairs (\eg camera1-camera2 pair) sharing the same projector code are obvious. The features from different projectors can be connected using a common camera (\eg camera2 for projector1 and projector2), \ie the features from different projectors are regarded as identical, if their positions are close enough (less than 0.5 pixels in our experiments). Once they are connected, correspondences can be made for all combinations of cameras and projectors (\eg even correspondence for a projector-projector pair can be made). In contrast to the fact that the method~\cite{Li7} only uses the correspondences of all camera-camera pairs, we use all correspondences including projectors, which results in denser 3D points as shown in Fig.~\ref{fig:3dresult}.

\subsubsection{3D point and projector-camera pose estimation}

The set of all obtained correspondences is then fed into a standard SfM pipeline~\cite{Schoenberger,Snavely} to estimate the 3D points, the projector poses, and the camera poses. In the SfM pipeline, we modify the bundle adjustment formulation~\cite{Lourakis} so as to minimize the following weighted reprojection errors.

\begin{equation}
E = \sum_i \sum_k w_i \| \bm{x}_{k,i}-H_{i}(\bm{p}_k)\|^2,
\label{eq:reprojection}
\end{equation}
where $\bm{p}_k$ is the 3D coordinate of point $k$, $\bm{x}_{k,i}$ is the corresponding pixel coordinates in $i$-th viewpoint (camera or projector), and $H_{i}(\bm{p})$ is a function that projects the 3D point to $i$-th viewpoint (camera or projector) using intrinsic and extrinsic parameters for each projector and each camera. In Eq.~(\ref{eq:reprojection}), we set a larger weight to impose higher penalties for the reprojection errors of the projector as
\begin{equation}
w_i = \begin{cases}
		1,& \text{if viewpoint $i$ is a camera} \\
		w_p,& \text{if viewpoint $i$ is a projector}
		\end{cases},
\label{eq:weight}
\end{equation}
where $w_p>1$,
because it can be regarded that ``feature" positions of projectors have almost no errors. Through the bundle adjustment, 3D points and whole system parameters, including projector and camera positions and intrinsic parameters
for both projectors and cameras are estimated without any pre-calibration.

\subsection{Spectral reflectance estimation}
\label{ssec:spectral}

Given the estimated 3D points, projector positions, and camera poses, we next estimate the spectral reflectance of each 3D point. For this purpose, we use multispectral images captured under the uniform color illuminations. In what follows, we first introduce our proposed rendering model and then explain cost optimization to estimate the spectral reflectance using multi-view multispectral images. 

\subsubsection{Rendering model}
We here introduce our rendering model for each 3D point using a single projector-camera pair. Suppose the object surface is modeled by Lambertian reflectance and the camera response is linear, the camera's pixel intensity $y$ for $k$-th 3D point captured by $m$-th camera channel and $n$-th projected illumination is modeled as
\begin{equation}
y_{k,m,n}(\bm{x}_k)= s_k\int_{\Omega_\lambda}c_{m}(\lambda)l_n(\lambda)r(\bm{p}_k,\lambda)d\lambda,
\label{eq:rendering}
\end{equation}
where $\bm{x}_k$ is the projected pixel coordinate for $k$-th point, $r(\bm{p}_k,\lambda)$ is the spectral reflectance of $k$-th point, $l_n(\lambda)$ is the spectral power distribution of $n$-th projected illumination, $c_{m}(\lambda)$ is the camera spectral sensitivity of $m$-th channel, $s_k$ is the shading factor for $k$-th point, and $\Omega_\lambda$ is the target wavelength range. In practice, the continuous wavelength domain is discretized to $N_\lambda$ dimension (typically, sampled at every 10nm from 400nm to 700nm, \ie $N_\lambda$=31). Suppose the camera has three (\ie RGB) channels and $N_l$ illuminations are projected, the observed multispectral intensity vector for $k$-th point $\bm{y}_k \in \mathbb{R}^{3N_l}$ can be expressed as
\begin{equation}
\bm{y}_k = s_k \bm{C}^T \bm{L} \bm{r}_k,
\label{eq:matrixform}
\end{equation}
where $\bm{r}_k \in \mathbb{R}^{N_\lambda}$ represents the spectral reflectance, $\bm{L} = [\bm{L}_1;\cdots;\bm{L}_{N_l}] \in \mathbb{R}^{N_lN_\lambda \times N_\lambda}$ is the illumination matrix, where $\bm{L}_n \in \mathbb{R}^{N_\lambda \times N_\lambda}$ is the $n$-th diagonal illumination matrix, and $\bm{C}^T = blockdiag(\bm{C}_{rgb}^T,\cdots,\bm{C}_{rgb}^T) \in \mathbb{R}^{3N_l\times N_lN_\lambda}$ is the block diagonal matrix, where $\bm{C}_{rgb}^T \in \mathbb{R}^{3 \times N_\lambda}$ is the camera sensitivity matrix. In this work, we assume that the spectral power distributions of the projected illuminations and the camera sensitivity (\ie $\bm{C}^T$ and $\bm{L}$) are known or preliminarily estimated (\eg by~\cite{Jiang,Oh}). 

\subsubsection{Spectral reflectance model}
It is known that the spectral reflectance of natural objects is well represented by a small number of basis functions~\cite{Parkkinen}. Based on this observation, we adopt a widely used basis model~\cite{Han2,Park}, where the spectral reflectance is modeled as
\begin{equation}
\bm{r}_k=\bm{B} \bm{\alpha}_k,
 \label{eq:reflectance_basis}
\end{equation}
where $\bm{B} \in \mathbb{R}^{N_\lambda \times N_b}$ is the basis matrix, where $N_b$ is the number of basis functions, and $\bm{\alpha}_k \in \mathbb{R}^{N_b}$ is the coefficient vector. The basis model can reduce the number of parameters (since $N_b < N_\lambda$) for spectral reflectance estimation. Using the basis model, Eq.~(\ref{eq:matrixform}) is rewritten as
\begin{equation}
\bm{y}_k = s_k \bm{C}^T \bm{L} \bm{B} \bm{\alpha}_k.
\label{eq:matrixform2}
\end{equation}

\begin{figure}[!tbp]
  \centering
  \includegraphics[width=0.6\linewidth]{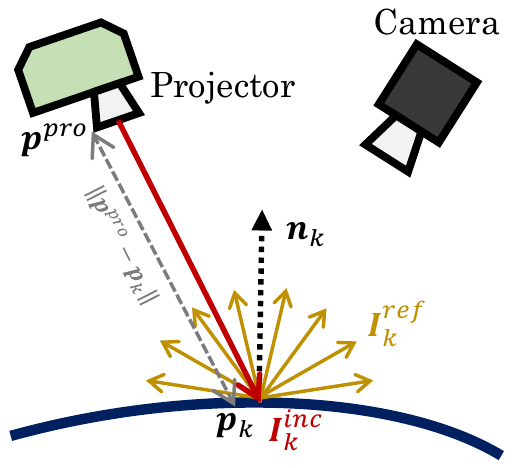}
  \caption{Geometric relationship between the projector $\bm{p}^{pro}$ and the 3D point $\bm{p}_k$ to calculate the shading factor. We assume that the point has Lambertian reflectance and the projected illumination follows the inverse-square law.}
  \label{fig:shading}
\end{figure}

\subsubsection{Shading model}
Different from common single-view image-based methods (\eg \cite{Han2,Hirai,Oh,Park}), we take the shading factor into account for spectral reflectance estimation, which results in more accurate model and estimation. Figure~\ref{fig:shading} illustrates the relationship between the projector position~$\bm{p}^{pro}$, $k$-th 3D point~$\bm{p}_k$, and the point normal~$\bm{n}_k$ (which can be calculated using~\cite{Hoppe}). Since the shading factor is wavelength independent and determined by the geometric relationship between the projector and the 3D point (under the Lambertian reflectance assumption), we rewrite the illumination power as $l(\lambda)=l$ in the following derivation. Then, the shading factor at $k$-th point is modeled as 
\begin{equation}
s_k= I_k^{ref}/l,
\label{eq:shading}
\end{equation}
where $I_k^{ref}$ is the irradiance of the reflected light incoming to the camera. In our model, the shading factor determines how much the amount of the projected illumination reaches the camera, irrespective of the spectral reflectance. If we assume Lambartian reflectance, $I_k^{ref}$ is independent of the camera position and expressed as
\begin{equation}
I^{ref}_k=I^{inc}_k\times\frac{\bm{p}^{pro}-\bm{p}_k}{\|\bm{p}^{pro}-\bm{p}_k\|}\cdot \bm{n}_k,
\label{eq:lambartian}
\end{equation}
where $I^{inc}_k$ is the irradiance of the incident light at $k$-th point and $(\bm{p}^{pro}-\bm{p}_k)/\|\bm{p}^{pro}-\bm{p}_k\|\cdot \bm{n}_k$ represents the inner product of the normalized lighting vector and the point normal (see Fig.~\ref{fig:shading}). Based on the near-by light model and the inverse-square law, $I^{inc}_k$ is inversely proportional to the square of the distance from the projector to the 3D point as
\begin{equation}
I^{inc}_k=\frac{l}{\|\bm{p}^{pro}-\bm{p}_k\|^2}.
\label{eq:inversesquare}
\end{equation}
If we assume that the ambient light is negligible and omit interreflection from the model, the shading factor is modeled from Eqs.~(\ref{eq:shading})--(\ref{eq:inversesquare}) as   
\begin{equation}
s_{k}=\frac{\bm{p}^{pro}-\bm{p}_k}{\|\bm{p}^{pro}-\bm{p}_k\|^3}\cdot \bm{n}_k.
 \label{eq:shading2}
\end{equation}
Based on this model, the shading factor can be calculated from the 3D point, the point normal, and the projector position that we have already obtained by the self-calibrating 3D reconstruction. The final rendering model is derived by substituting Eq.~(\ref{eq:shading2}) into Eq.~(\ref{eq:matrixform2}).

\subsubsection{Visibility calculation}
To estimate the spectral reflectance using multi-view images, we need to calculate the visibility of each 3D point. For this purpose, the object surface is reconstructed using Poisson surface reconstruction~\cite{Kazhdan,Kazhdan2}. Then, for each projector-camera pair, a set of 3D points that are visible from both the camera and the projector is calculated. By calculating the visibility, we discount the effects of cast shadows for spectral reflectance estimation.

\subsubsection{Cost optimization}
Using the rendering model of Eq.~(\ref{eq:matrixform2}), we solve an optimization problem to estimate the spectral reflectance of each 3D point from multi-view images obtained from all projector-camera pairs. The cost function is defined as
\begin{equation}
\arg\min_{\bm{\alpha}_k}~E_{ren}(\bm{\alpha}_k)+ \gamma E_{ssm}(\bm{\alpha}_k),
\label{eq:costfunction}
\end{equation}
where $E_{ren}$ is the rendering term and expressed as
\begin{equation}
E_{ren}(\bm{\alpha}_k)=\sum_{c \in \mathcal{V}(k)}\frac{\|{\bm{y}}_{k,c}^{obs}-\bm{y}_{k,c}(\bm{\alpha}_k)\|^2}{|\mathcal{V}(k)|},
 \label{eq:costren}
\end{equation}
where ${\bm{y}}_{k,c}^{obs} \in \mathbb{R}^{3N_l}$ is the observed multispectral intensity vector obtained from $c$-th projector-camera pair, ${\bm{y}}_{k,c}(\bm{\alpha}_k)$ is the estimated intensity vector based on the rendering model, and $\mathcal{V}(k)$ is the visible set for $k$-th point. This term evaluates the data fidelity between the observed and the rendered intensities. $E_{ssm}$ is a commonly used spectral smoothness term~\cite{Han2,Oh,Park}, which is defined by
\begin{equation}
E_{ssm}(\bm{\alpha}_k)= \bm{D}\bm{B}\bm{\alpha}_k,
 \label{eq:costssm}
\end{equation}
where $\bm{D} \in \mathbb{R}^{N_\lambda \times N_\lambda}$ is the operation matrix to calculate the second-order derivative~\cite{Park}. This term evaluates the smoothness of the estimated spectral reflectance. The balance of $E_{ren}$ and $E_{ssm}$ is determined by the parameter $\gamma$.

\begin{figure}[!t]
  \centering
  {\renewcommand{\arraystretch}{0.5}
  \begin{tabular}{@{\hskip 0pt}c@{\hskip 0pt}c@{\hskip 0pt}c@{\hskip 0pt}c@{\hskip 0pt}}
    \includegraphics[width=0.25\hsize]{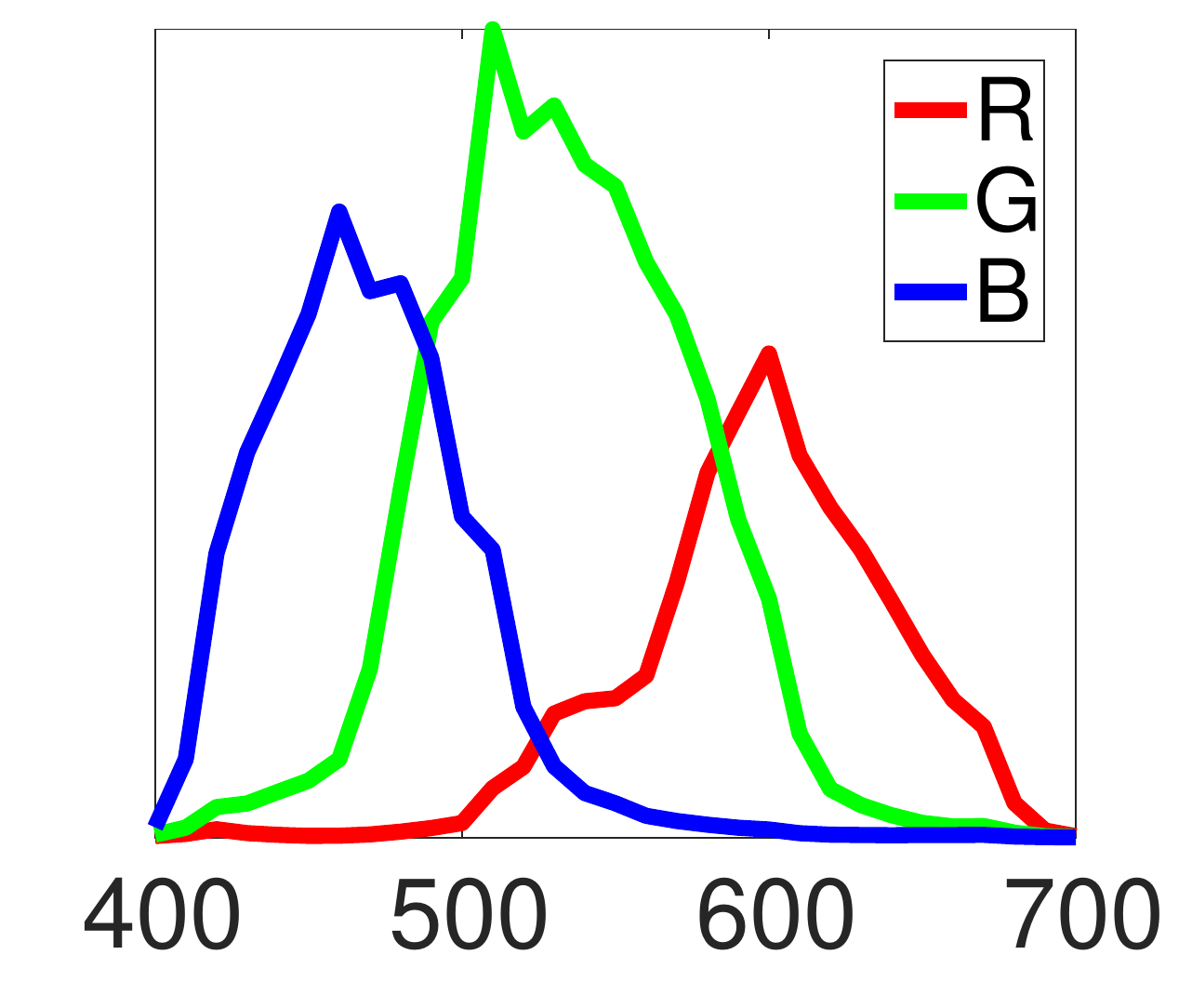}&
    \includegraphics[width=0.25\hsize]{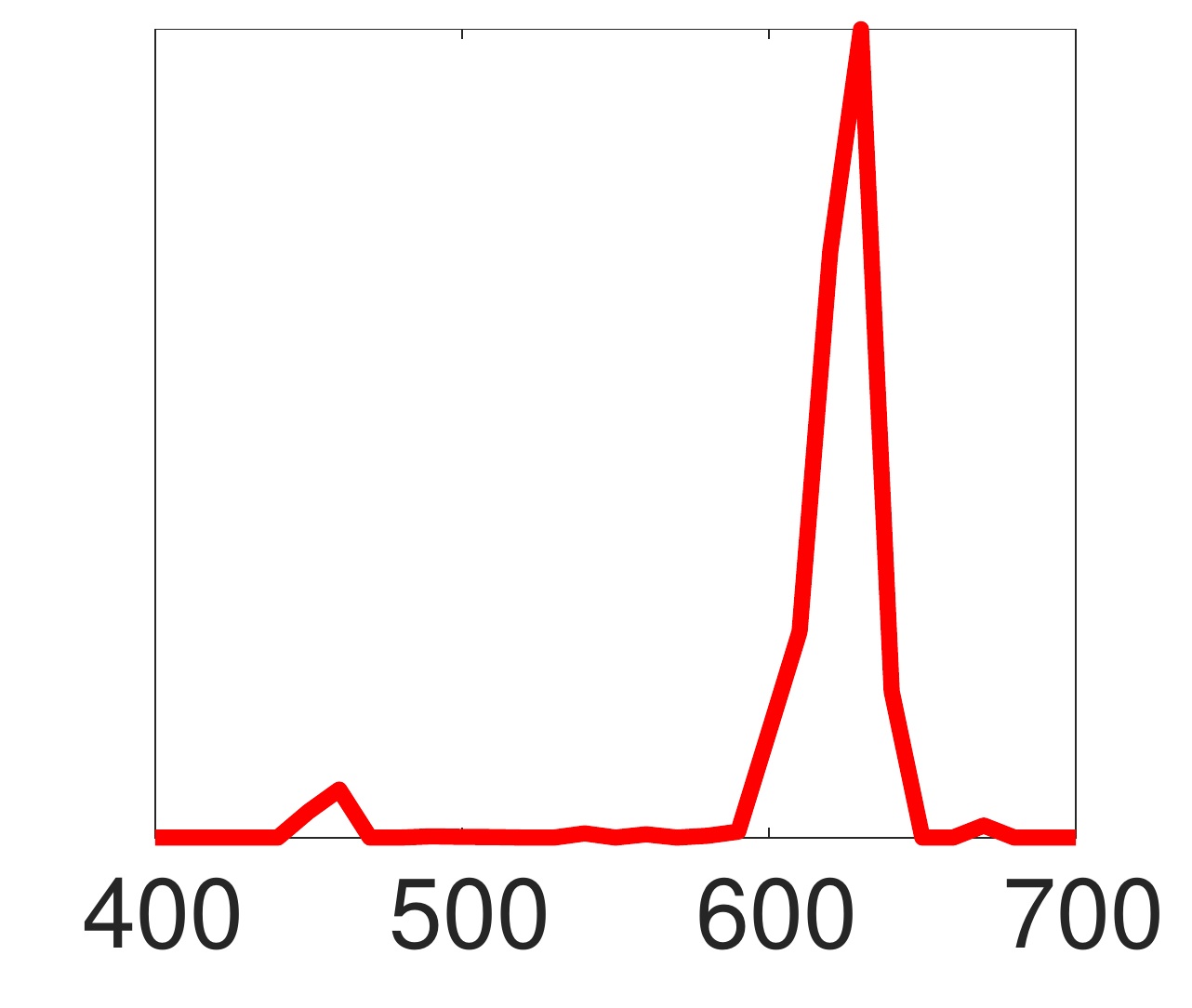}&
    \includegraphics[width=0.25\hsize]{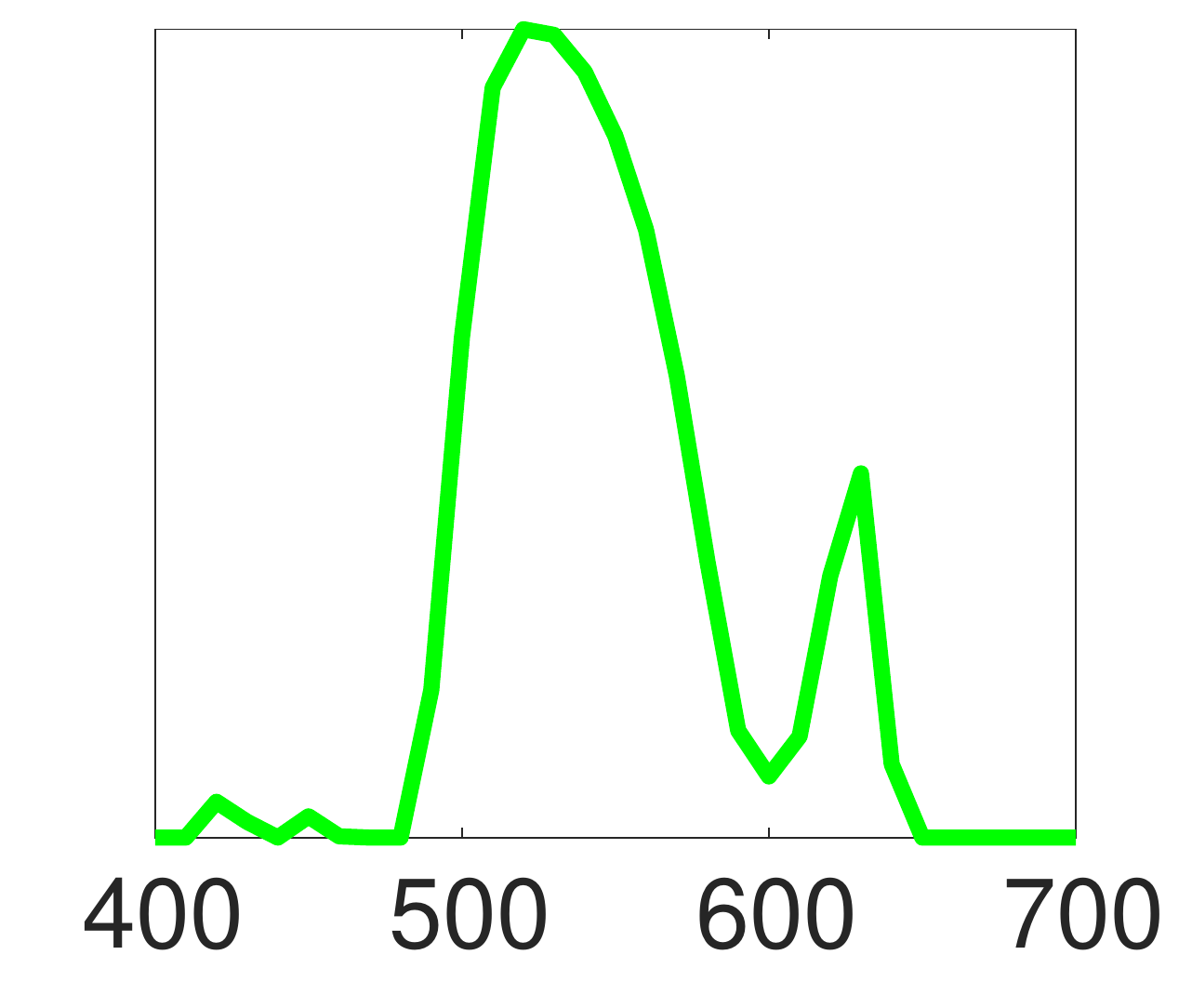}&
    \includegraphics[width=0.25\hsize]{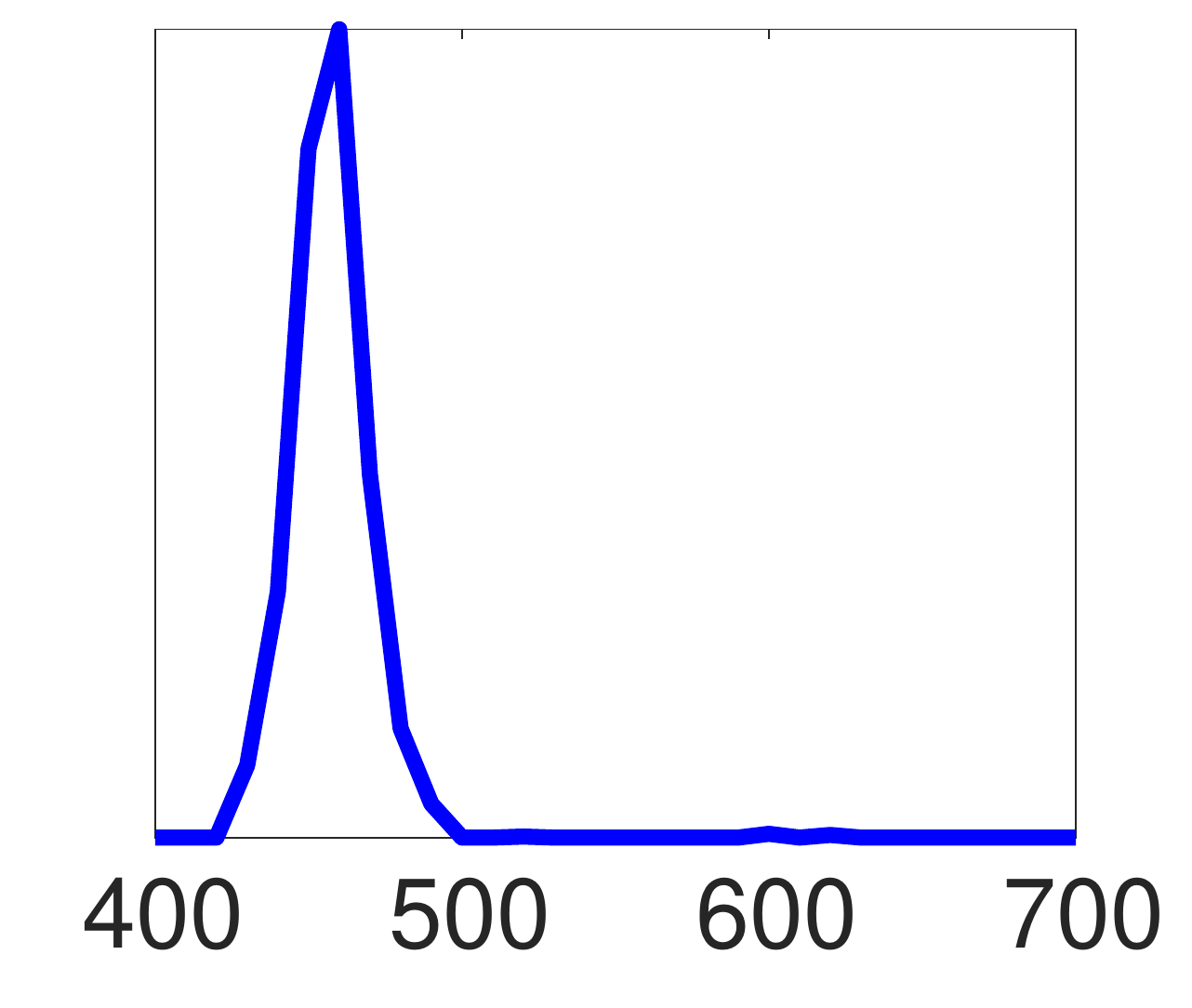}\\
    {\footnotesize Camera sensitivity}&{\footnotesize Red}&{\footnotesize Green}&{\footnotesize Blue} \vspace{2mm} \\ 
    \includegraphics[width=0.25\hsize]{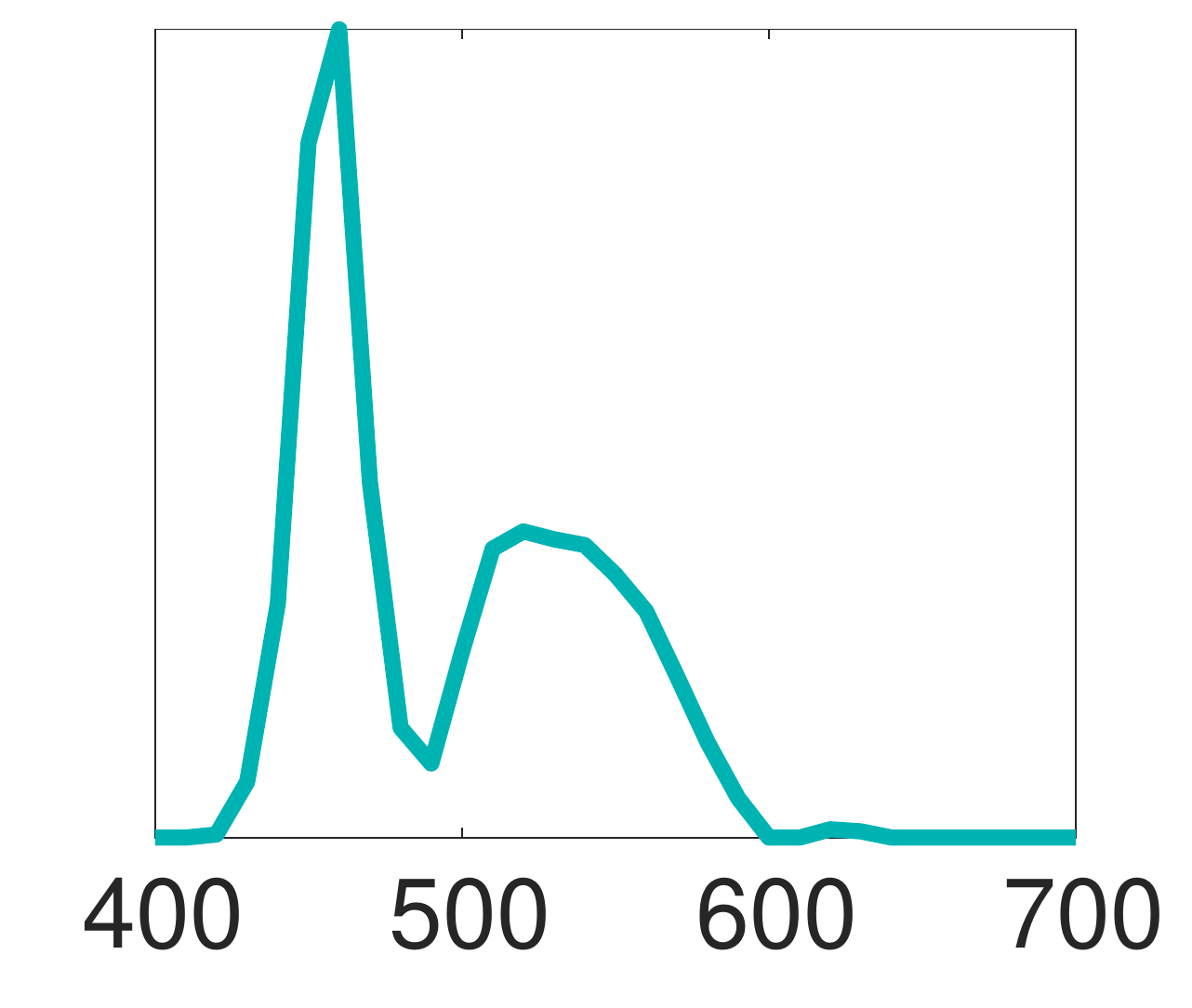}&
    \includegraphics[width=0.25\hsize]{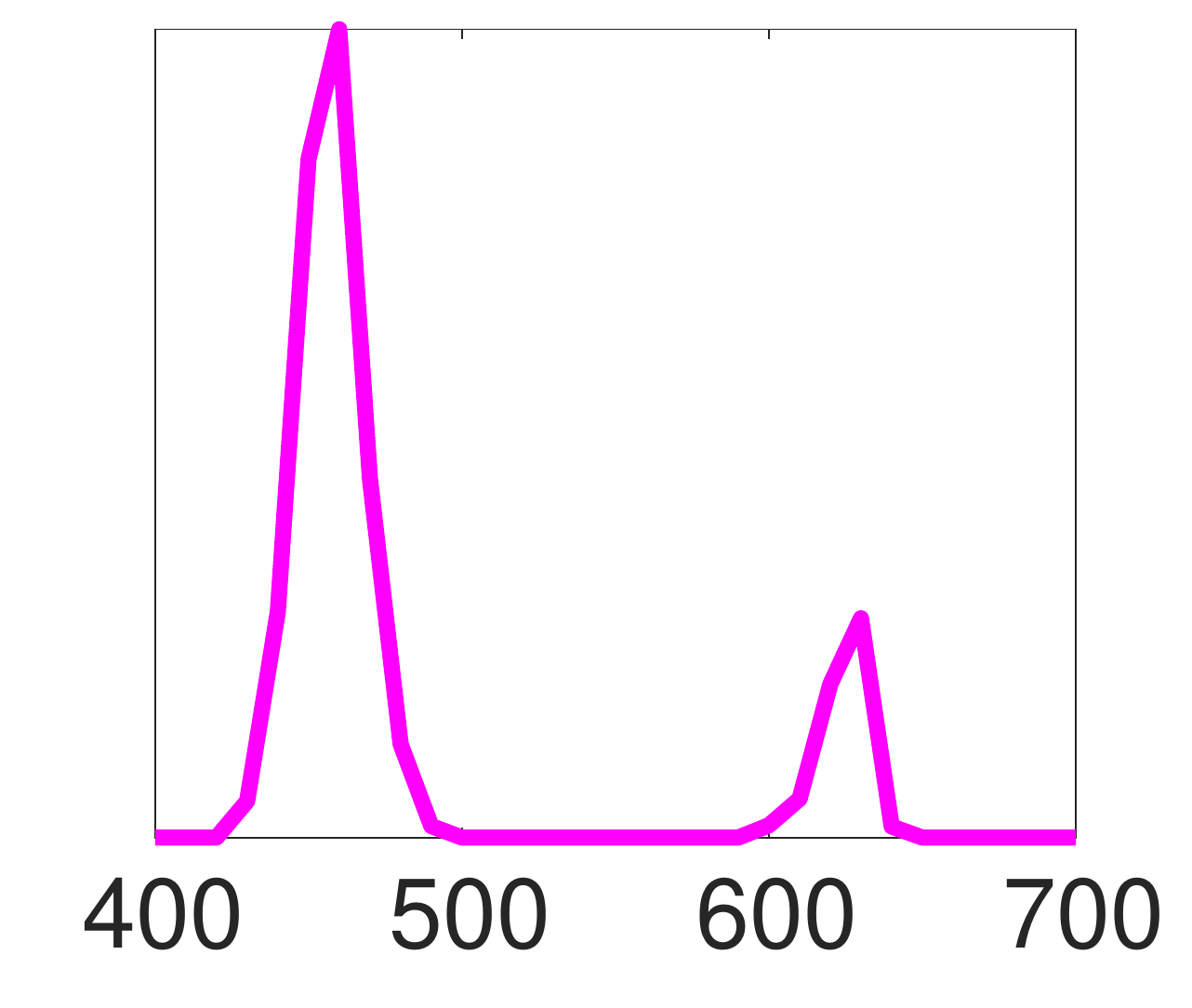}&
    \includegraphics[width=0.25\hsize]{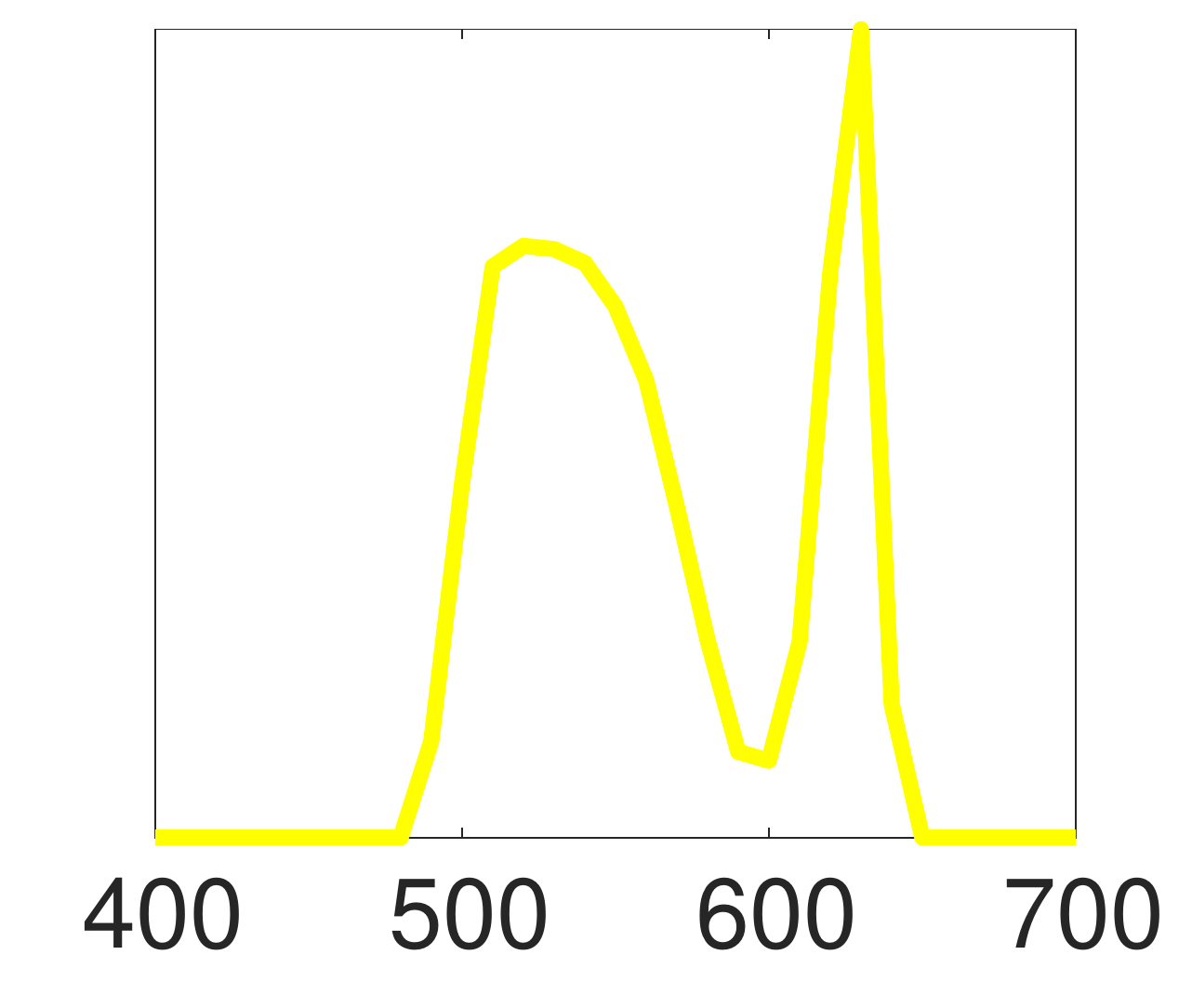}&
    \includegraphics[width=0.25\hsize]{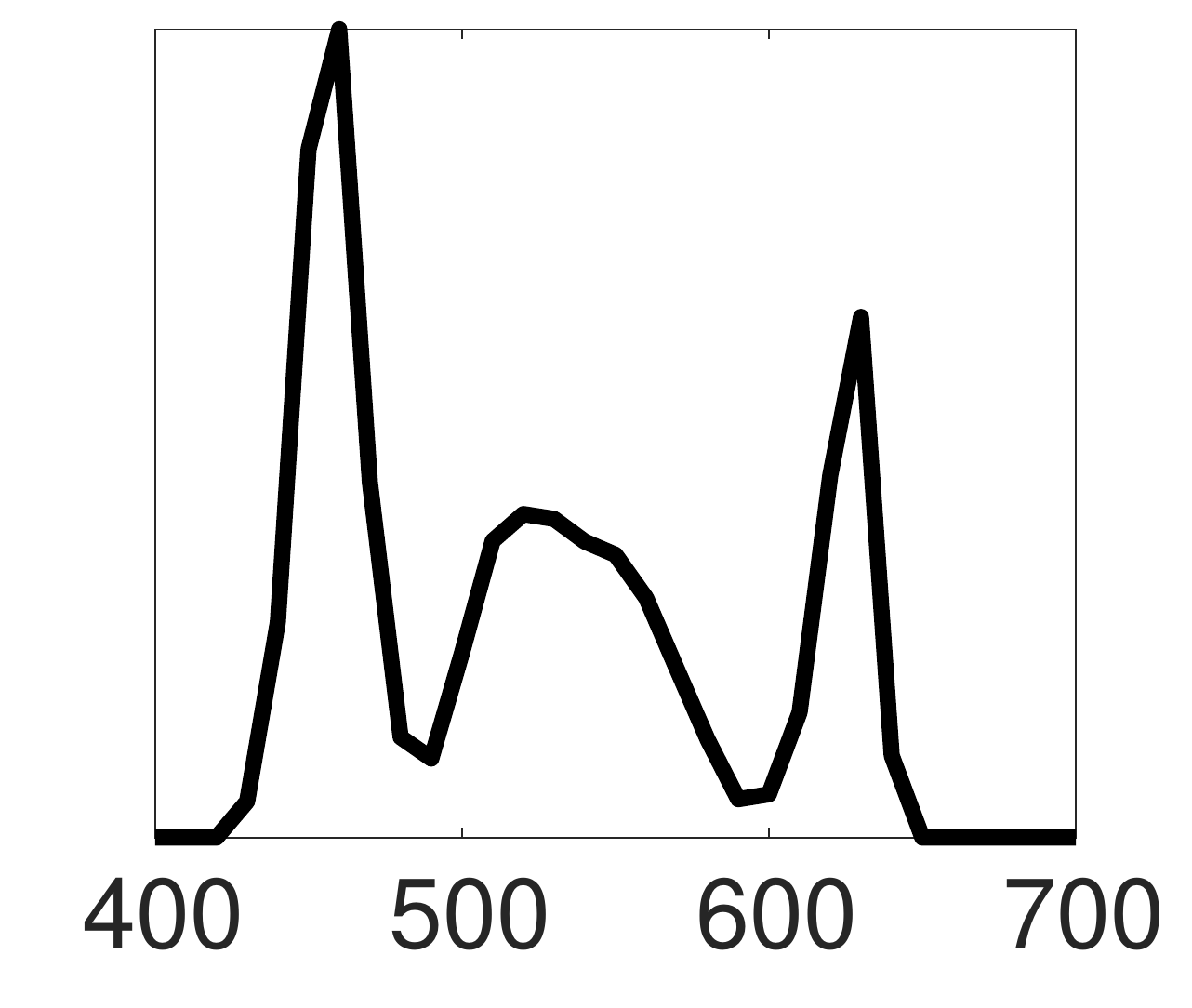}\\
    {\footnotesize Cyan}&{\footnotesize Magenta}&{\footnotesize Yellow}&{\footnotesize White}
  \end{tabular}
  }
  \caption{Camera spectral sensitivity and spectral power distribution of each uniform color illumination.}
  \label{fig:spectrum}
\end{figure}

\begin{figure*}[!tbp]
  \centering
      \begin{minipage}{0.195\hsize}
  \centering
    \includegraphics[width=\hsize]{topviewprojector.pdf}\\
    \small{(a) Synthesized top view}
    \end{minipage}
  \begin{minipage}{0.16\hsize}
  \centering
    \includegraphics[width=\hsize]{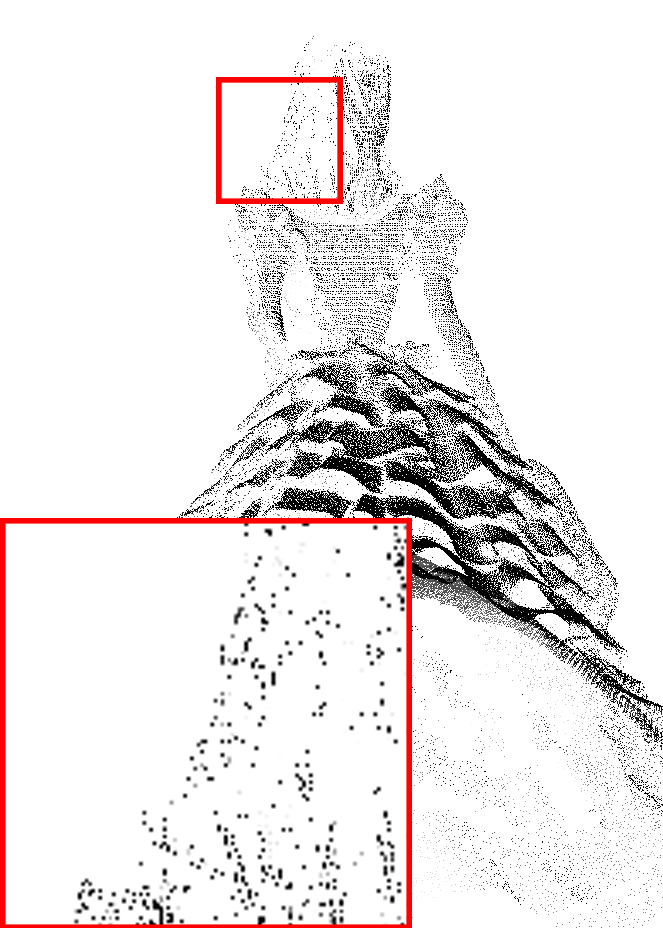}\\
  \small{(b) Method~\cite{Li7}}
    \end{minipage}
    \begin{minipage}{0.16\hsize}
  \centering
    \includegraphics[width=\hsize]{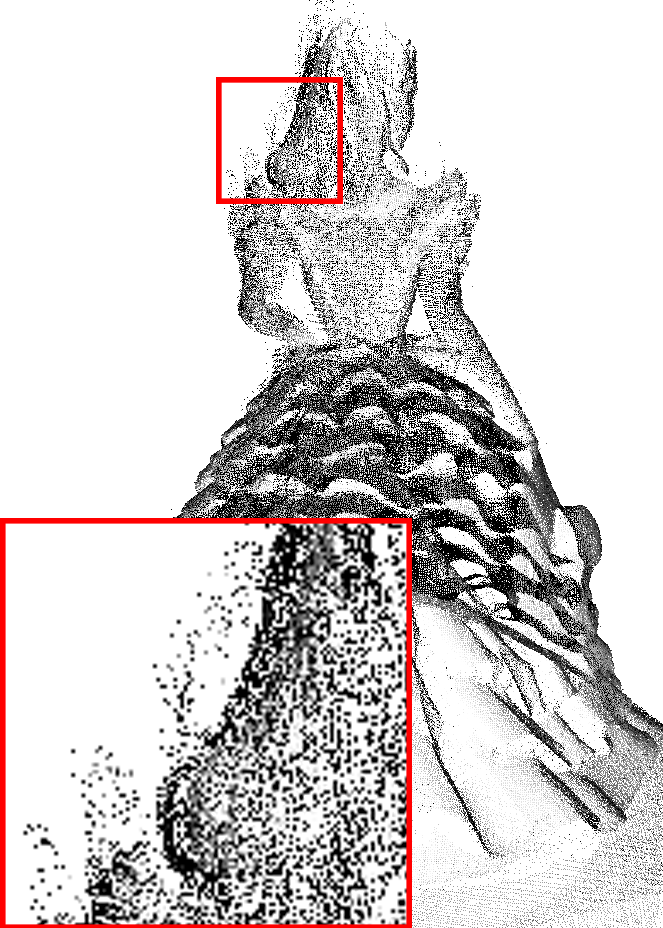}\\
  \small{(c) Ours w/o weight}
  \end{minipage}
    \begin{minipage}{0.16\hsize}
  \centering
    \includegraphics[width=\hsize]{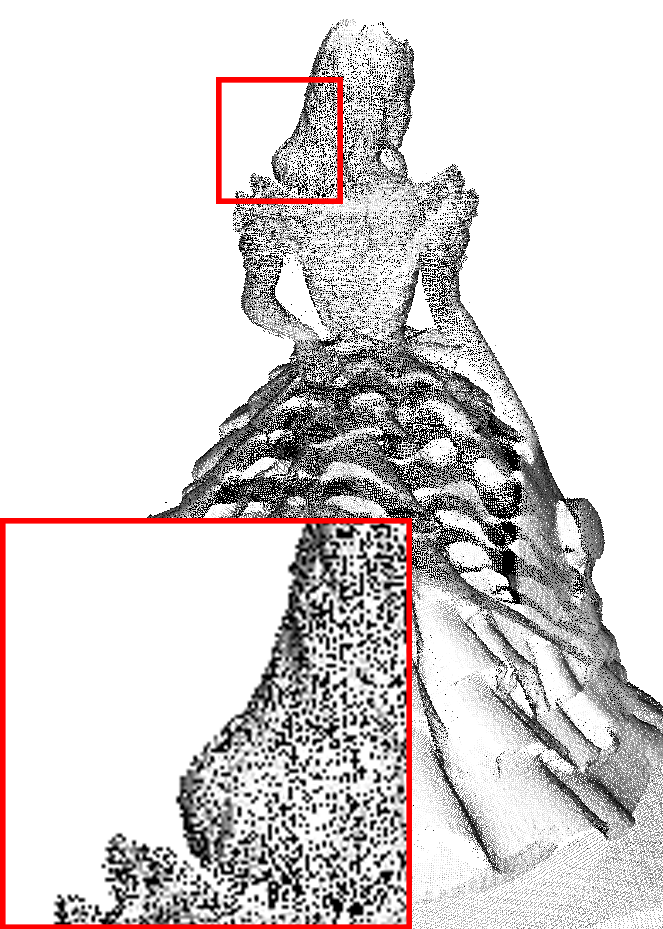}\\
    \small{(d) Our method}
    \end{minipage}
 \begin{minipage}{0.295\hsize}
    \centering
    \begin{minipage}{0.48\hsize}
    	\includegraphics[width=\hsize]{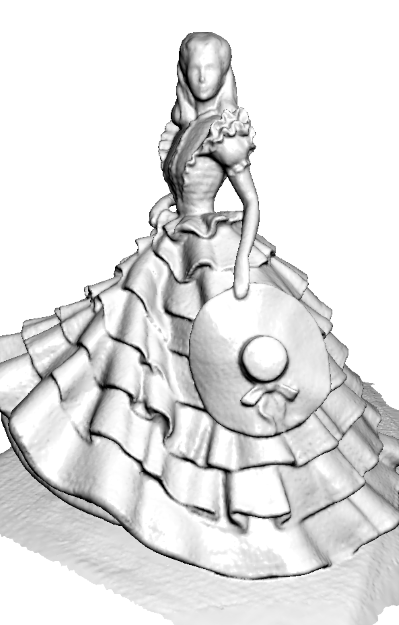}
    \end{minipage}
    \begin{minipage}{0.50\hsize}
    	\includegraphics[width=\hsize]{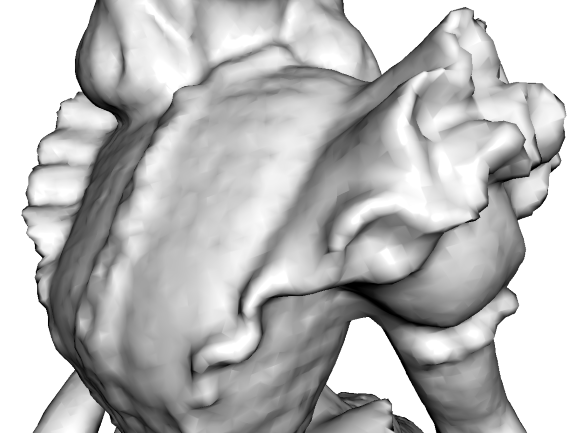}
        \includegraphics[width=\hsize]{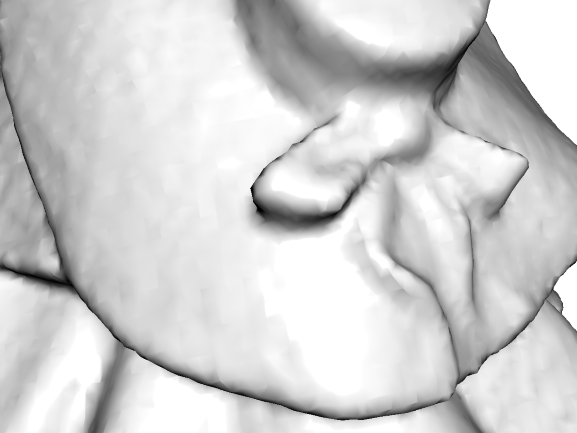}
    \end{minipage}
    \vspace{0.01em}\\
    \small{(e) 3D surface result of our method}
  \end{minipage}\\ \vspace{1.5mm}
  \caption{3D reconstruction results: (a) The synthesized top-view image to schematically show the layout of the moved projector and camera positions. The estimated projector and camera positions are overlaid as the red and the green triangular pyramids; (d) Our method generates {\bf210,523} points, while (b) the method~\cite{Li7} generates {\bf105,915} points; (c) The result of our method without the bundle adjustment weight, which leads to reconstruction errors; (e) The 3D surface result reconstructed from the 3D points (d).}
  \label{fig:3dresult}
\end{figure*}

\begin{figure}[t]
\vspace{-1em}
\centering 
\begin{minipage}{0.46\hsize}
\centering
\includegraphics[width=\hsize]{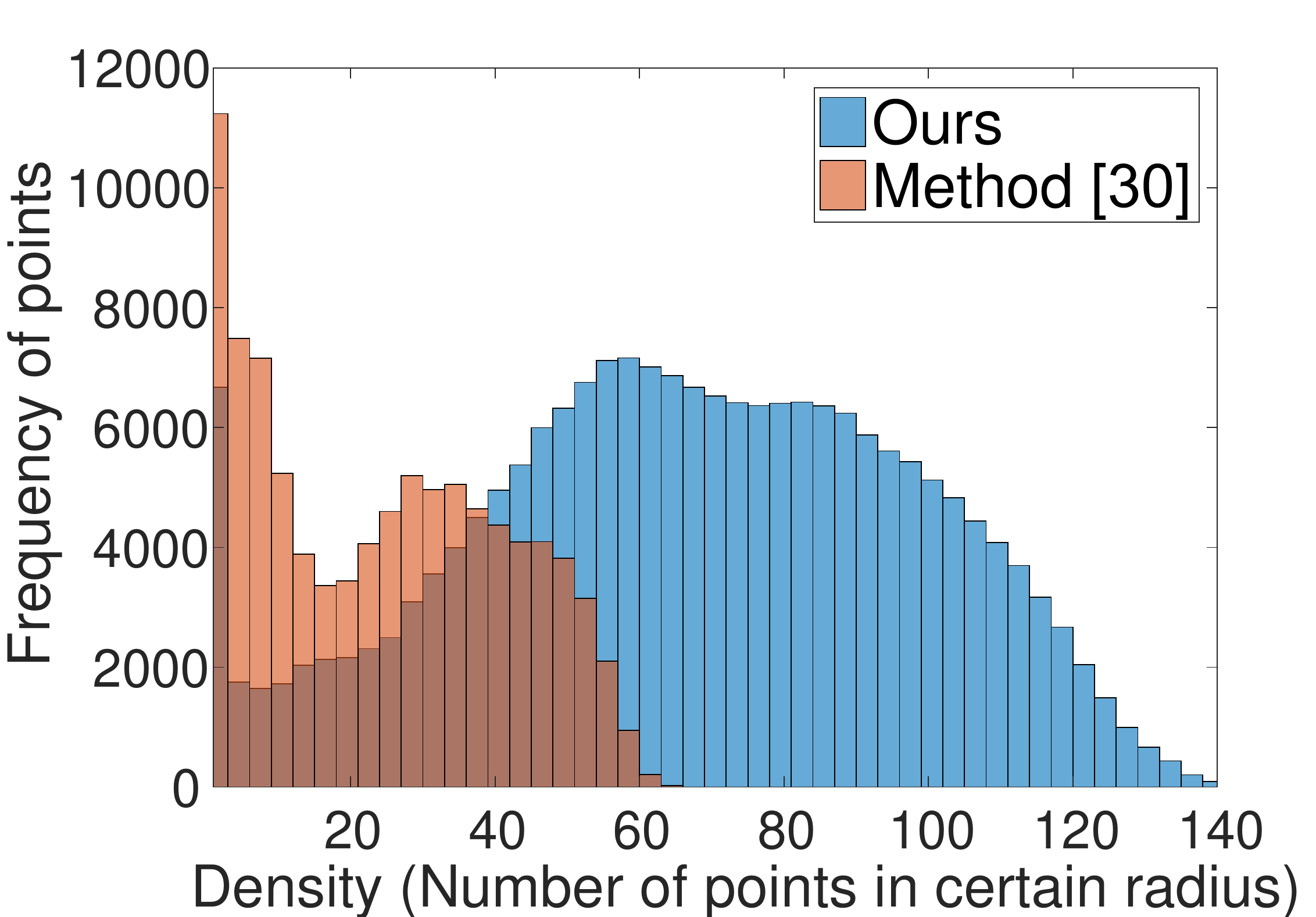}
\end{minipage}
\begin{minipage}{0.52\hsize}
\centering
    \begin{minipage}{0.415\hsize}
  \centering
    \includegraphics[width=\hsize]{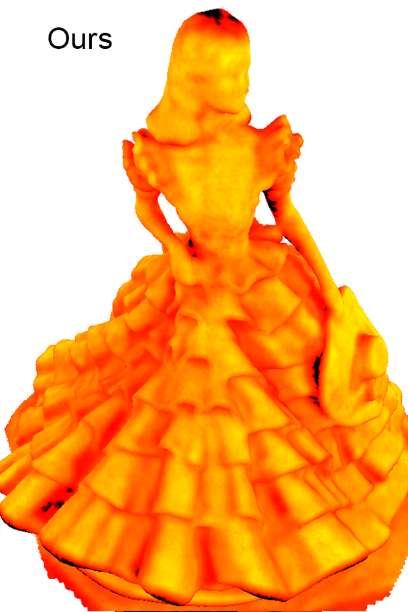}
  \centering
  \centering
  \end{minipage}
    \begin{minipage}{0.415\hsize}
    \includegraphics[width=\hsize]{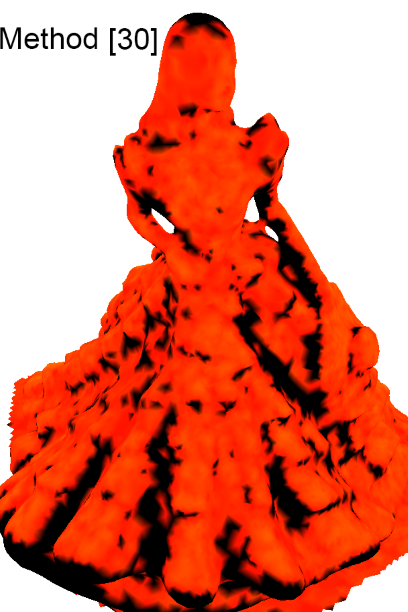}
  \centering
  \end{minipage}  
  \begin{minipage}{0.13\hsize}
    \centering
    \includegraphics[width=\hsize]{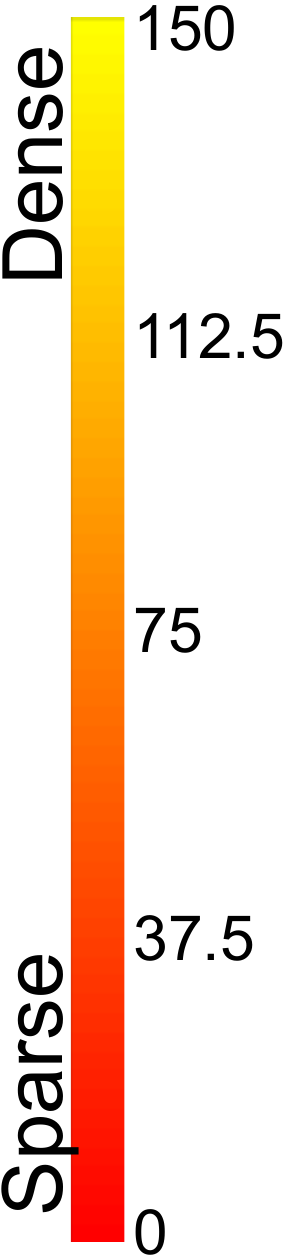}
  \end{minipage}
\end{minipage}
\vspace{0.5em}
\caption{Evaluation of 3D point cloud density. Left: The density histogram of reconstructed points. Right: Visualization of its spacial distribution.}
\label{fig:density}
\vspace{-1em}
\end{figure}

\section{Experimental Results}
\label{sec:results}

\subsection{Setup and implementation details}

We used an ASUS P3B projector and a Canon EOS 5D Mark-II digital camera. The sequence of the structured light patterns was captured using a video format with 1920$\times$1080 resolution, while the color illuminations were captured using a RAW format, which has a linear camera response, with a higher resolution. The RAW images were then resized to have the same resolution with the video format. As shown in~Fig.~\ref{fig:spectrum}, the camera spectral sensitivity of Canon EOS 5D Mark-II was obtained from the camera sensitivity database~\cite{Jiang} and the spectral power distributions of the color illuminations were measured using a spectrometer. 

For the 3D reconstruction, we used Colmap~\cite{Schoenberger} to run the SfM pipeline and Poisson surface reconstruction~\cite{Kazhdan,Kazhdan2}, which is integrated with Meshlab~\cite{Cignoni}, to visualize the obtained 3D model. For the spectral reflectance estimation, we set the target wavelength range as 410nm to 670nm with every 10nm intervals because the used projector illuminations only have the spectral power within this range. We used eight basis functions, which were calculated using the spectral reflectance data of 1269 Munsell color chips~\cite{Munsell} by principal component analysis. The spectral smoothness weight in Eq.~(\ref{eq:costfunction}) was determined by an empirical manner and set as $\gamma = 0.06$ for the intensity range [0,1]. The C++ Ceres solver~\cite{ceres-solver} was used to solve the non-linear optimization problem of Eq.~(\ref{eq:costfunction}).

\subsection{3D reconstruction results}

Figure~\ref{fig:3dresult} shows the self-calibrating 3D reconstruction results of a clay sculpture with roughly 30 centimeter height. To schematically show the layout of the moved projector and camera positions, we show a synthesized top-view image of Fig.~\ref{fig:3dresult}(a).
The estimated projected and camera positions by our method are overlaid as the red and the green triangular pyramids using a manually aligned scale. It is demonstrated that the projector and the camera positions are correctly estimated by our method.

Figure~\ref{fig:3dresult}(b) and~\ref{fig:3dresult}(d) show the 3D point cloud results of the method~\cite{Li7} and our method, which were reconstructed using exactly the same setup and images. The difference is that our method uses both camera and projector images for SfM computation, while the method~\cite{Li7} uses only camera images. Our method can reconstruct the 210,523 points, which are almost double of the 105,915 points reconstructed by the method~\cite{Li7}. To quantitatively evaluate the point cloud density, we counted the number of reconstructed 3D points within a certain radius from each point. Figure~\ref{fig:density} shows the density histogram of reconstructed points (left) and the visualization of its spacial distribution (right), where our method achieves much denser 3D reconstruction.

Another improvement can be achieved by introducing the weight of Eq.~(\ref{eq:reprojection}) for bundle adjustment, which poses larger penalties to the projector's reprojection errors. In our experiments, we set as $w_p$ = 100, though a larger value more than 10 does not make a big difference. We experimentally observed that the estimation of the projector's positions and internal parameters often fails if we do not use the weight, which leads to reconstruction errors as can be seen in Fig.~\ref{fig:3dresult}(c). Figure~\ref{fig:3dresult}(e) shows the final reconstructed surface result for our method, where the detail structures of the sculpture are precisely reconstructed.

\begin{figure}[!t]
\vspace{-1em}
  \centering
    \includegraphics[width=0.61\hsize]{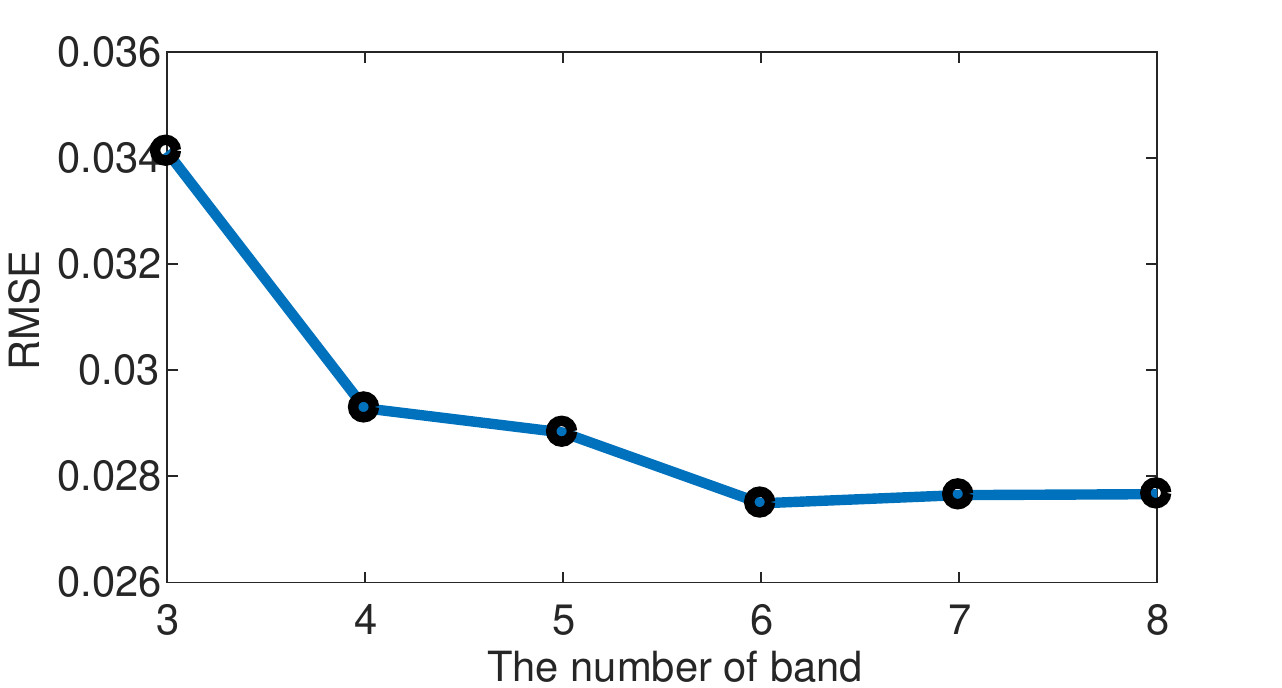}
  \caption{RMSE for the 24 patches of the colorchart when using the selected best band set for each number of spectral bands.}
  \label{fig:multibandselection}
  \vspace{-0.5em}
\end{figure}

\begin{figure*}[!tbp]
  \centering
  \begin{minipage}{\hsize}
      \begin{minipage}{0.21\hsize}
      \centering
      \includegraphics[width=\hsize]{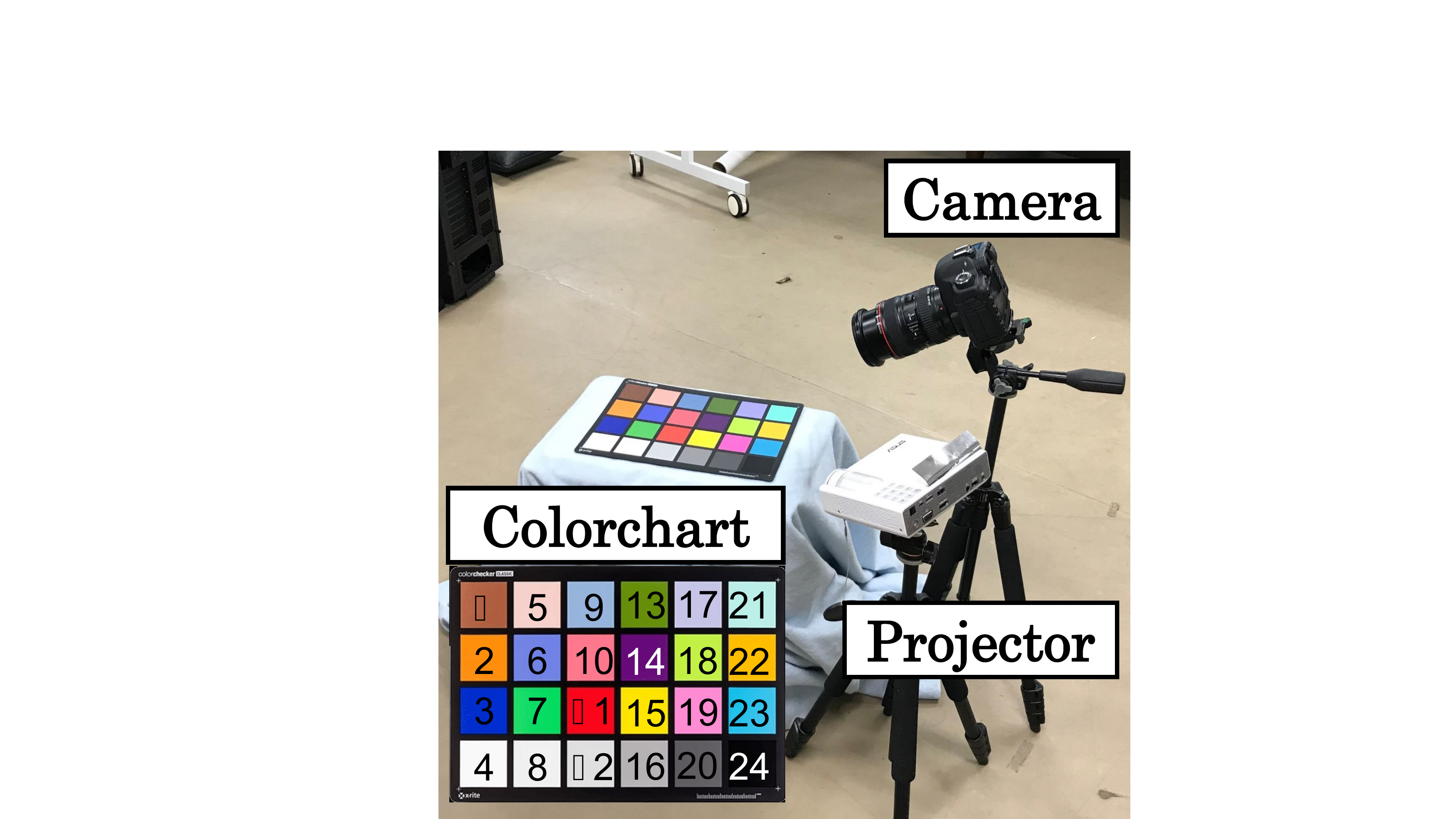}\\
      \small{(a) Experimental setup} 
      \vspace{2mm}\\
      \includegraphics[width=\hsize]{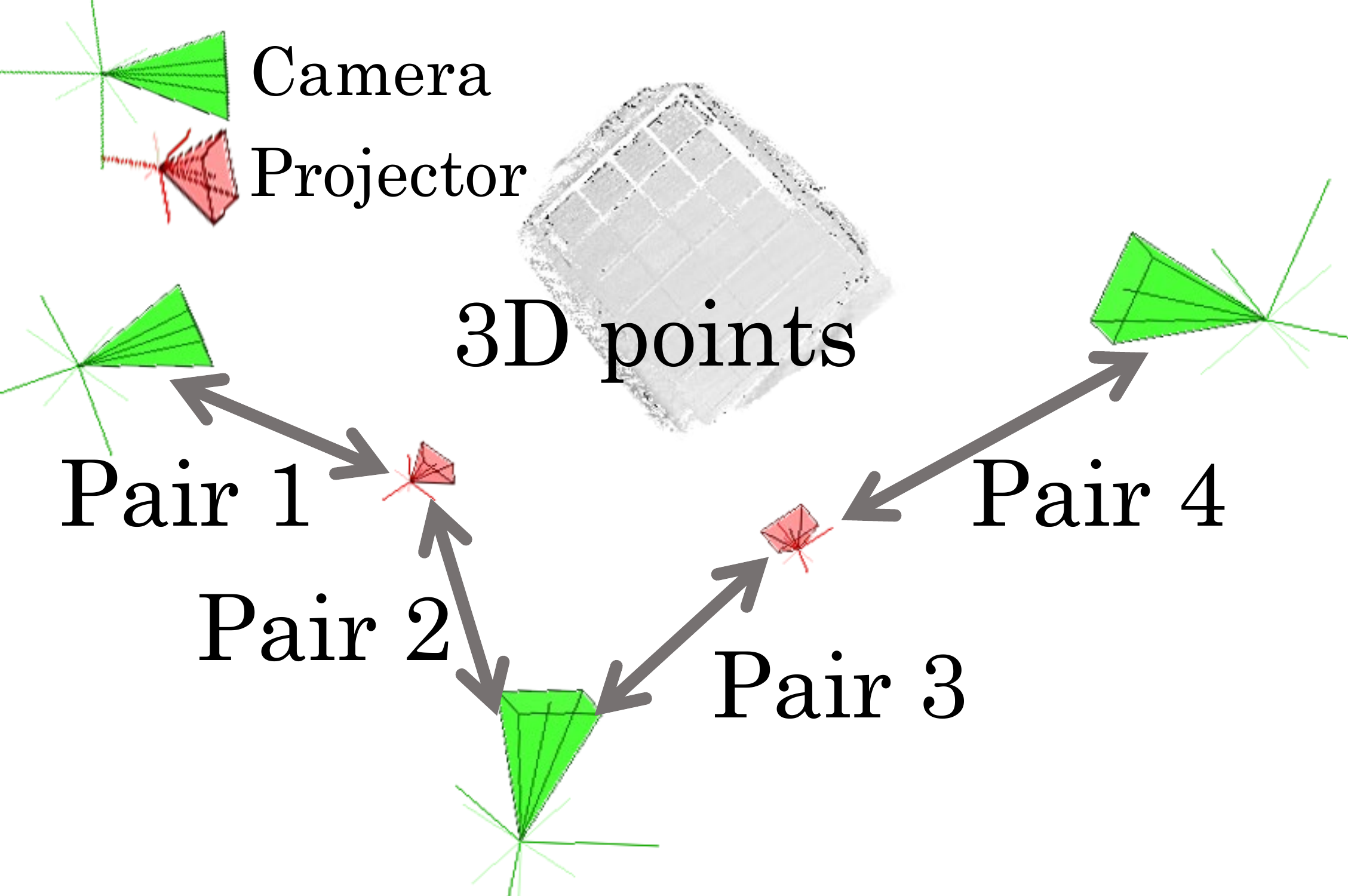}\\
      \small{(b) Estimated 3D points and projector and camera poses}
      \end{minipage}
　　　\hspace{1mm}
      \begin{minipage}{0.78\hsize}
      \centering
      \begin{tabular}{@{\hskip 0pt}c@{\hskip 0pt}c@{\hskip 0pt}c@{\hskip 0pt}c@{\hskip 0pt}c@{\hskip 0pt}c@{\hskip 0pt}}
    	\includegraphics[width=0.166\linewidth]{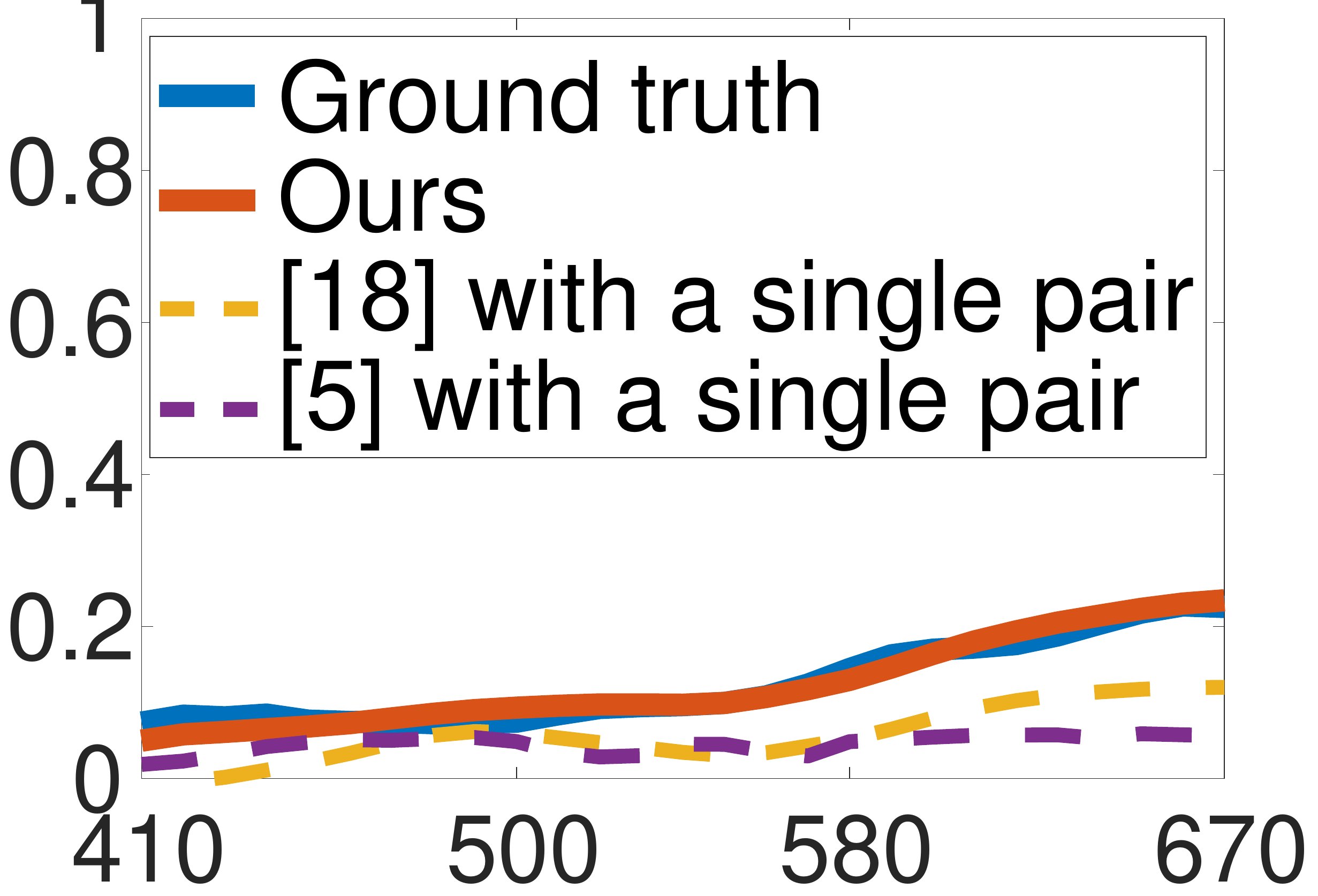} &
    	\includegraphics[width=0.166\linewidth]{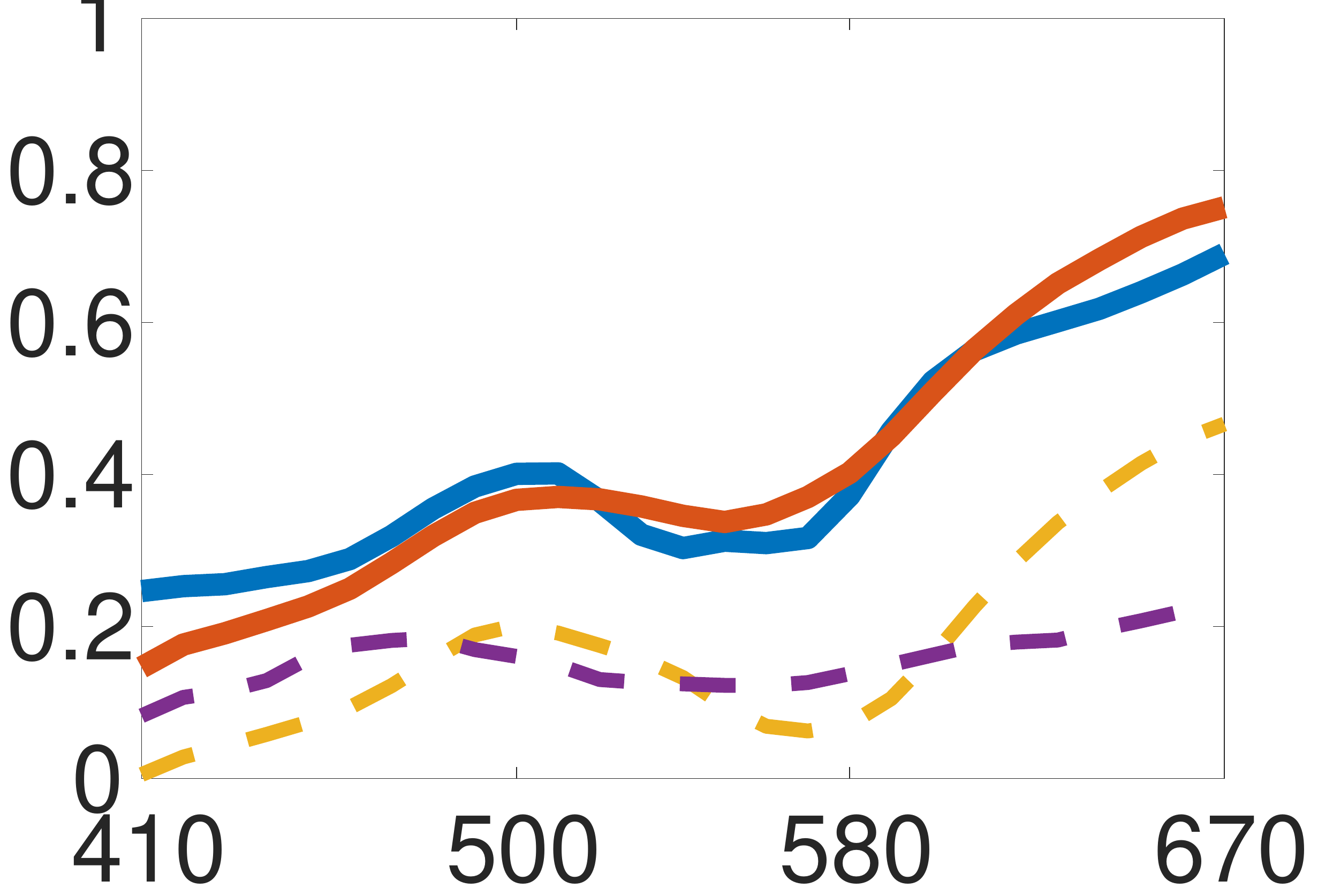} &
    	\includegraphics[width=0.166\linewidth]{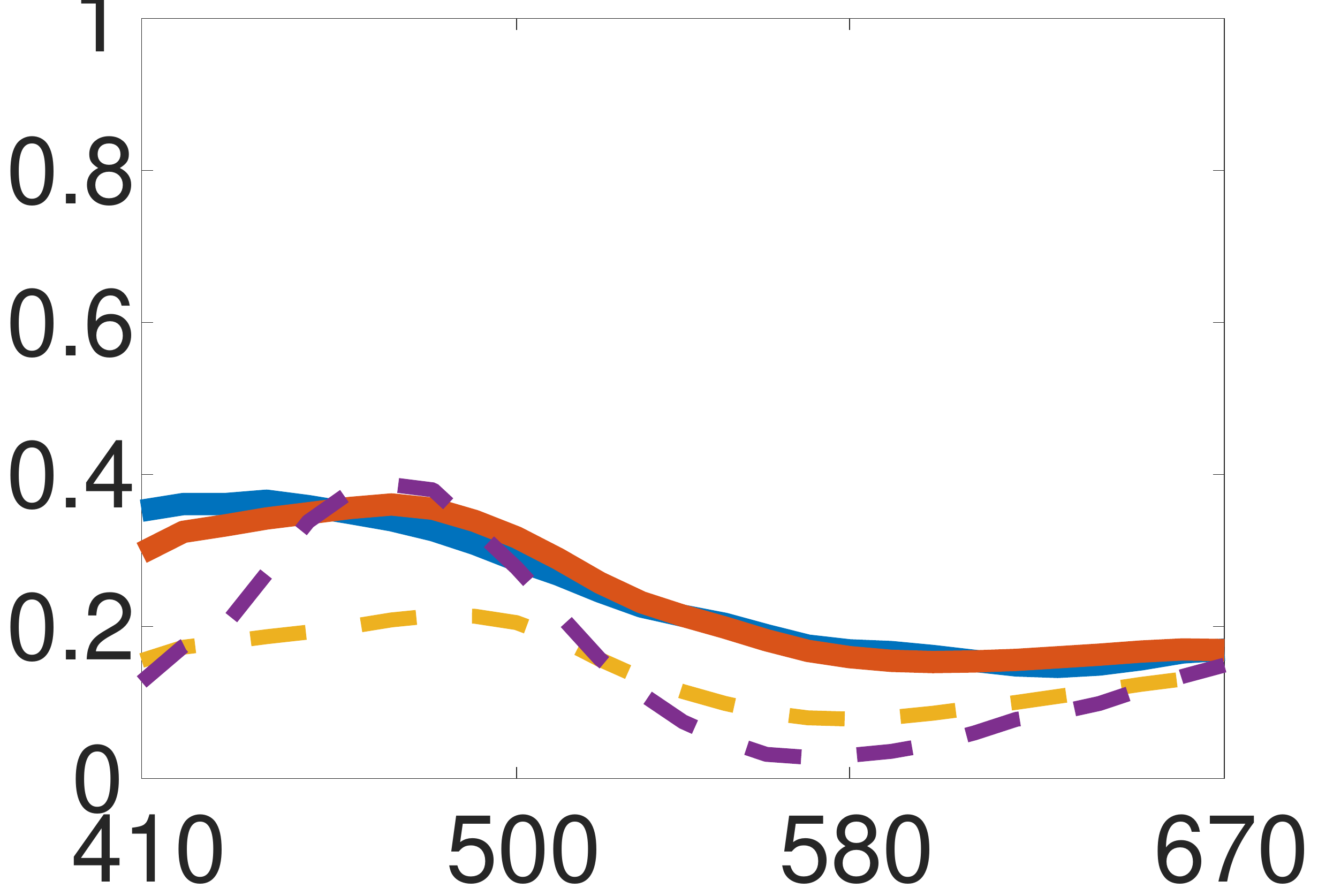} &
    	\includegraphics[width=0.166\linewidth]{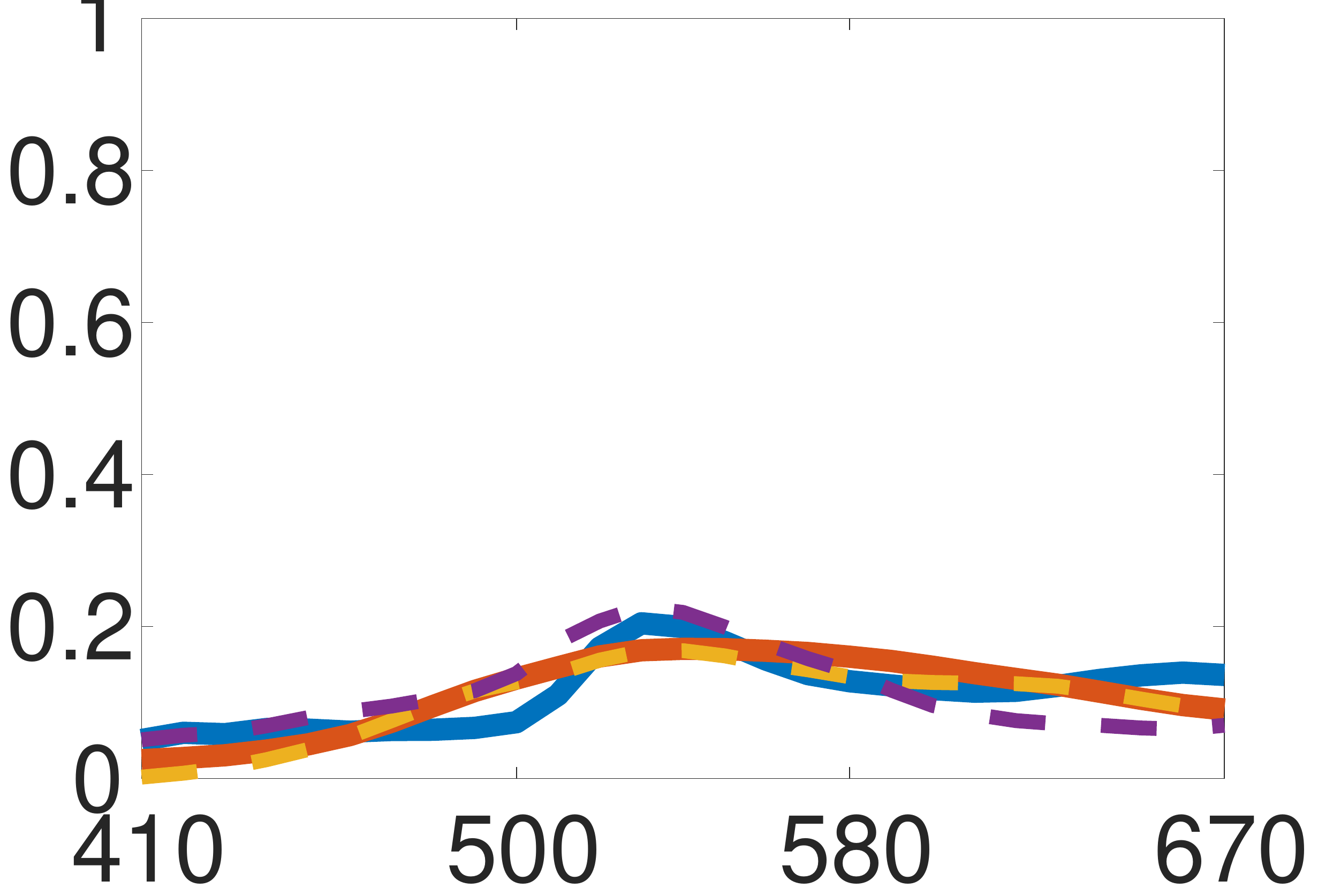} &
    	\includegraphics[width=0.166\linewidth]{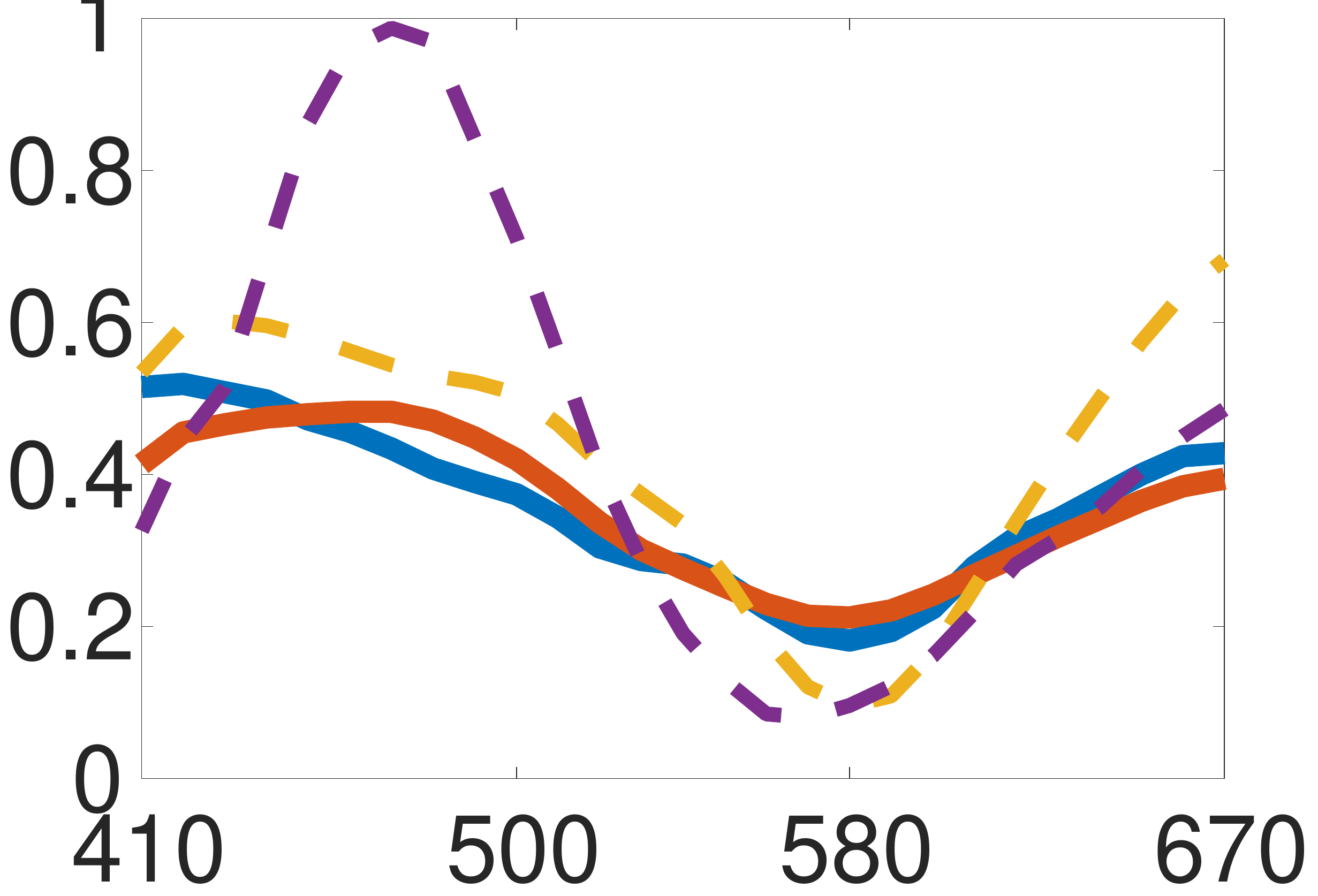} &
    	\includegraphics[width=0.166\linewidth]{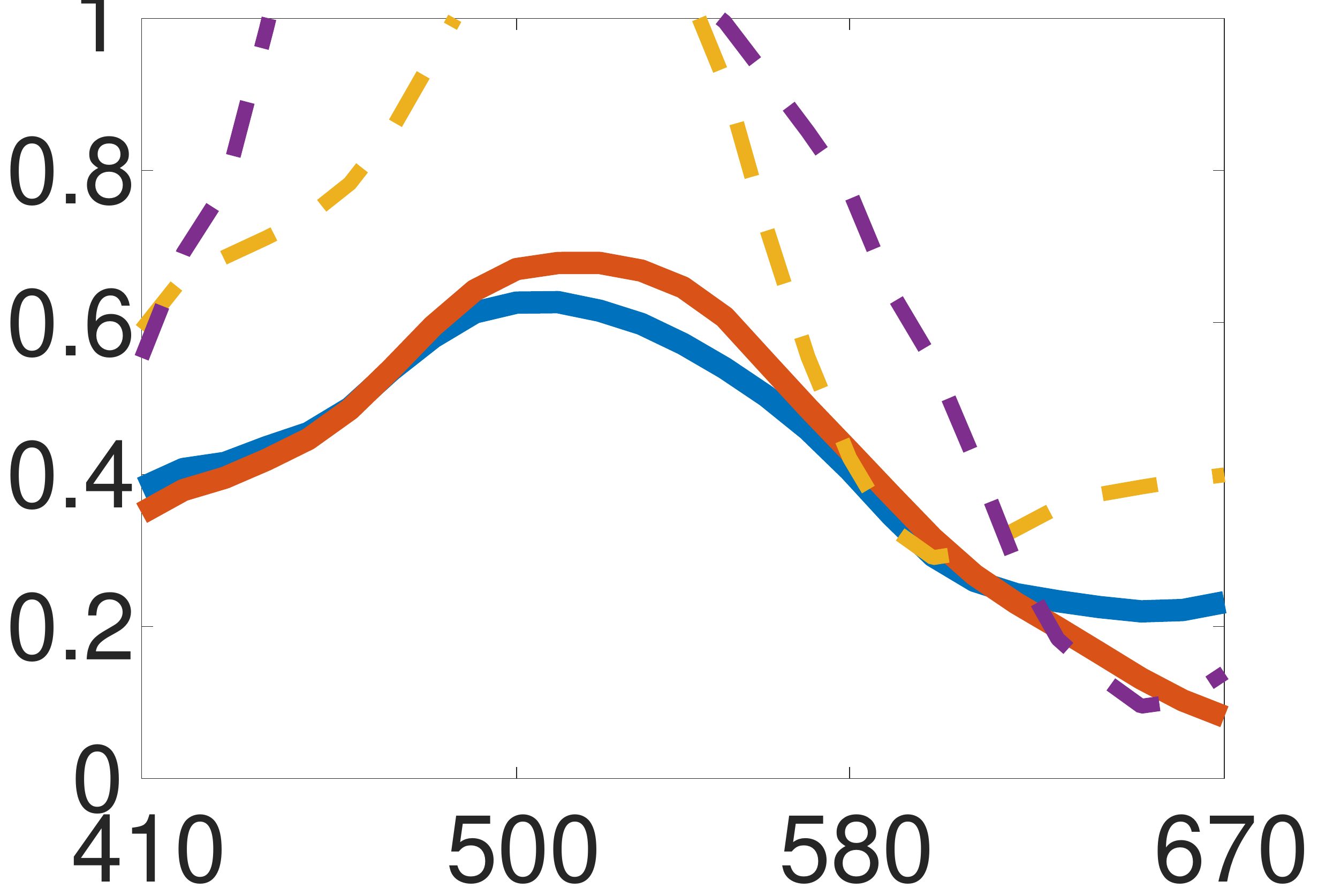} \\
    		\includegraphics[width=0.166\linewidth]{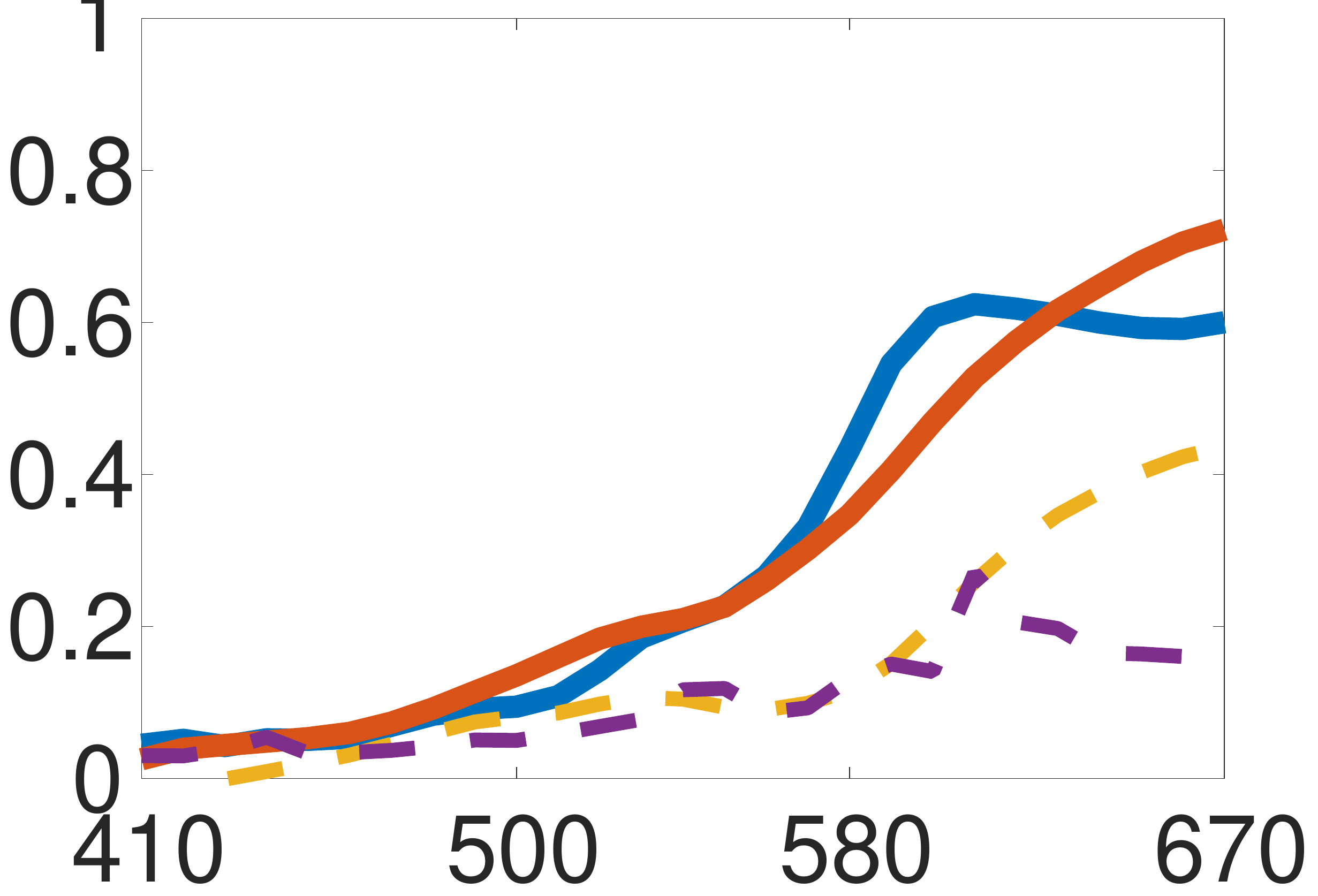} &
    	\includegraphics[width=0.166\linewidth]{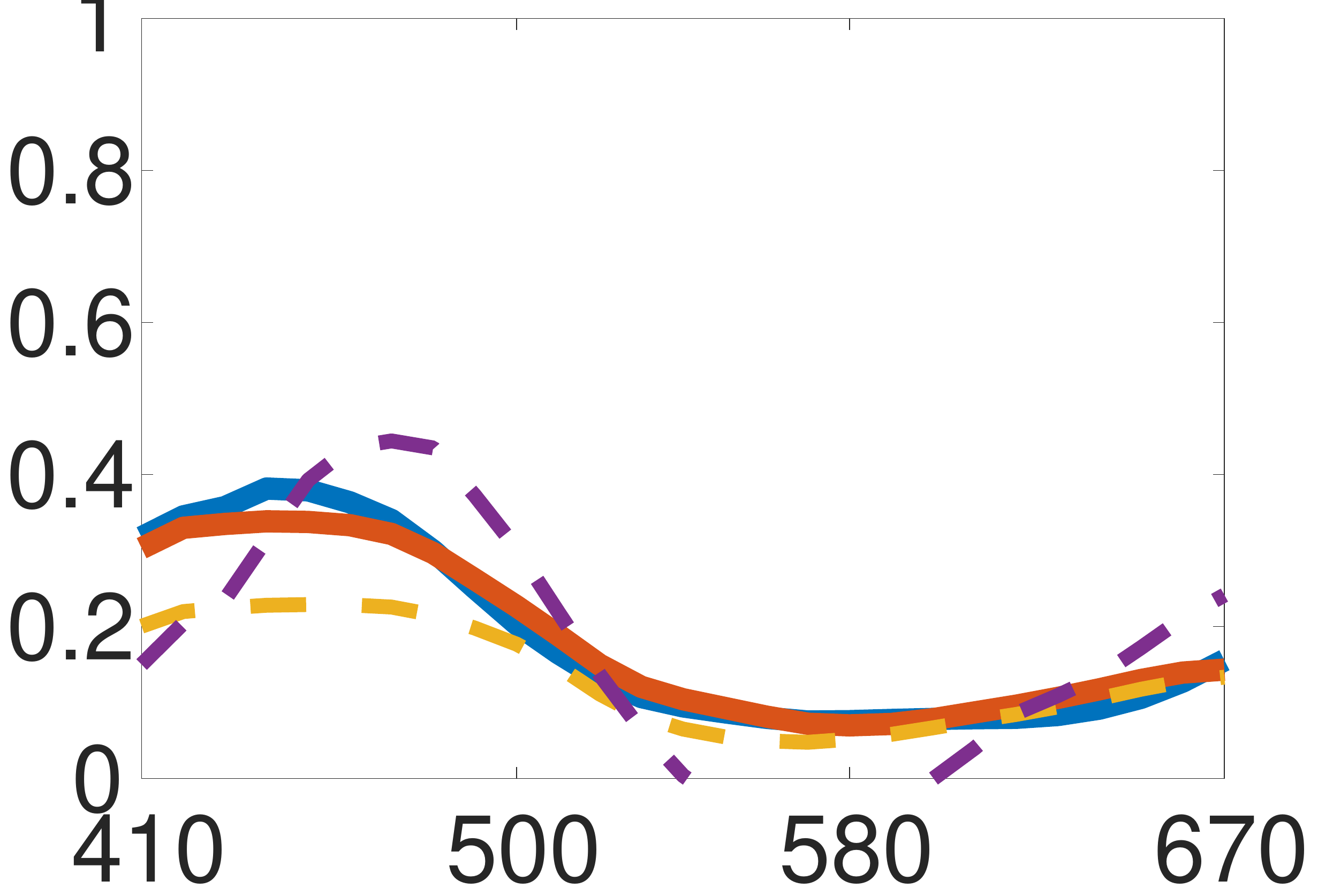} &
    	\includegraphics[width=0.166\linewidth]{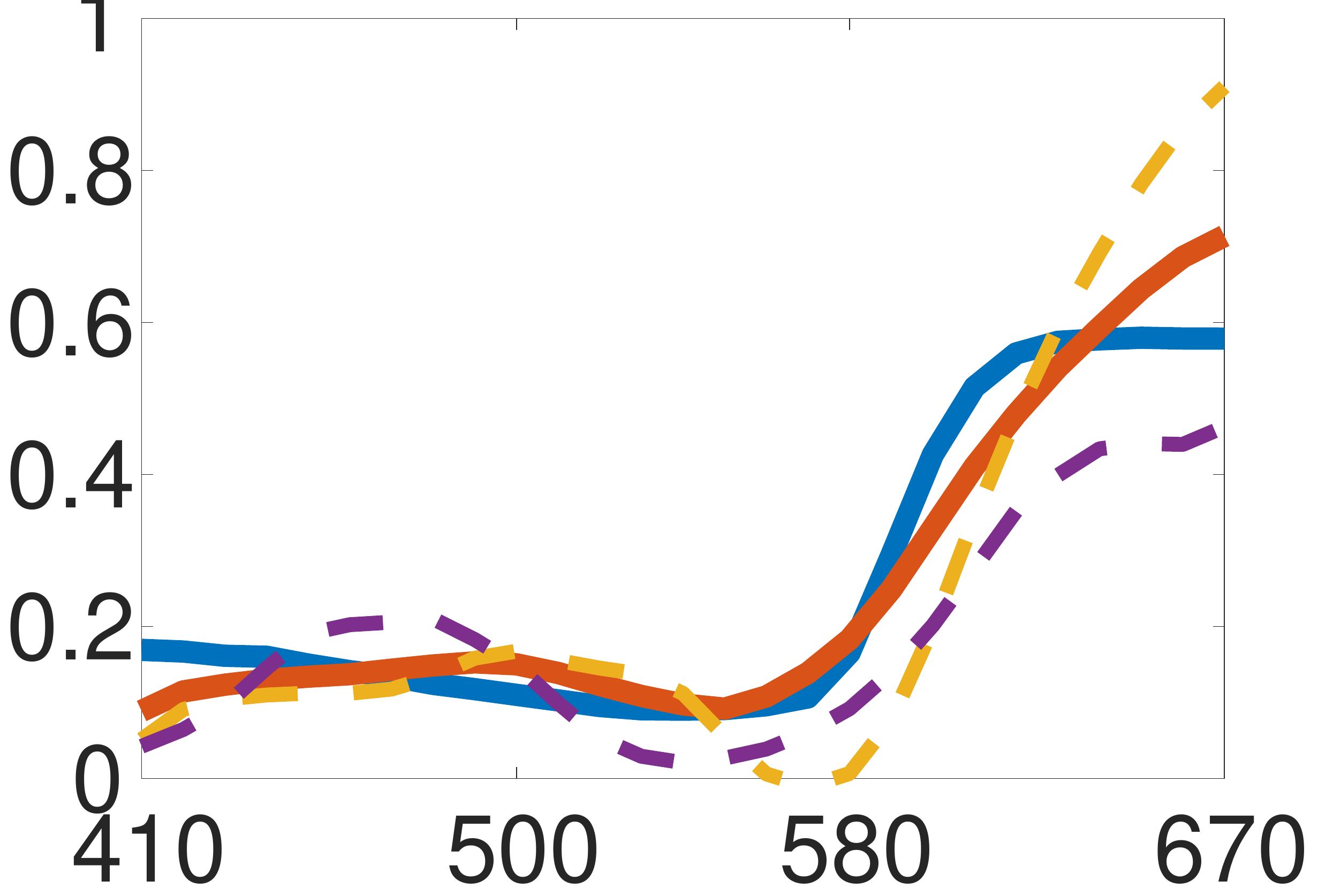} &
    	\includegraphics[width=0.166\linewidth]{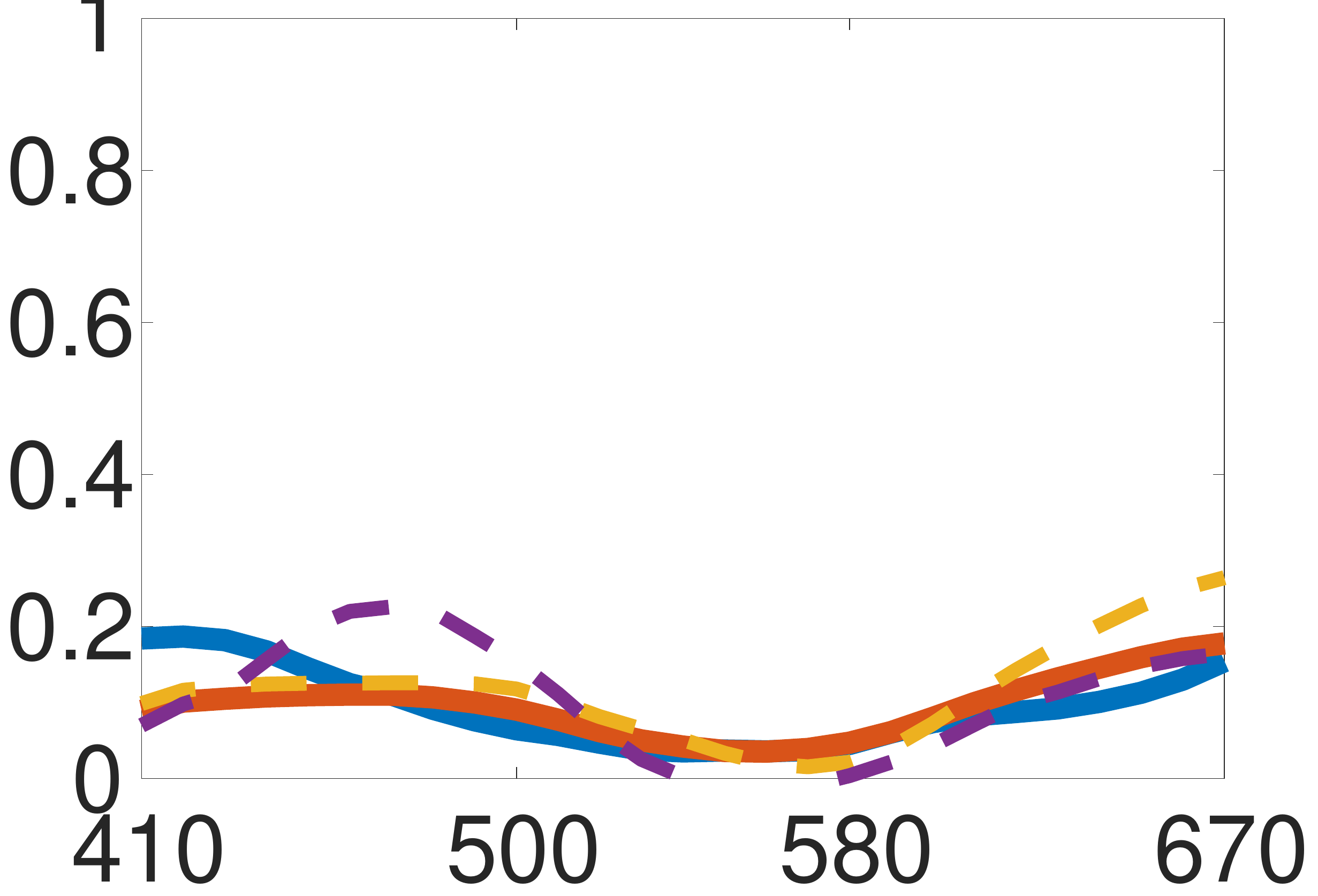} &
    	\includegraphics[width=0.166\linewidth]{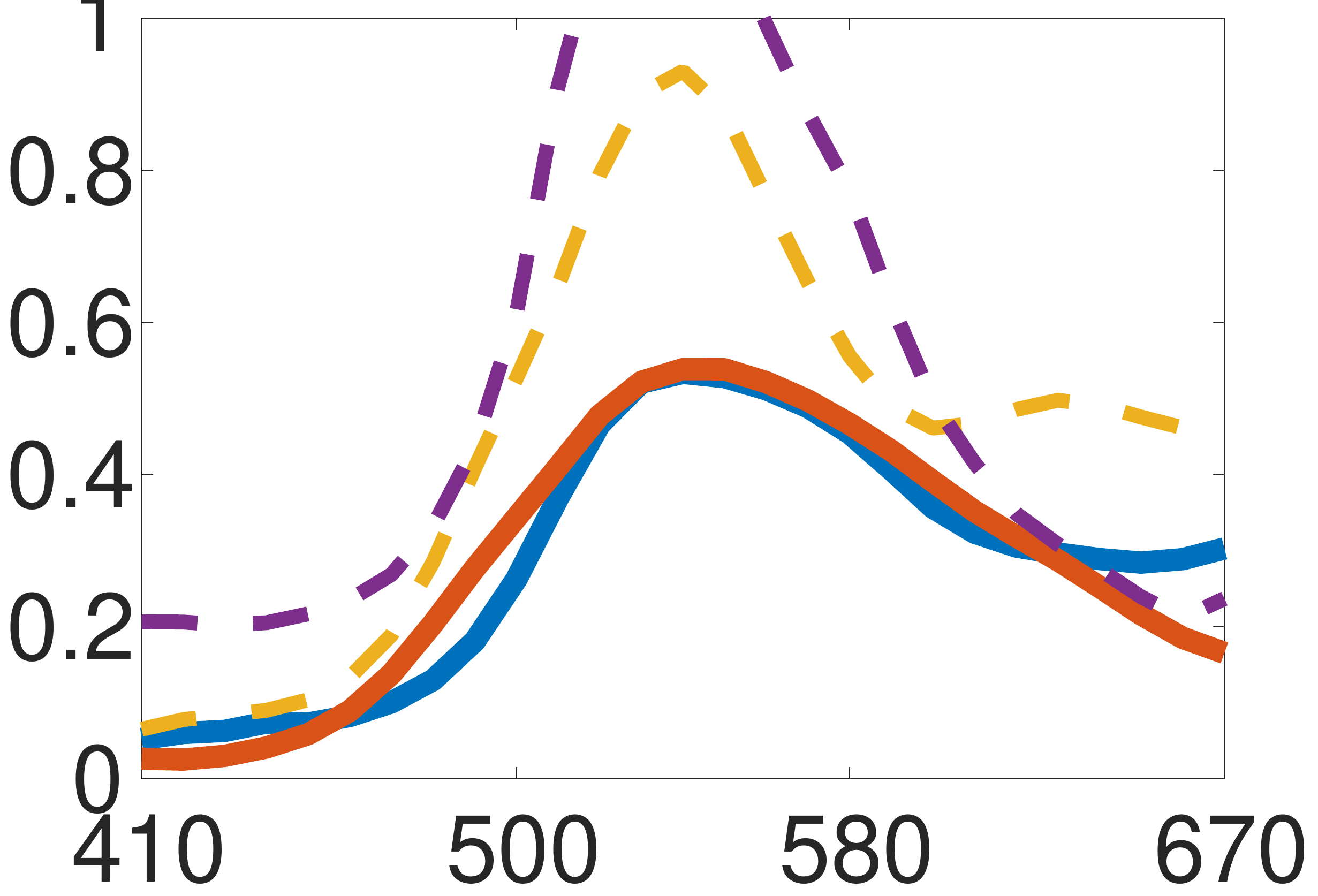} &
    	\includegraphics[width=0.166\linewidth]{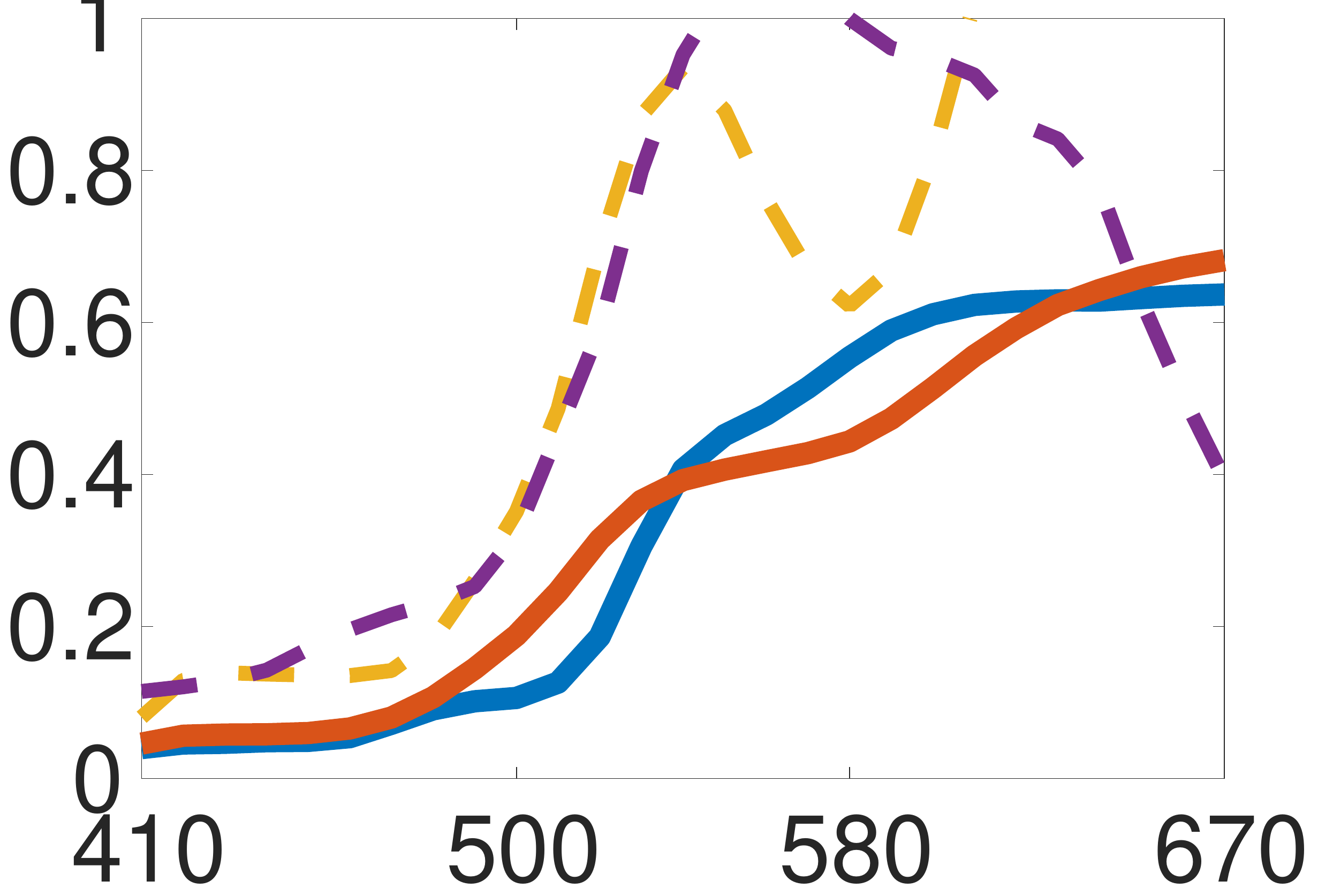} \\
    		\includegraphics[width=0.166\linewidth]{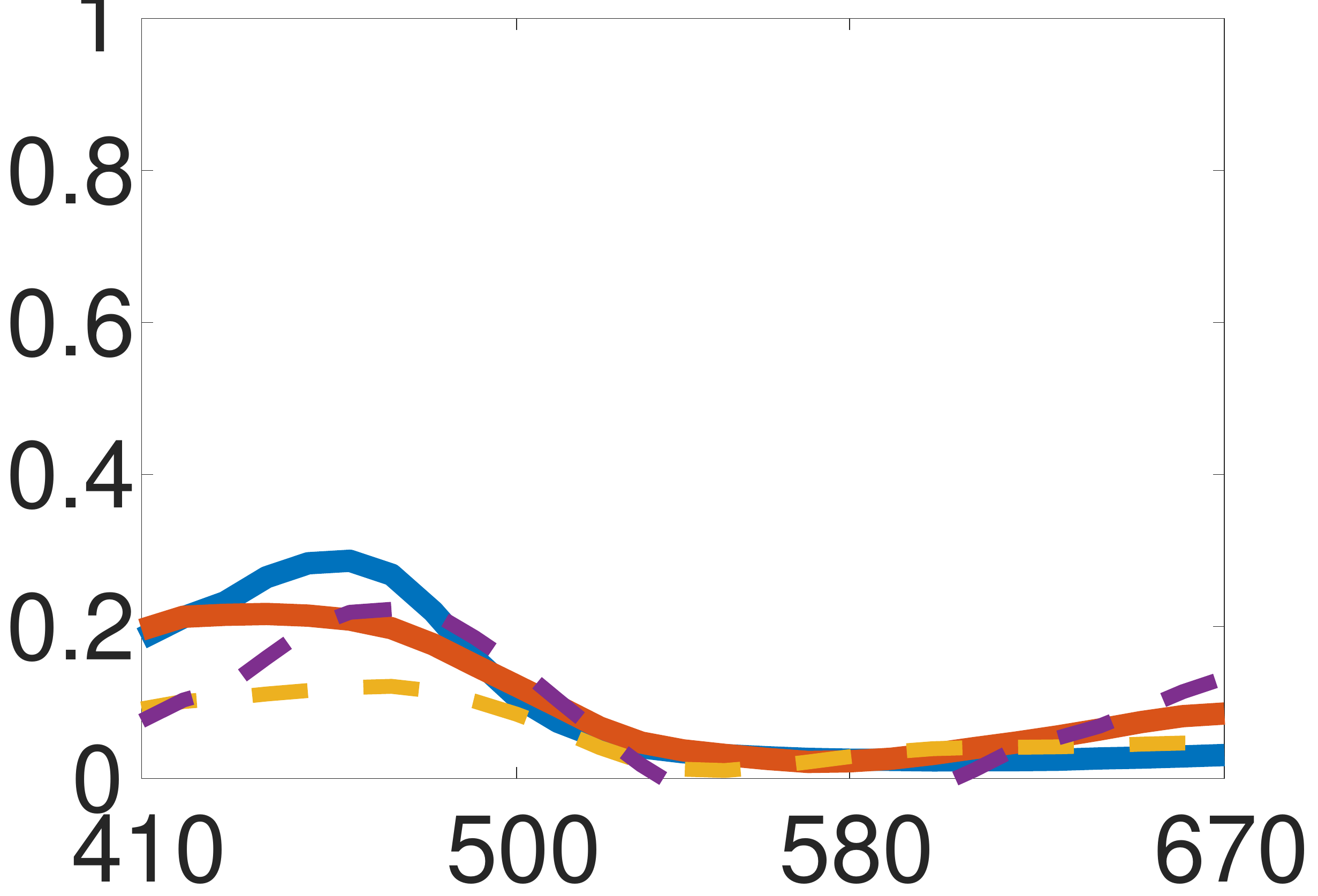} &
    	\includegraphics[width=0.166\linewidth]{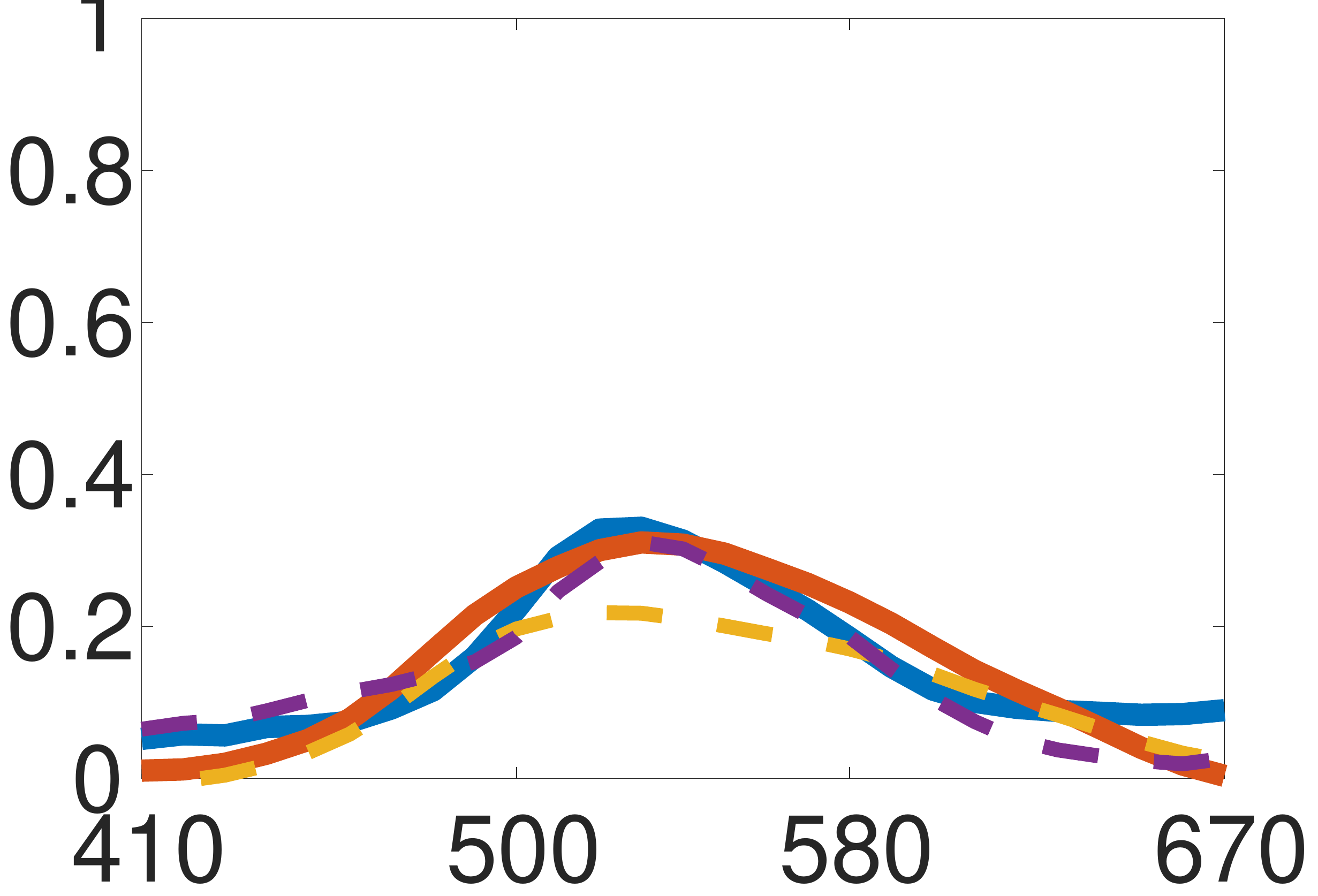} &
    	\includegraphics[width=0.166\linewidth]{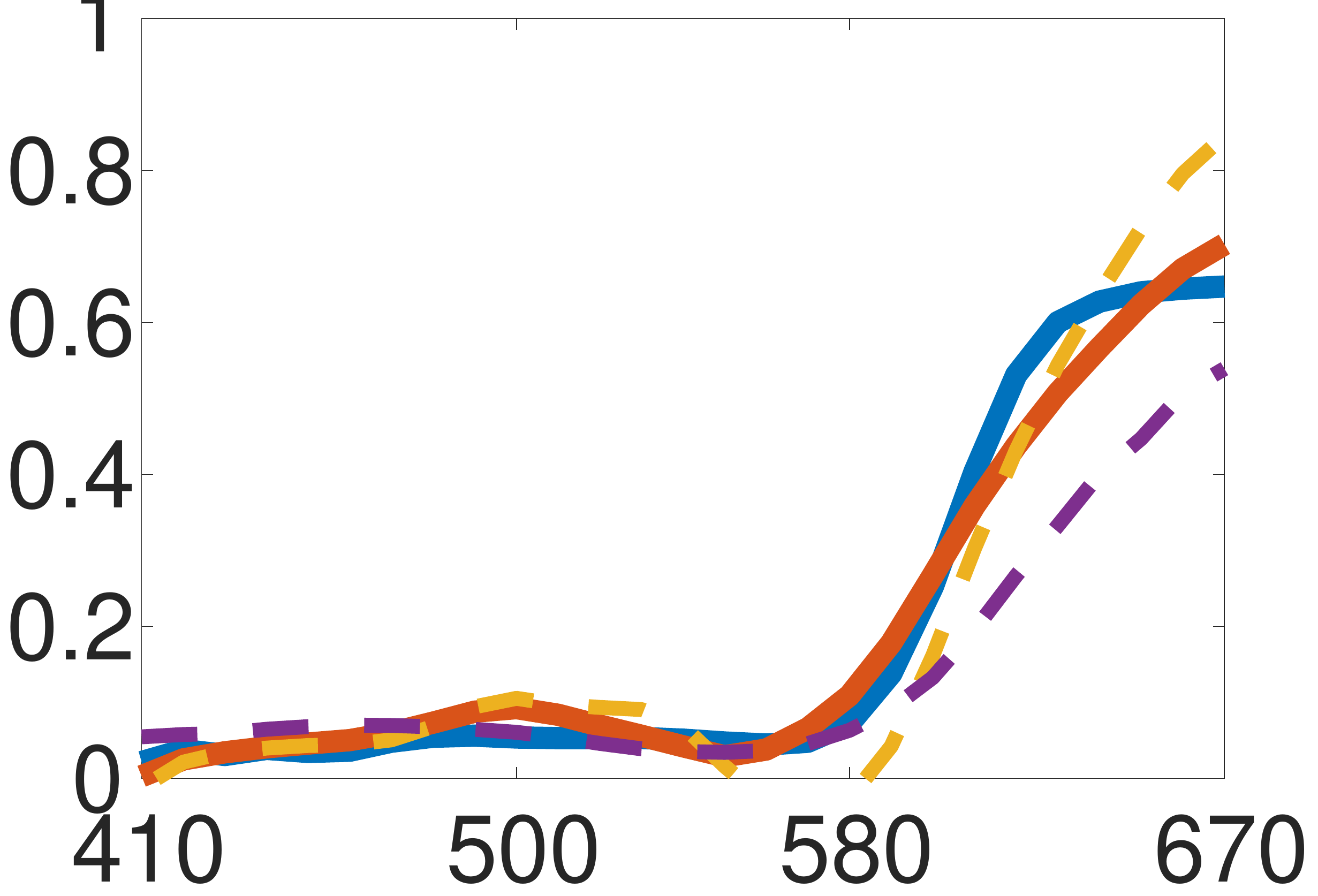} &
    	\includegraphics[width=0.166\linewidth]{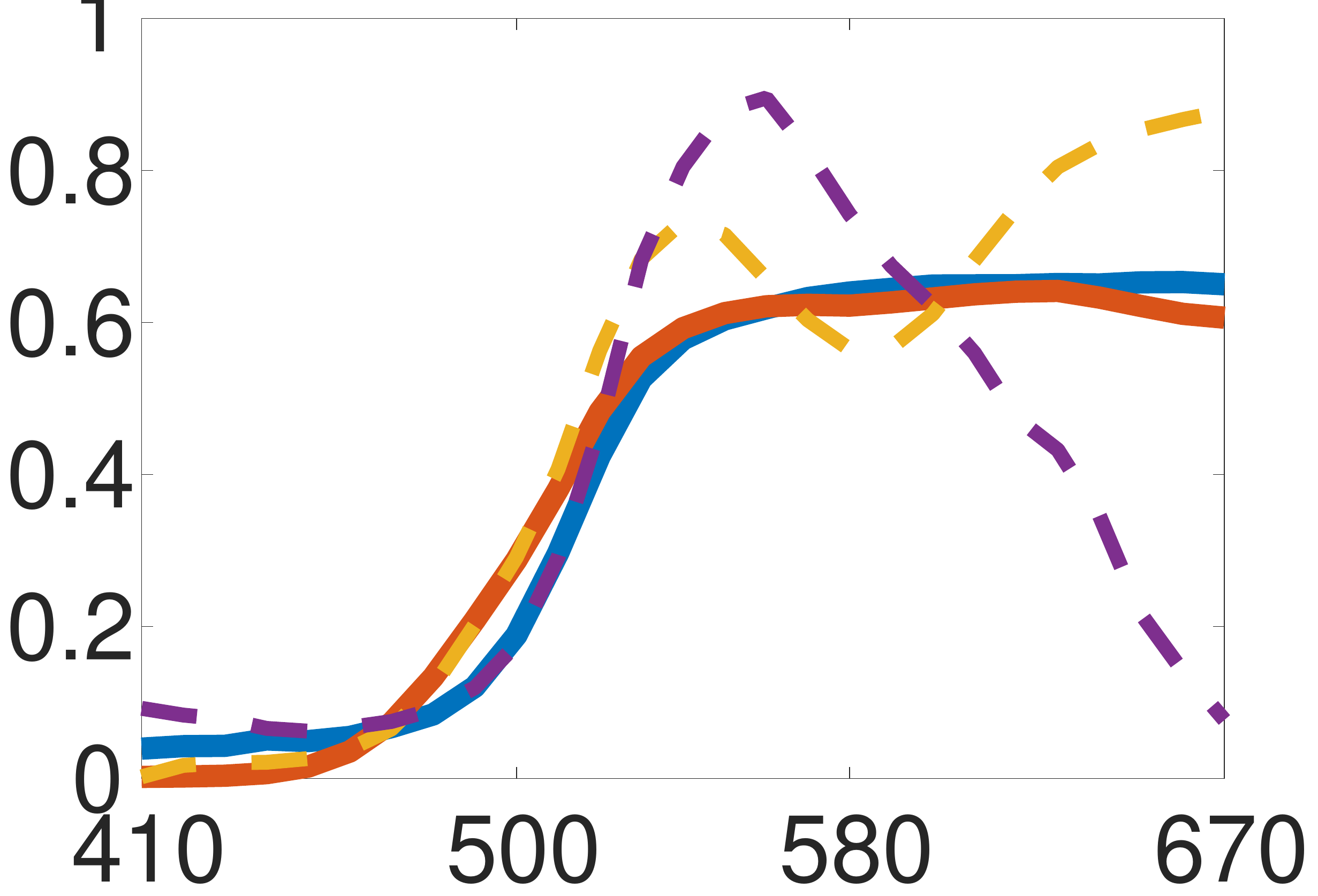} &
    	\includegraphics[width=0.166\linewidth]{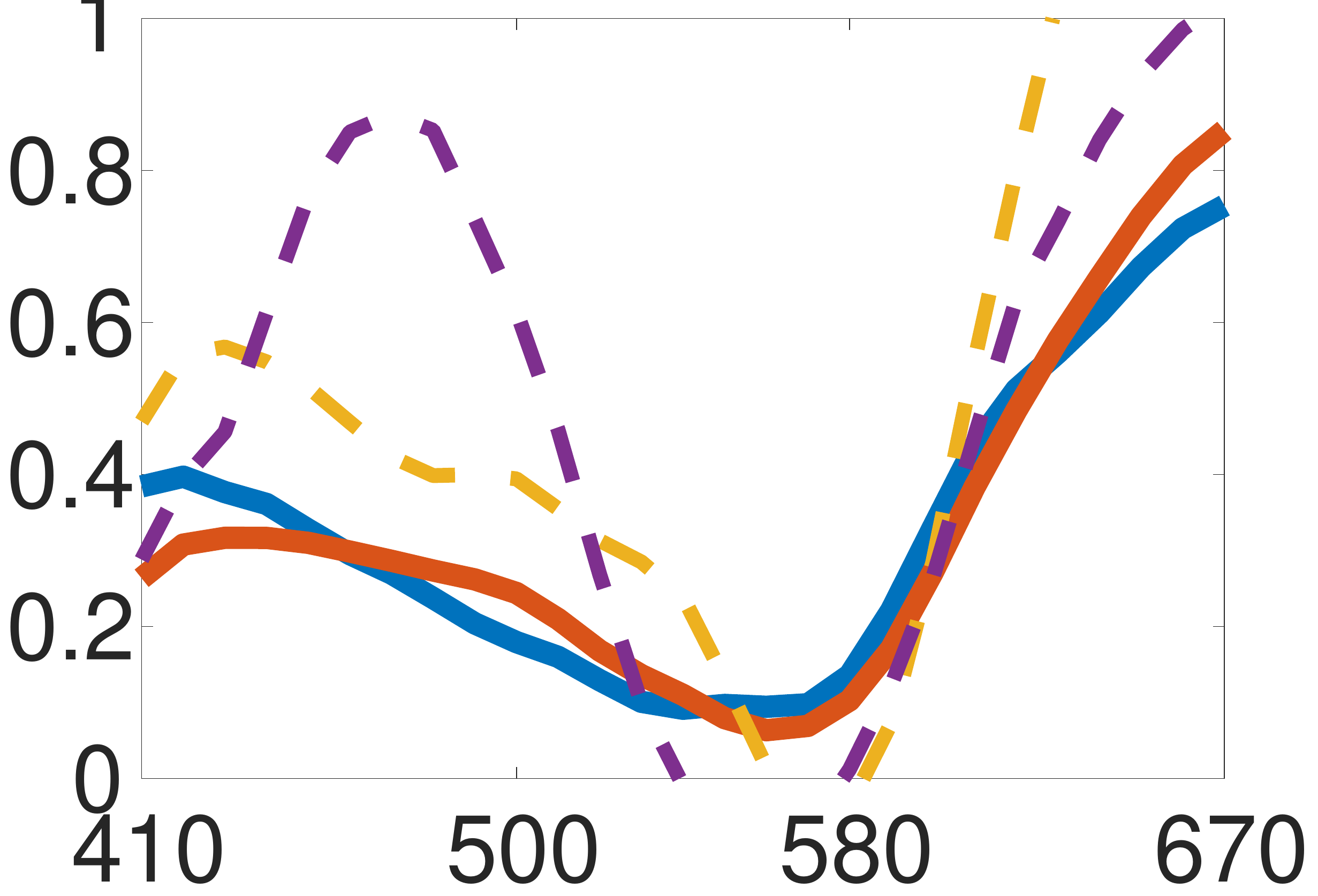} &
    	\includegraphics[width=0.166\linewidth]{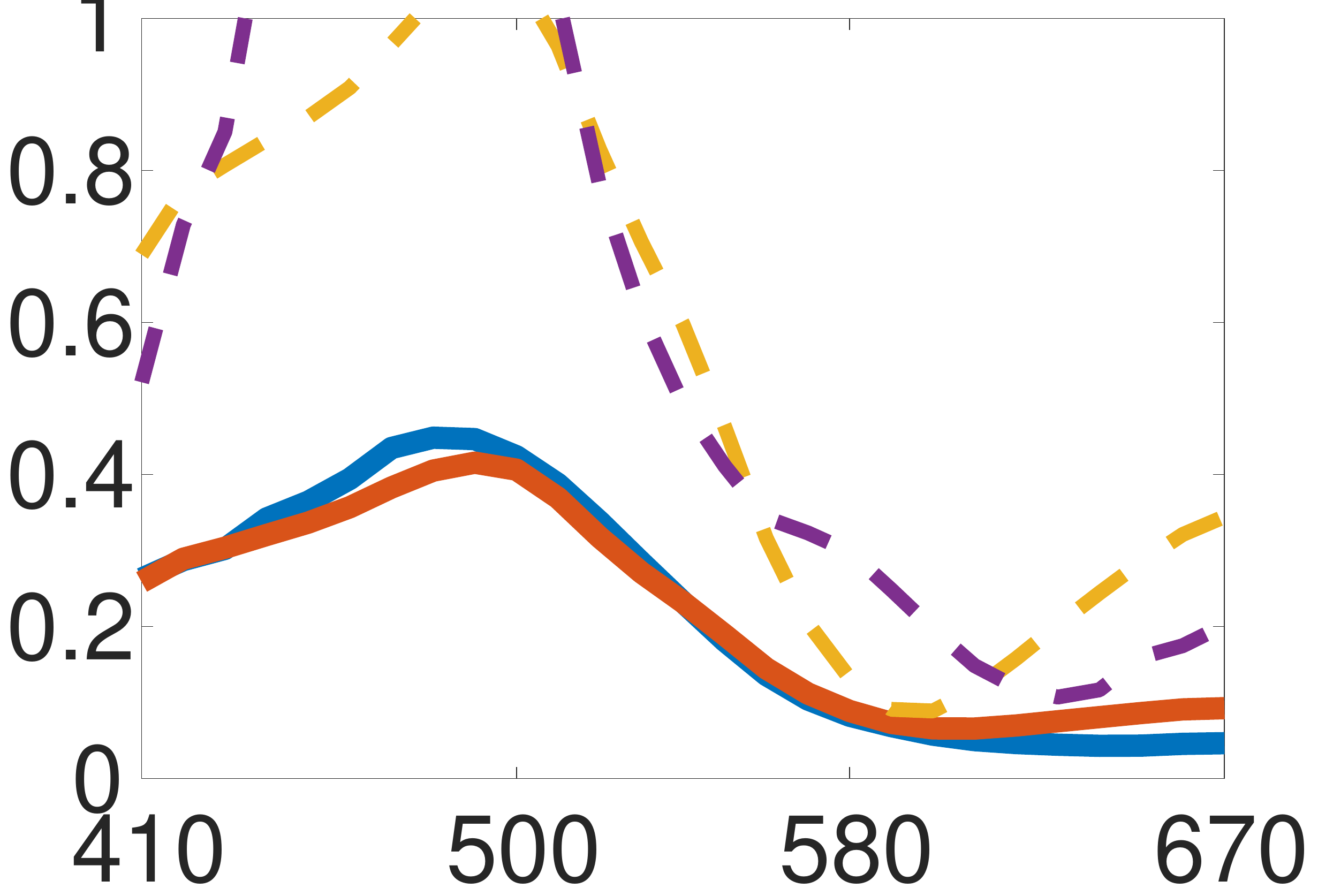}\\
    		\includegraphics[width=0.166\linewidth]{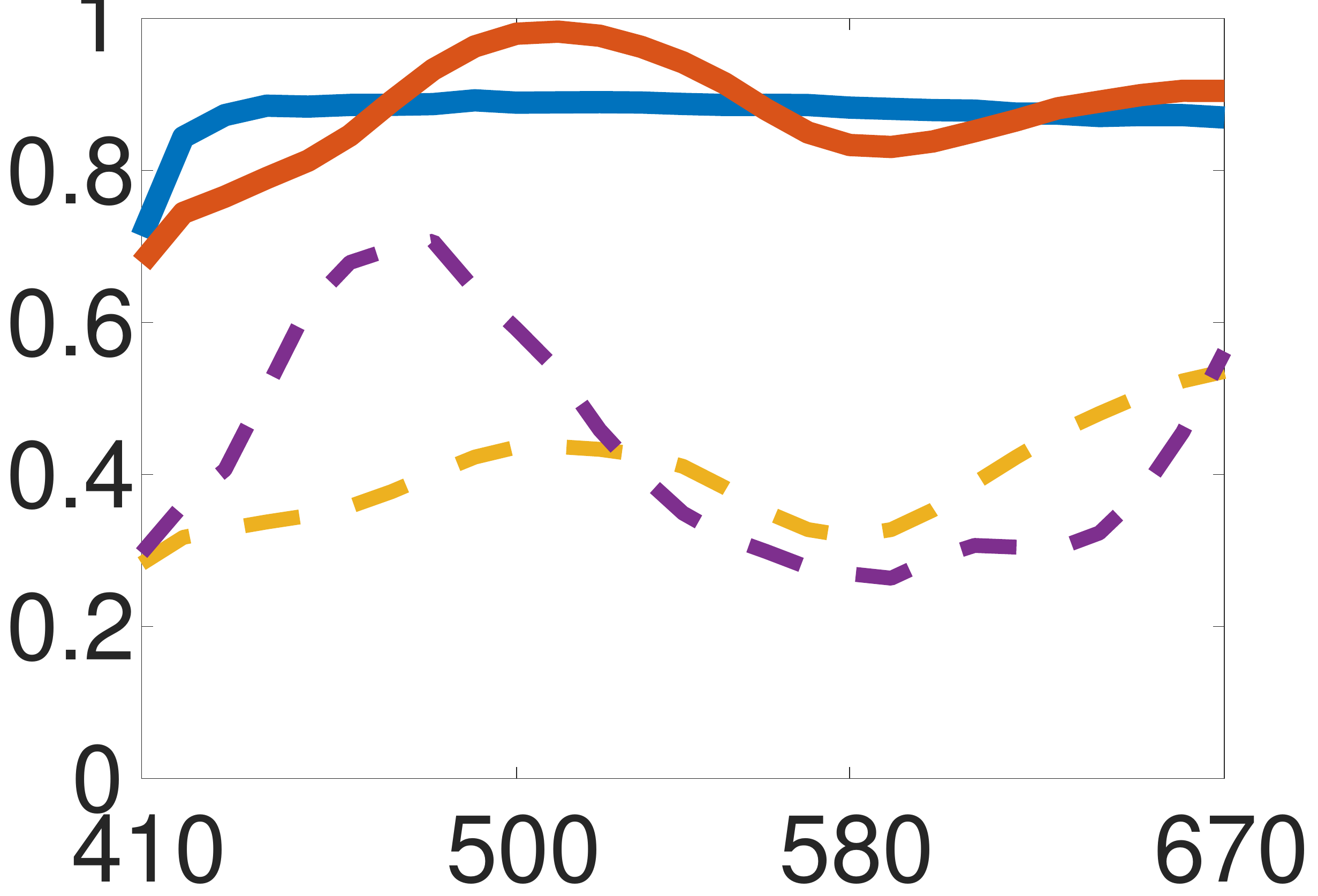} &
    	\includegraphics[width=0.166\linewidth]{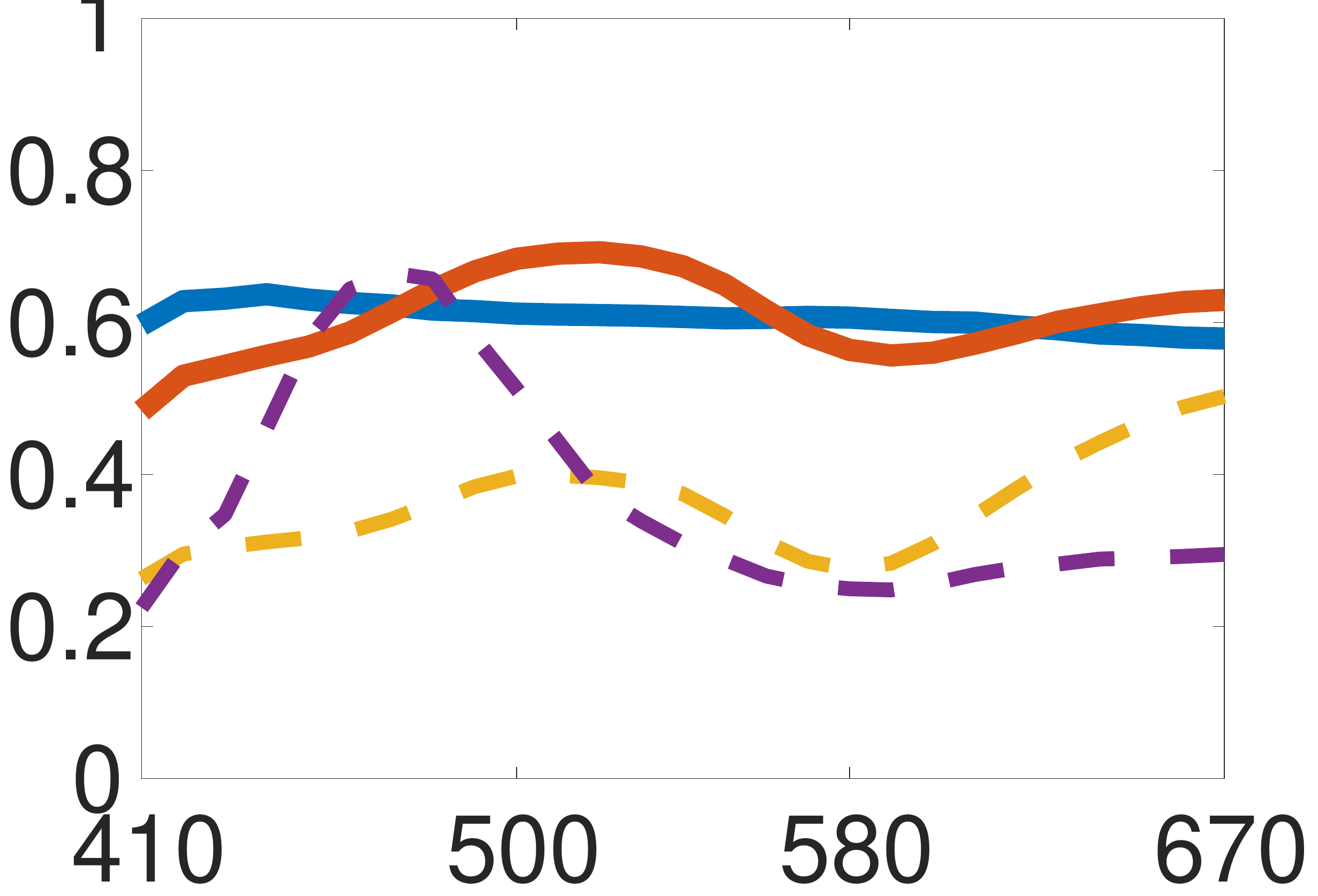} &
    	\includegraphics[width=0.166\linewidth]{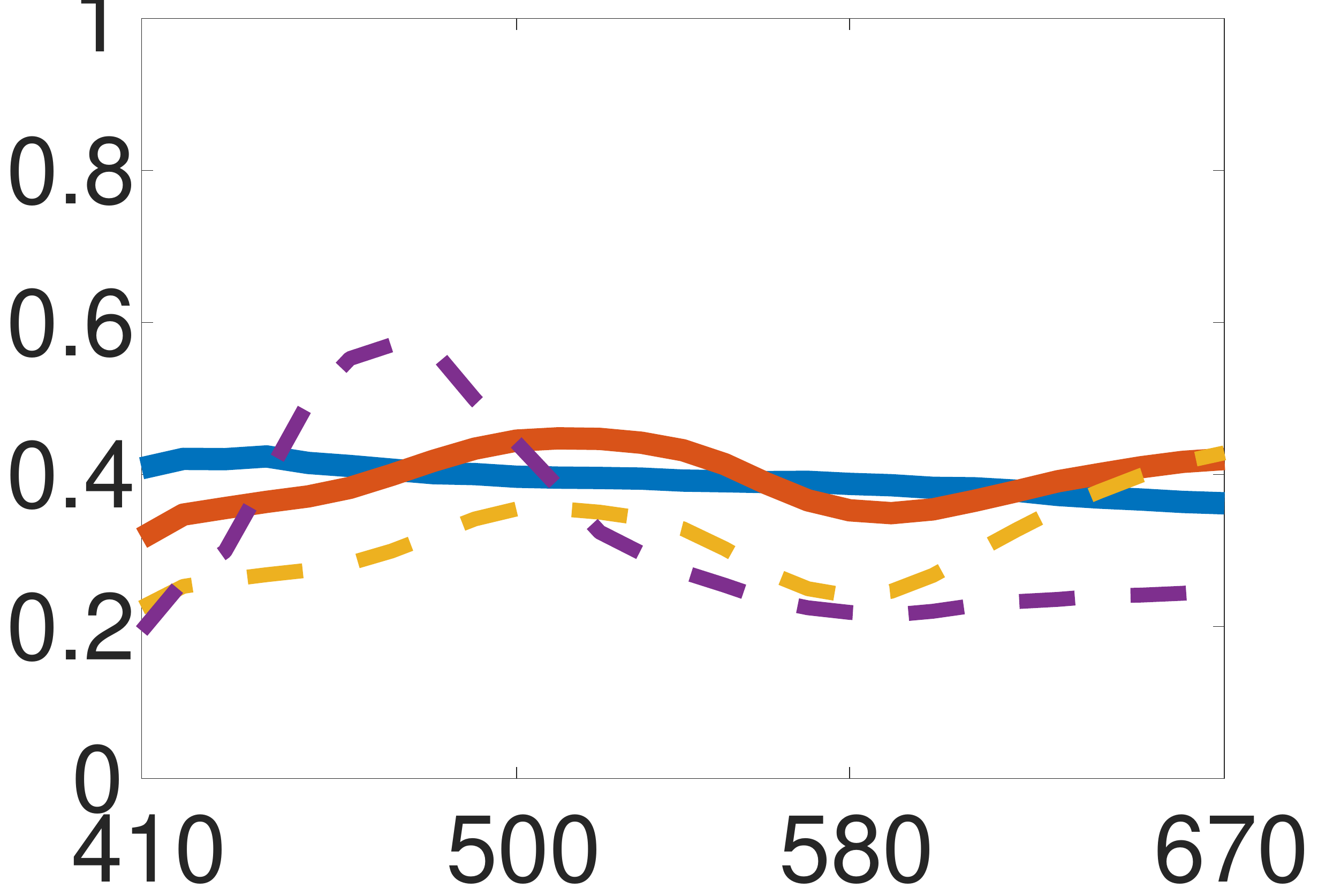} &
    	\includegraphics[width=0.166\linewidth]{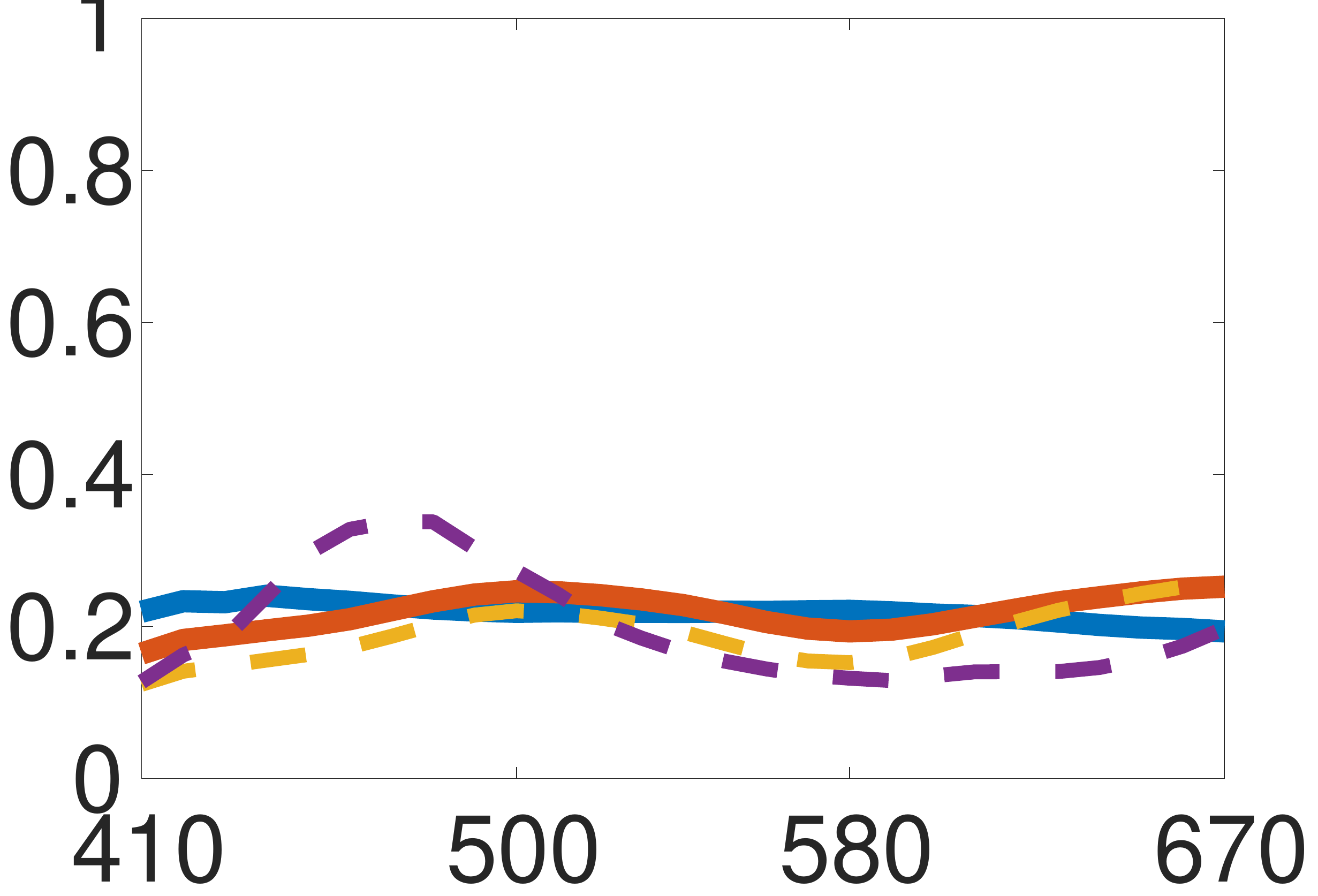} &
    	\includegraphics[width=0.166\linewidth]{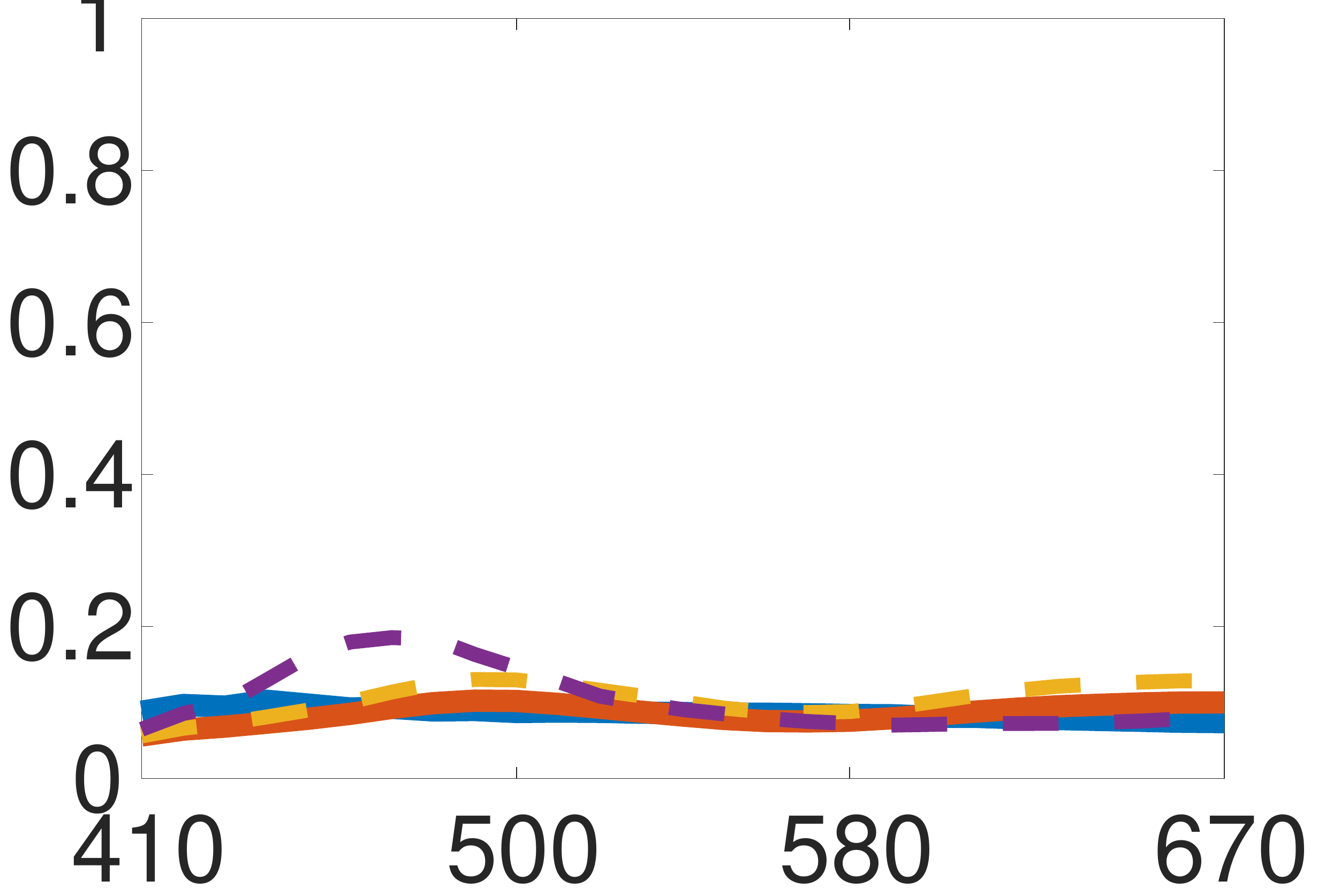} &
    	\includegraphics[width=0.166\linewidth]{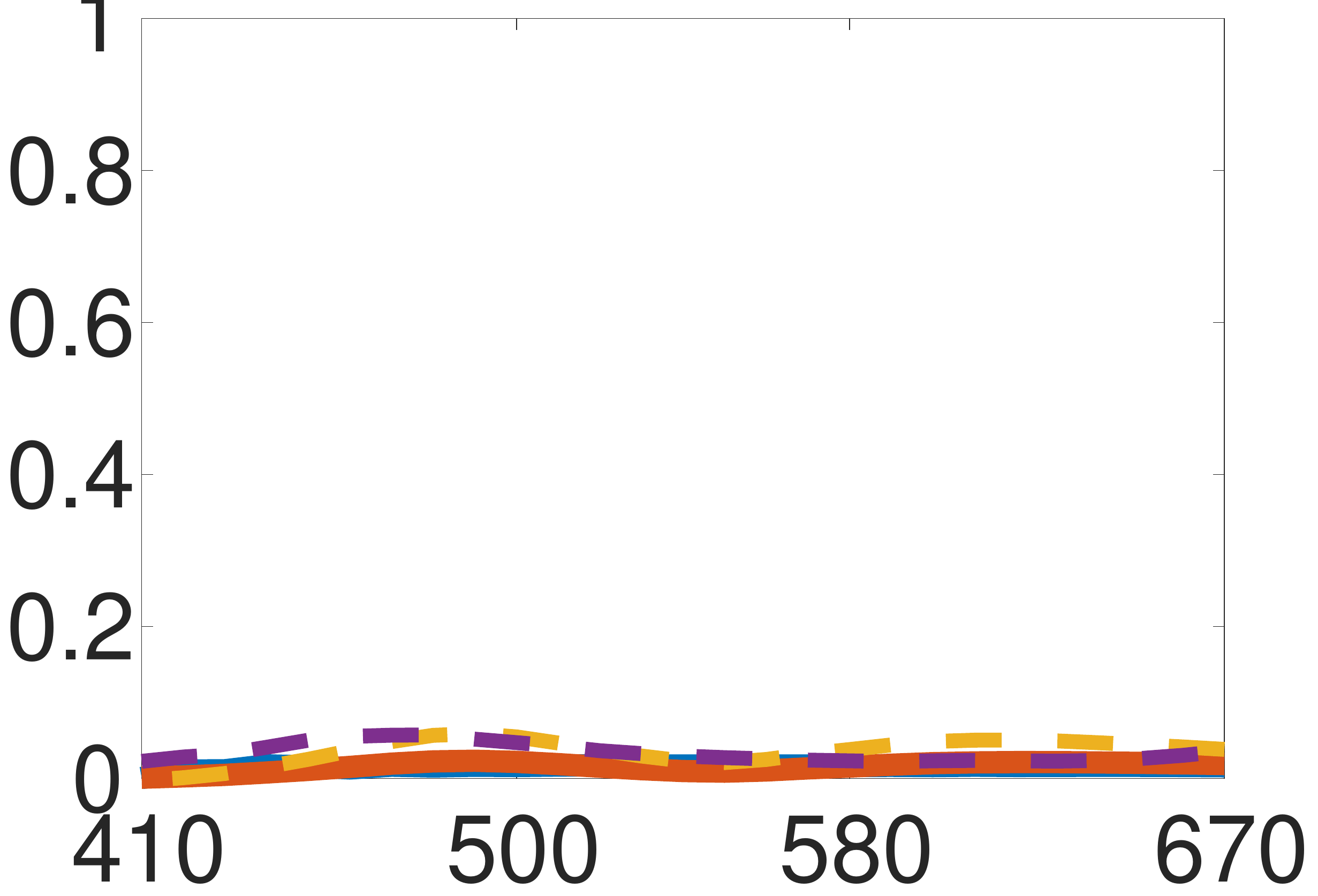} \vspace{1mm} \\
      \end{tabular}\\
      \small{(d) Comparison of the estiamted spectral reflectance results by our method (red line) and two existing single-view methods (yellow line~\cite{Han2} and purple line~\cite{Arad}) using the projector-camera pair 4}\\
      \end{minipage}
  \end{minipage}
  \vspace{2mm}\\
  \begin{minipage}{\hsize}
      \begin{minipage}{0.27\hsize}
        \centering
      \vspace{1.8mm}
          {\renewcommand{\arraystretch}{0.5}
          \begin{tabular}{@{\hskip 2pt}c@{\hskip 2pt}c@{\hskip 2pt}}
          \includegraphics[width=0.49\hsize]{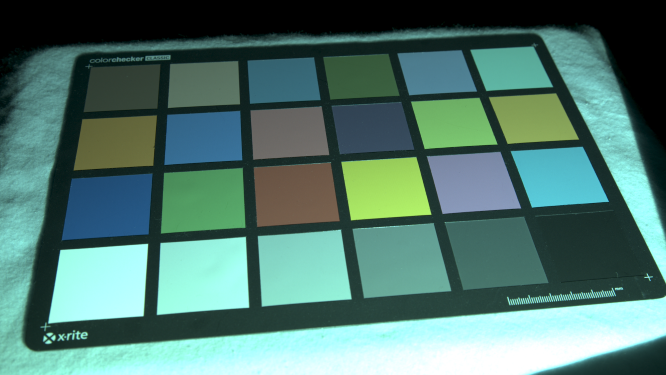}&
          \includegraphics[width=0.49\hsize]{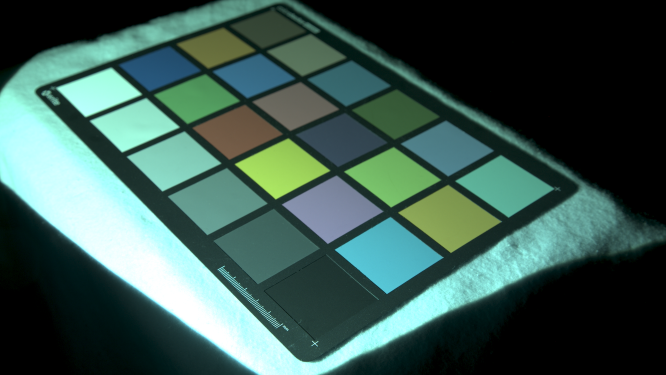}\\
            \footnotesize{Pair 1}&
          \footnotesize{Pair 2}\\
          \includegraphics[width=0.49\hsize]{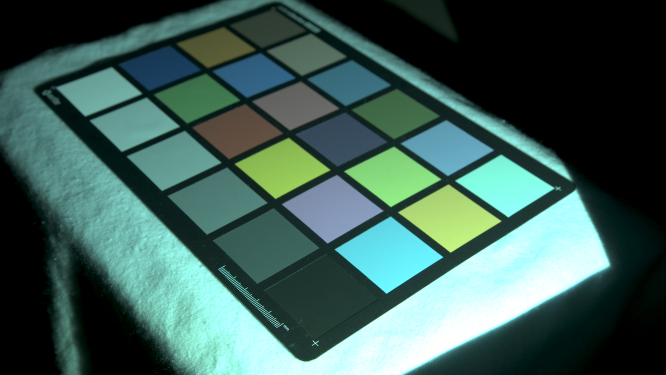}&
          \includegraphics[width=0.49\hsize]{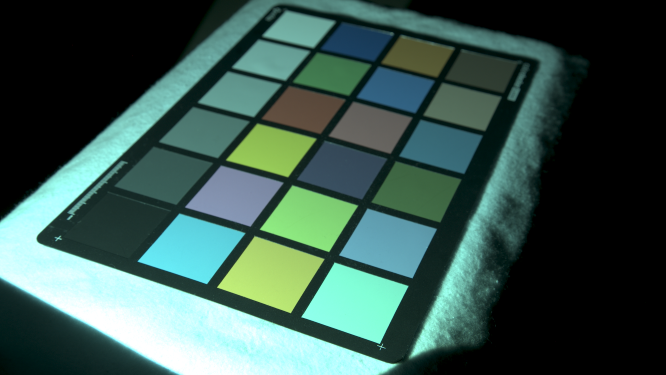}\\
          \footnotesize{Pair 3}&
          \footnotesize{Pair 4} \vspace{1mm}\\
          \end{tabular}
          }
          \vspace{-0.1mm}\\
          \small{(c) Example captured images}
      \end{minipage}
　　　\hspace{1mm}
      \begin{minipage}{0.72\hsize}
        \centering
        \includegraphics[width=\hsize]{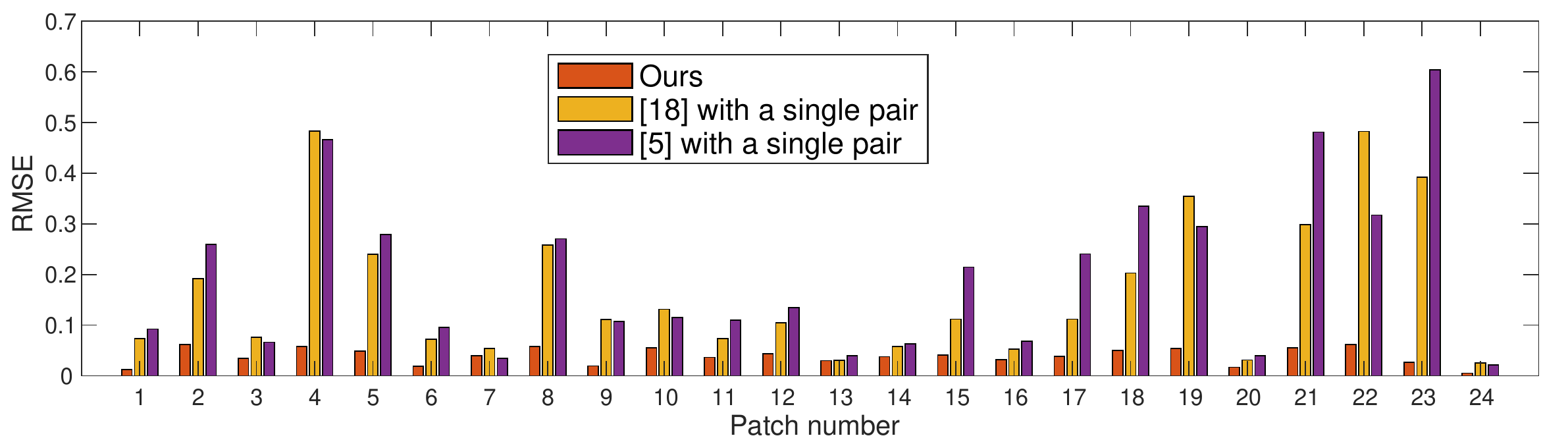}\\
        \small{(e) RMSE comparison for each patch of the colorchart}
      \end{minipage}
  \end{minipage}
  \vspace{1mm}
  \caption{Spectral reflectance estimation results on the 24 patches of the colorchart. As can be seen in (d) and (e), the existing single-view image-based methods~\cite{Han2,Arad} fail to correctly estimate the spectral reflectance of each patch (including relative scales between patches) due to the shading effect apparent in the colorchart setup, as shown in (c). In contrast, our method can accurately estimate the spectral reflectance including the relative scales by considering the geometric relationship between the 3D points and the projector positions.}
  \label{fig:colorchart}
\end{figure*}

\subsection{Spectral reflectance estimation results}

To evaluate the performance of our spectral reflectance estimation method, we used a standard colorchart with the 24 patches.
We first show the effect of the number of spectral bands. In our experiment, seven color illuminations and RGB camera channels, as shown in~Fig.~\ref{fig:spectrum}, were used, resulting in a total of 21-band measurements. To select the best band set, we evaluated all possible band sets for each number of spectral bands. Figure~\ref{fig:multibandselection} shows RMSE for the 24 patches of the colorchart when using the selected best band set for each number of spectral bands. We can observe that RMSE is reduced by using multispectral information and becomes very close when more than six bands are used. The six-band set of ($light$, $camera$) = ($L_{green}$, $C_{blue}$), ($L_{blue}$, $C_{green}$), ($L_{blue}$, $C_{blue}$), ($L_{cyan}$, $C_{red}$), ($L_{magenta}$, $C_{red}$), ($L_{yellow}$, $C_{green}$) provides the minimium RMSE among the evaluated all possible band sets.

We next demonstrate the effectiveness of our spectral reflectance estimation model considering the geometric information. As shown in~Fig.~\ref{fig:colorchart}(a), we laid the colorchart on a table and captured the structured light and multispectral data by four projector-camera pairs according to the data acquisition procedure of Pro-Cam SSfM. The estimated projector positions, camera positions, and 3D points of the colorchart are shown in Fig.~\ref{fig:colorchart}(b). The example captured images (under white illumination) by four projector-camera pairs are shown in~Fig.~\ref{fig:colorchart}(c). Figure~\ref{fig:colorchart}(d) compares the estimated spectral refletance results for the 24 patches, where the blue line is the ground truth, the red line is our result (average withing each patch) using all projector-camera pairs, and the yellow and purple dashed lines are the results of two existing single-view image-based methods~\cite{Han2,Arad} (average withing each patch) only using the projector-camera pair~4. Figure~\ref{fig:colorchart}(e) shows the corresponding RMSE comparison for each patch, where we can confirm that our method achieves much lower RMSE than the existing methods.

As can be seen in Fig.~\ref{fig:colorchart}(d) and~\ref{fig:colorchart}(e), the single-view methods fail to correctly estimate the spectral reflectance including the relative scales between the patches. This is due to the shading effect appeared in the colorchart setup, as shown in Fig.~\ref{fig:colorchart}(c). In contrast, our method can provide accurate estimation results with correct relative scales. The benefit of our method is to estimate the spectral reflectance while considering the shading effect, which is ignored in the single-view methods. With this essential difference, our method is especially beneficial when the shading exists in the scene. If the shading does not exist, the accuracy of our method could be similar to that of the existing methods. However, such no-shading condition is very special and possible only under fully controlled illumination.

\begin{figure}[!t]
  \centering
  \begin{minipage}{\hsize}
    \begin{minipage}{0.47\hsize}
        \centering
        \begin{minipage}{0.43\hsize}
            \centering
        	\includegraphics[width=\hsize]{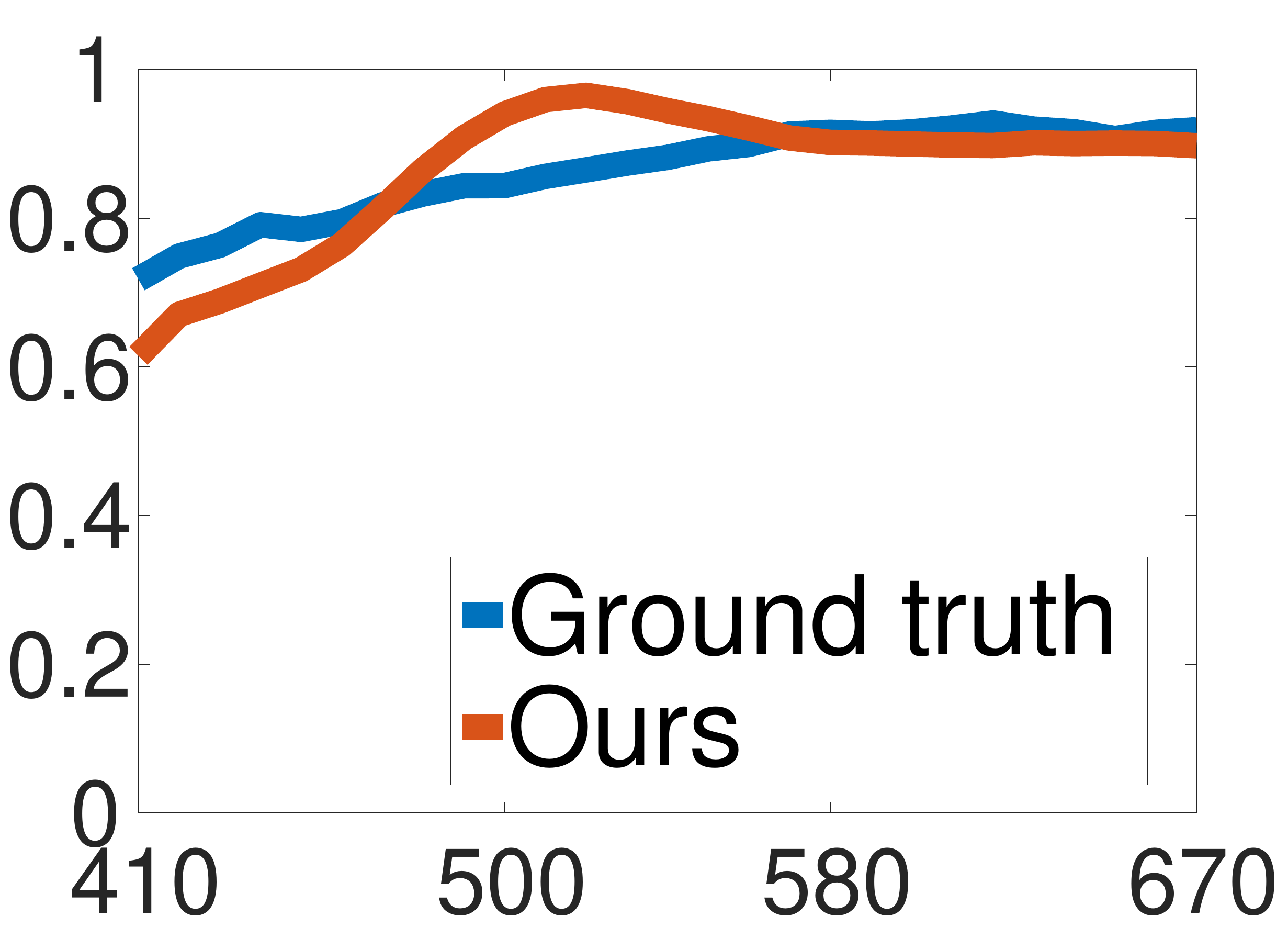}
        	\includegraphics[width=\hsize]{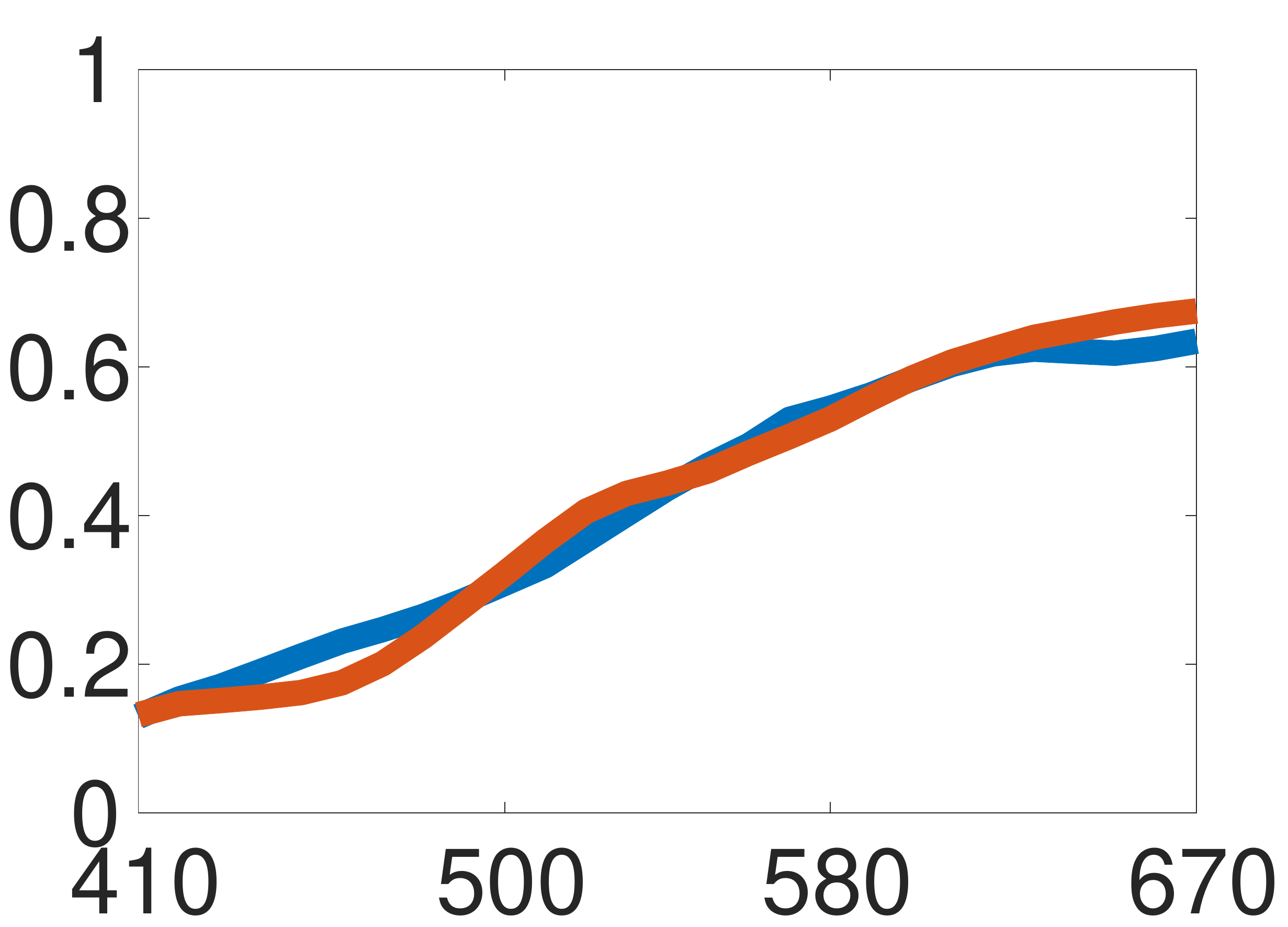}
        	\includegraphics[width=\hsize]{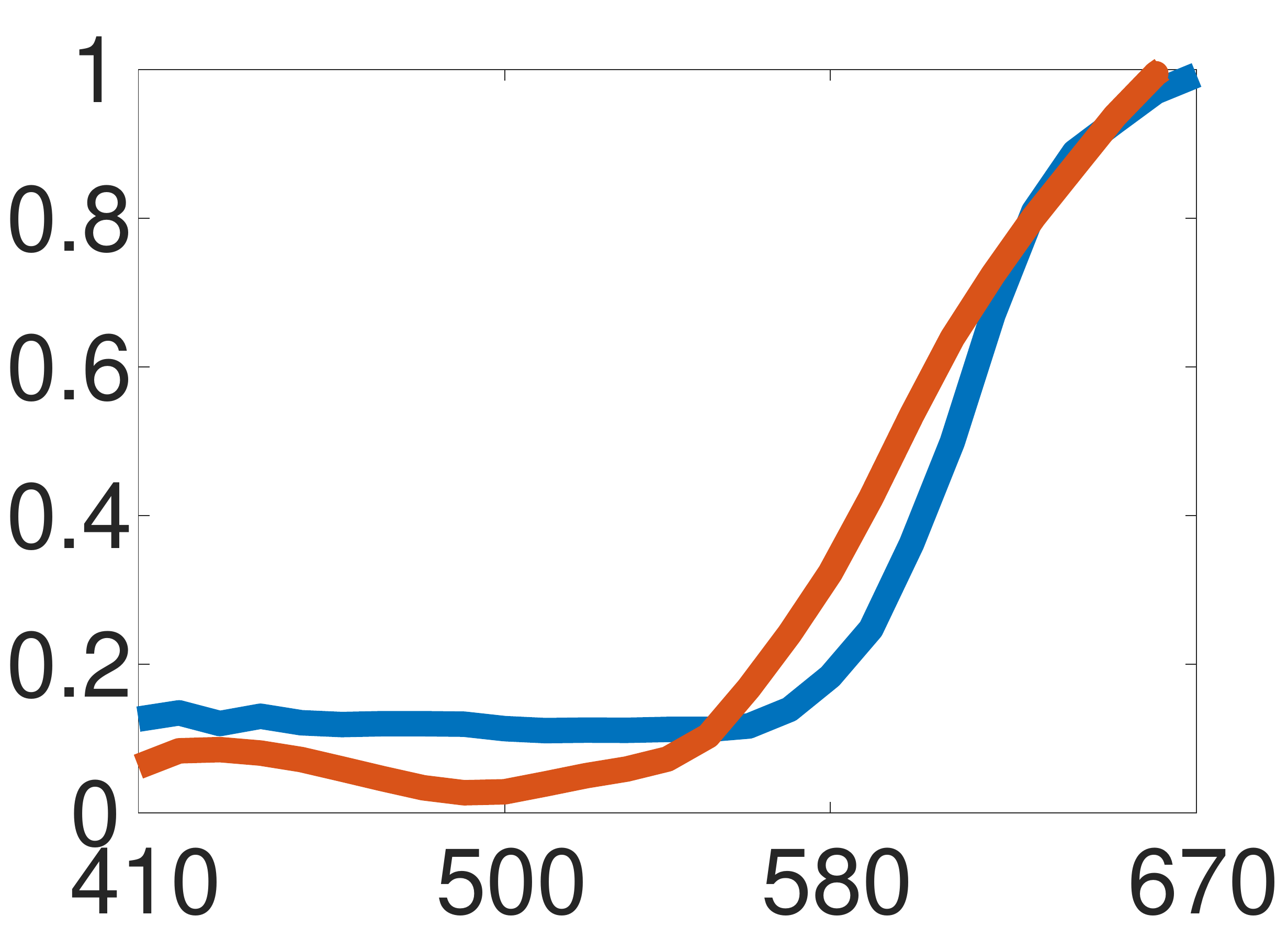}
    	\end{minipage}
    	\hspace{-1\fboxsep}
    	\begin{minipage}{0.54\hsize}
            \centering
    	    \includegraphics[width=\hsize]{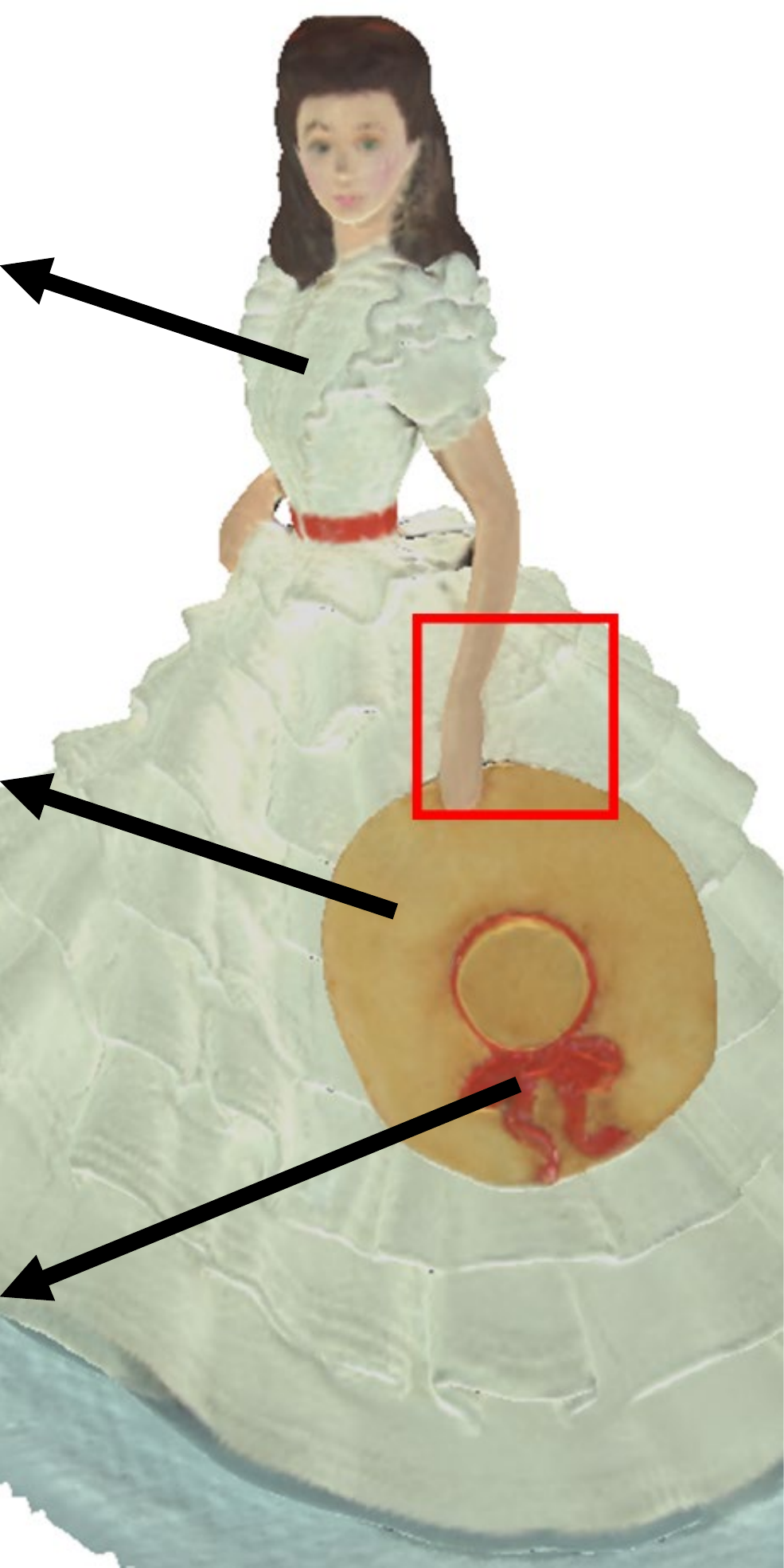}
        \end{minipage} \\
        \vspace{1mm}
	    \small{(a) Our result (sRGB)}
    \end{minipage}
    \begin{minipage}{0.255\hsize}
        \centering
	    \includegraphics[width=\hsize]{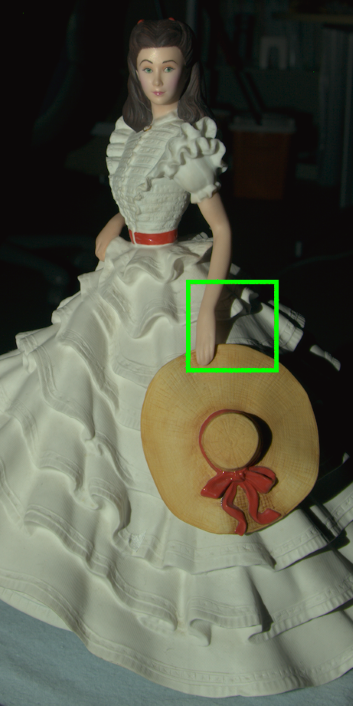}\\
	    \small{(b) Method~\cite{Han2}}
    \end{minipage}
    \begin{minipage}{0.255\hsize}
        \centering
	    \includegraphics[width=\hsize]{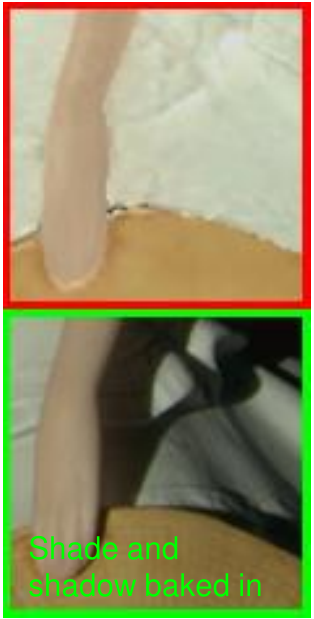}\\
	    \small{(c) Closeup}
    \end{minipage}
 \end{minipage}
 \begin{minipage}{\hsize}
    \centering
    \vspace{3mm}
    \begin{minipage}{0.66\hsize}
        \centering
        \begin{overpic}[scale=.108]{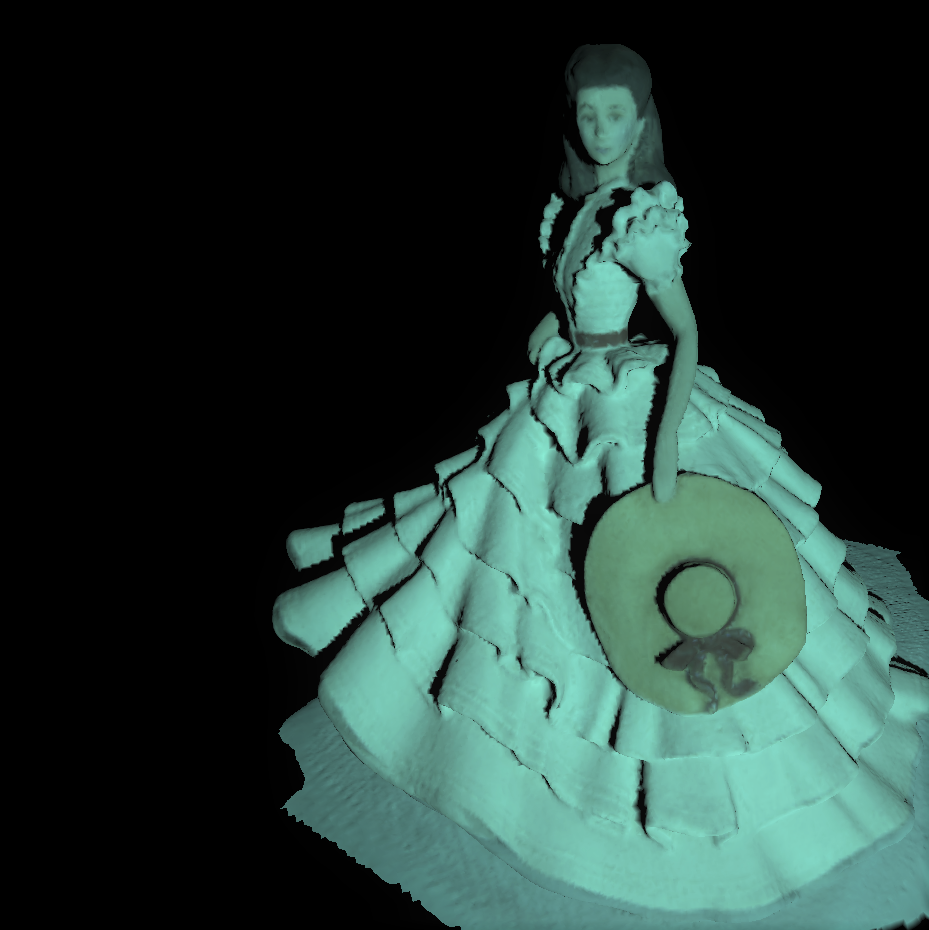}
            \put(0,55){\includegraphics[scale=.11]{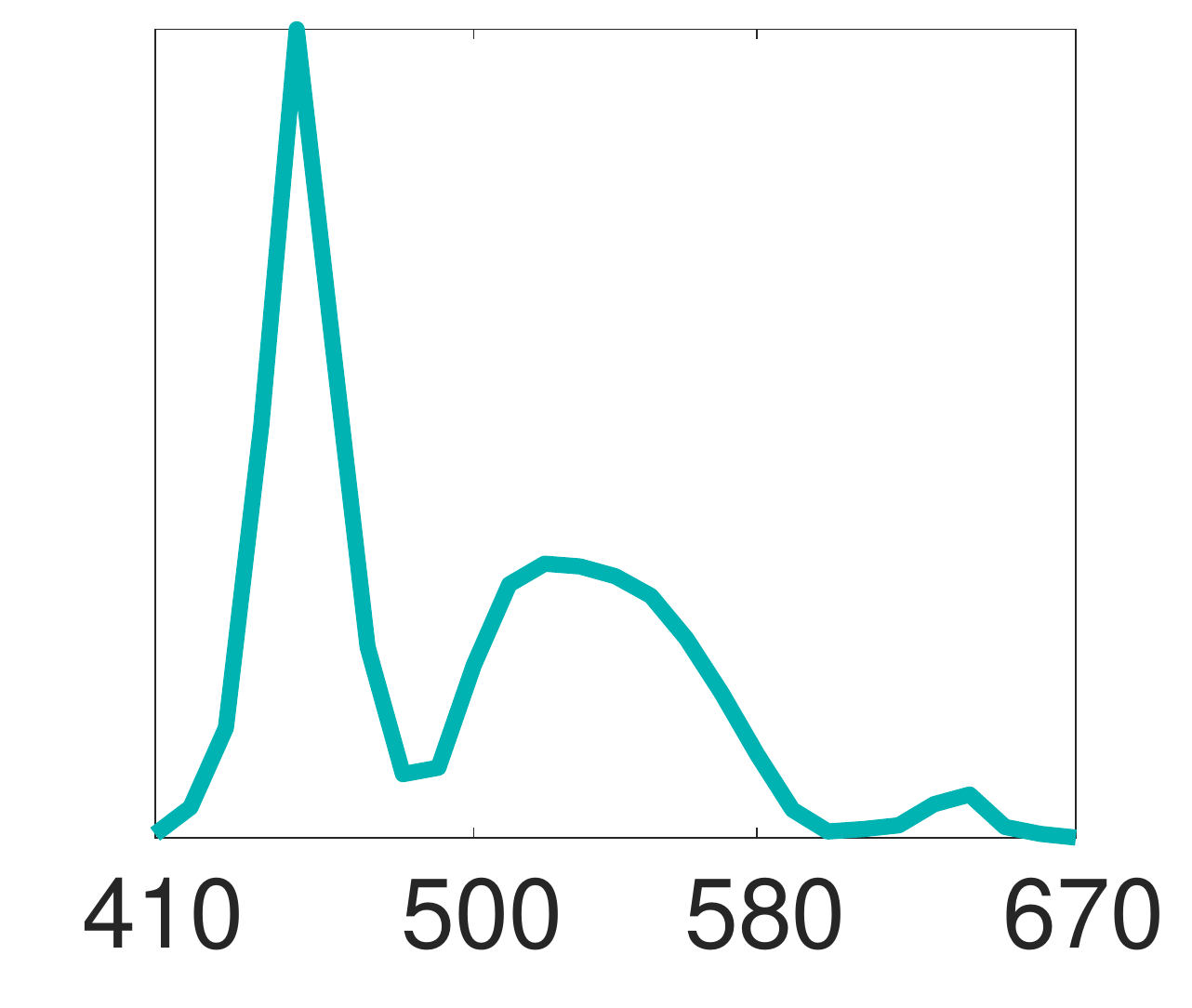}}
        \end{overpic}
        \begin{overpic}[scale=.107]{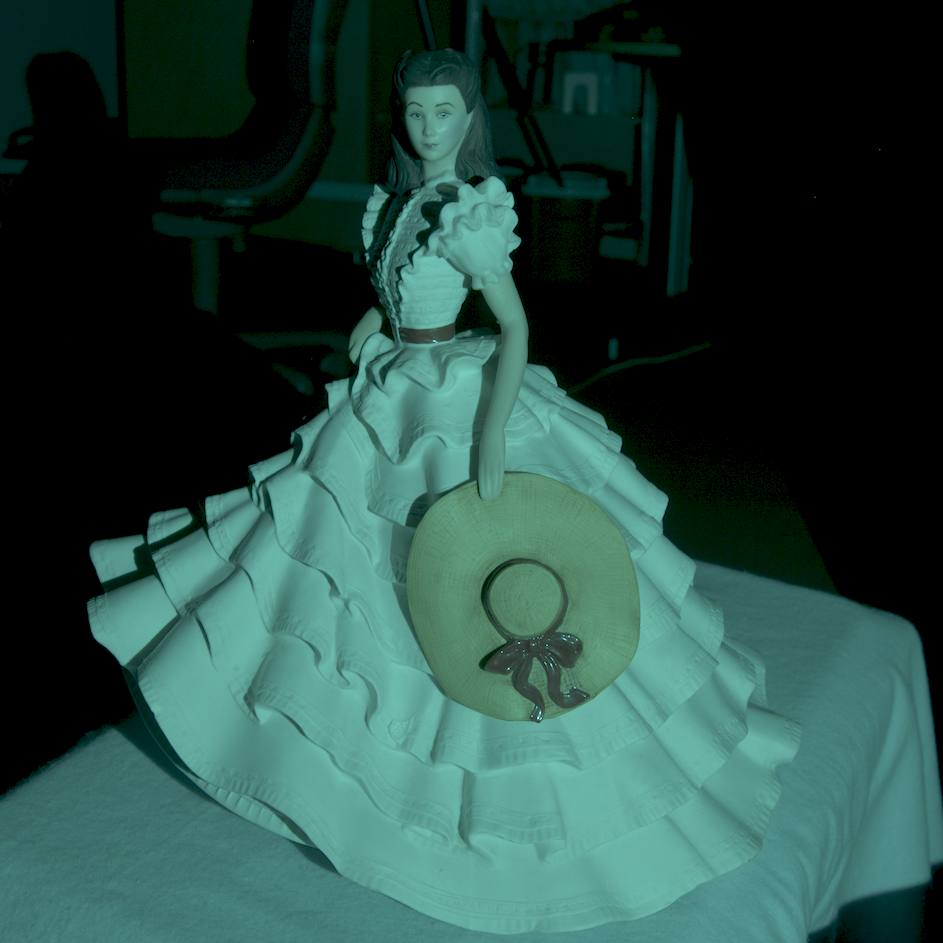}
        \end{overpic}\\
        \small{(d) Our 3D relighting result (left) and the reference actual image (right)}
    \end{minipage}
    \vspace{3mm}
    \begin{minipage}{0.33\hsize}
        \centering
        \begin{overpic}[scale=.12]{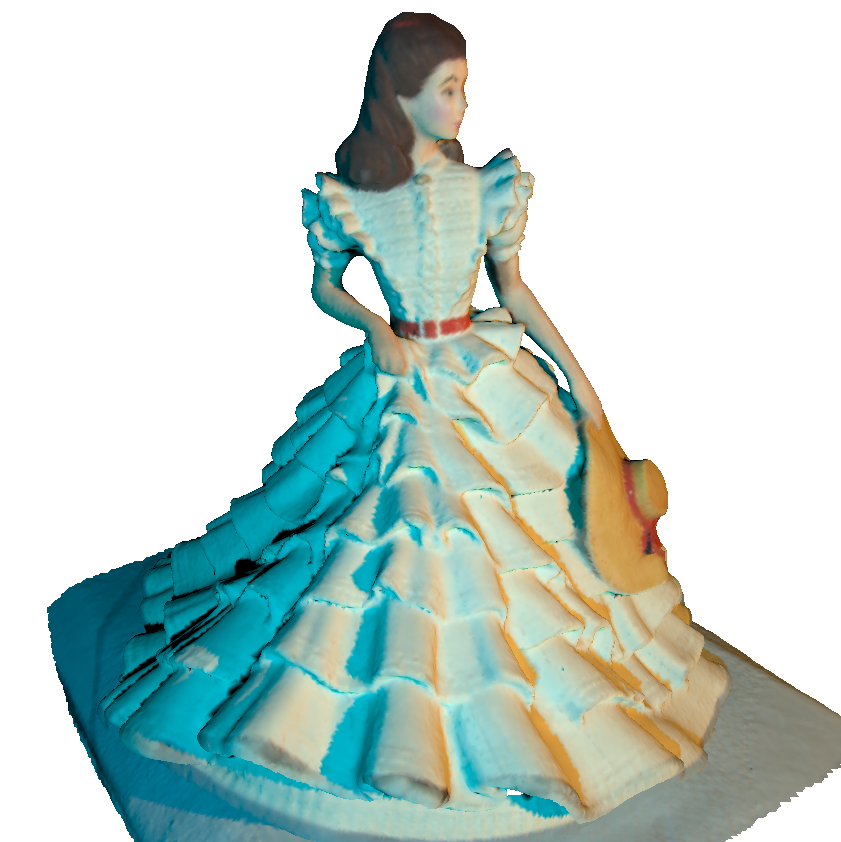}
            \put(-3,65){\includegraphics[scale=.08]{cyan.pdf}}
            \put(65,65){\includegraphics[scale=.08]{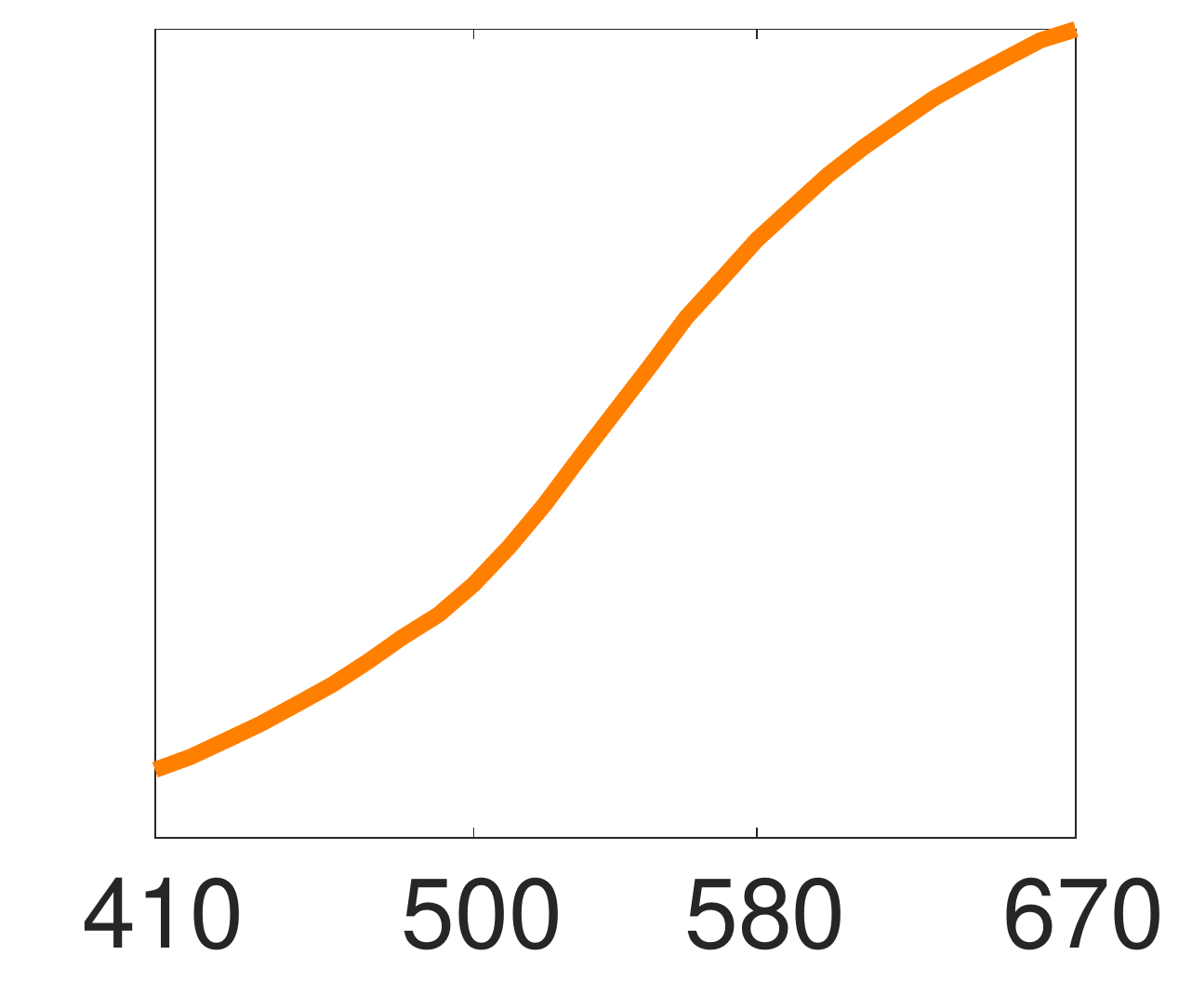}}
        \end{overpic}\\
        \small{(e) Relighting under two light sources}
    \end{minipage}
 \end{minipage}
 \begin{minipage}{\hsize}
        \centering
        \vspace{-3mm}
	    \includegraphics[width=0.32\hsize]{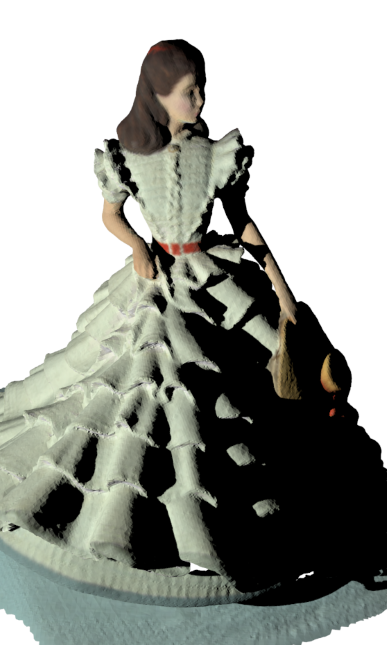}
	    \includegraphics[width=0.32\hsize]{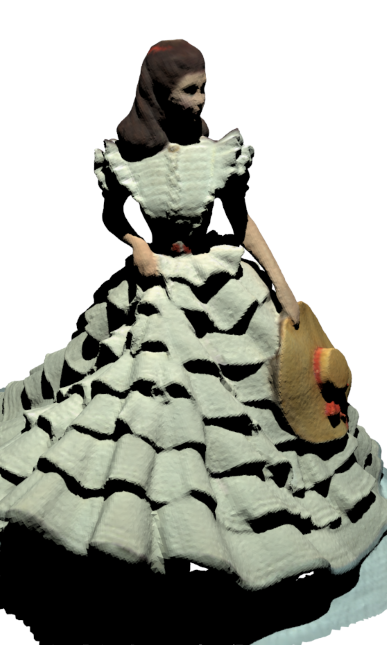}
	    \includegraphics[width=0.32\hsize]{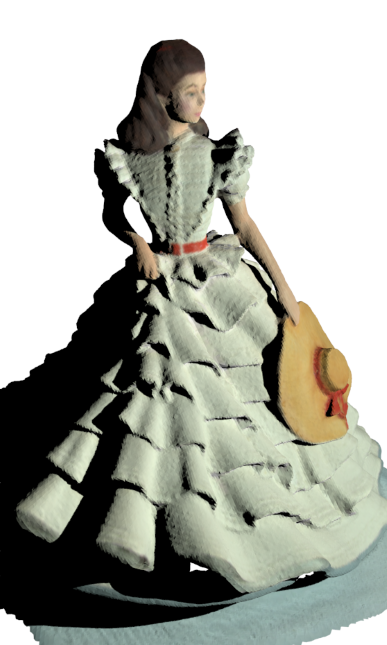}\\
	    \small{(f) 3D relighting results under different light orientations}
    \end{minipage}\\
    \vspace{2mm}
    \centering
  \caption{Results of the spectral 3D acquisition on a clay sculpture.}
  \label{fig:realdatascarlet}
  \vspace{-0.3em}
\end{figure}

\begin{figure}[!tbp]
  \centering
  \begin{minipage}{\hsize}
    \begin{minipage}{0.46\hsize}
        \centering
        \begin{minipage}{0.44\hsize}
            \centering
        	\includegraphics[width=\hsize]{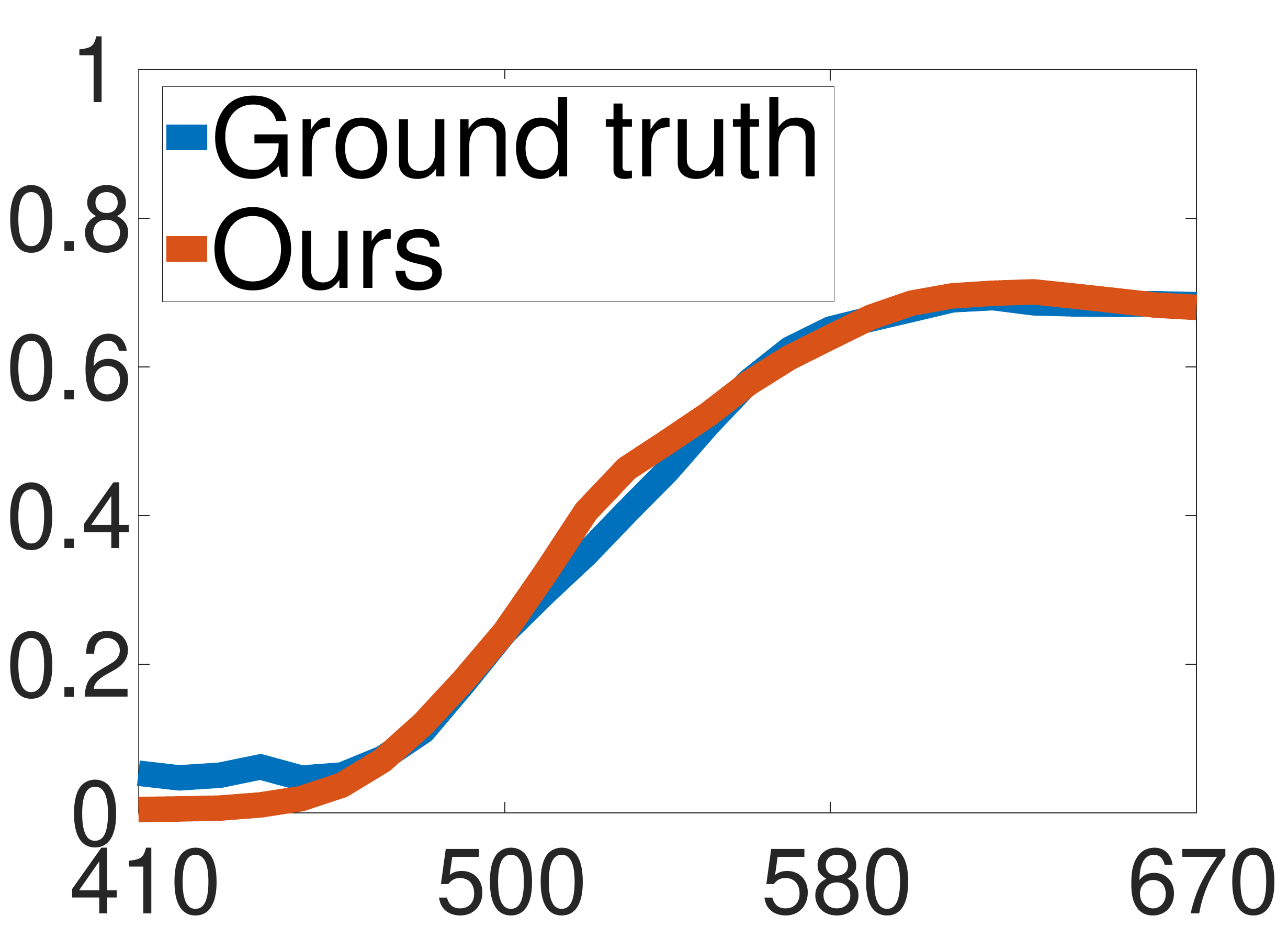}
        	\includegraphics[width=\hsize]{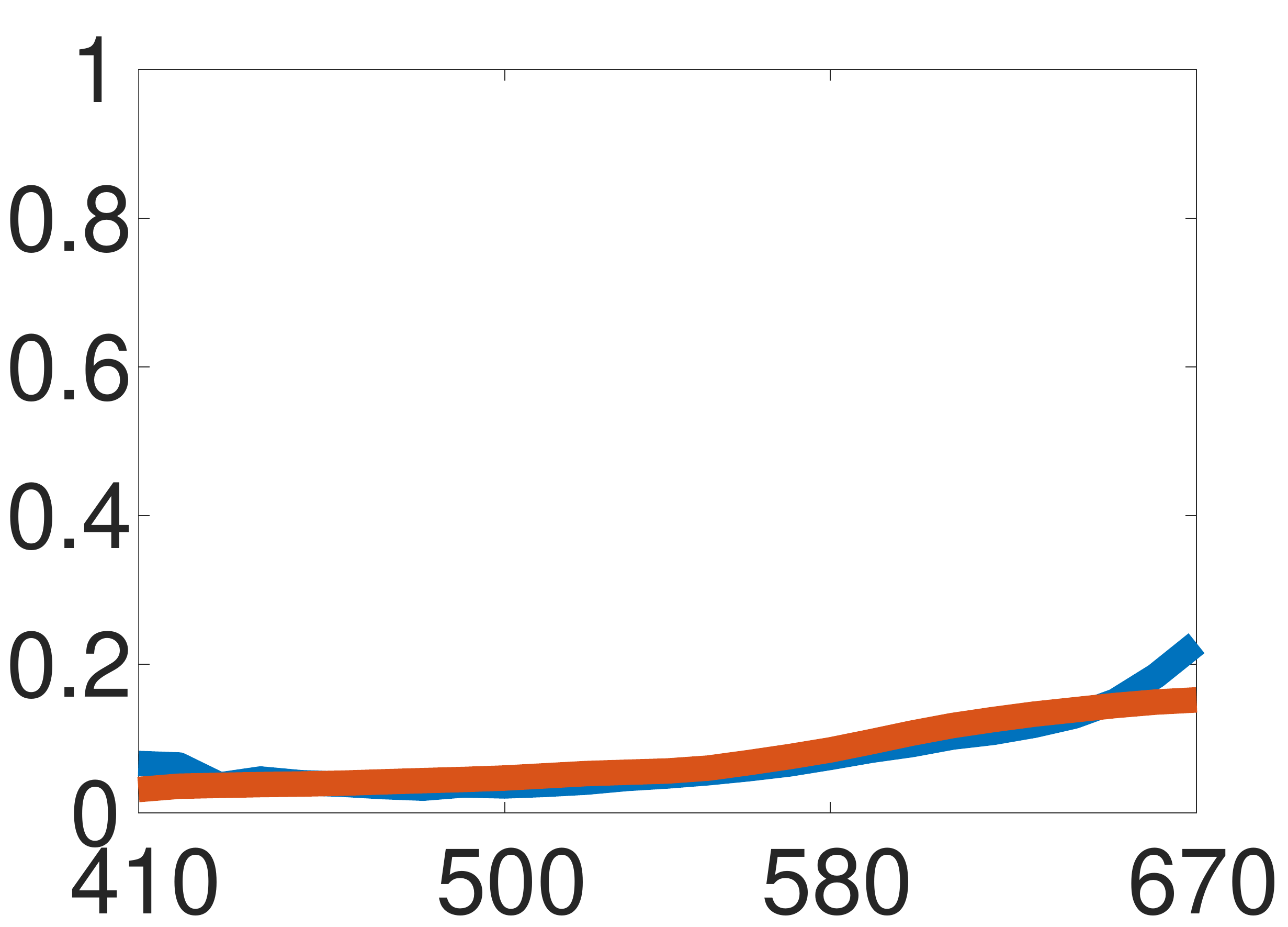}
        	\includegraphics[width=\hsize]{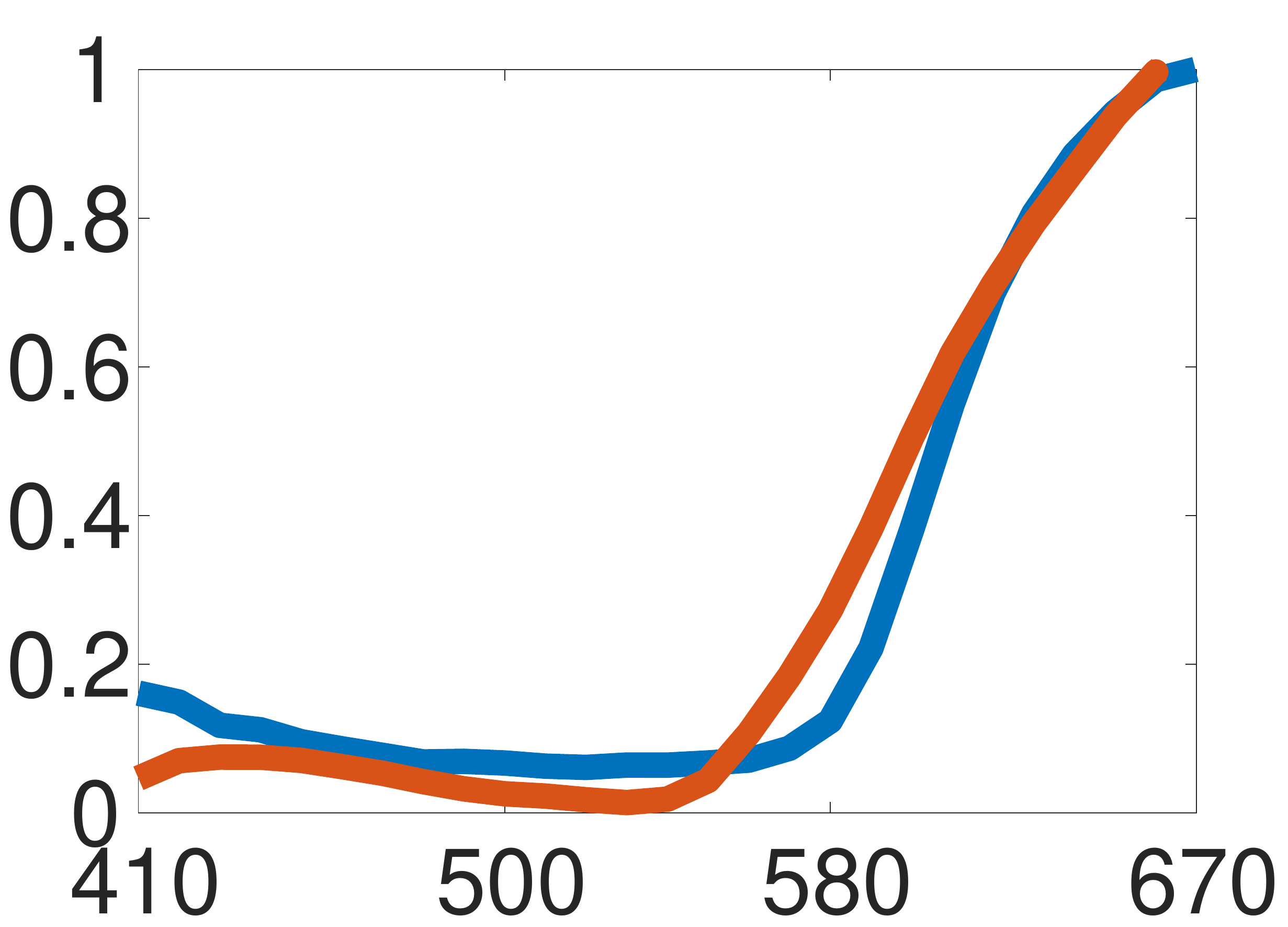}
    	\end{minipage}
    	\hspace{-1\fboxsep}
    	\begin{minipage}{0.54\hsize}
            \centering
    	    \includegraphics[width=\hsize]{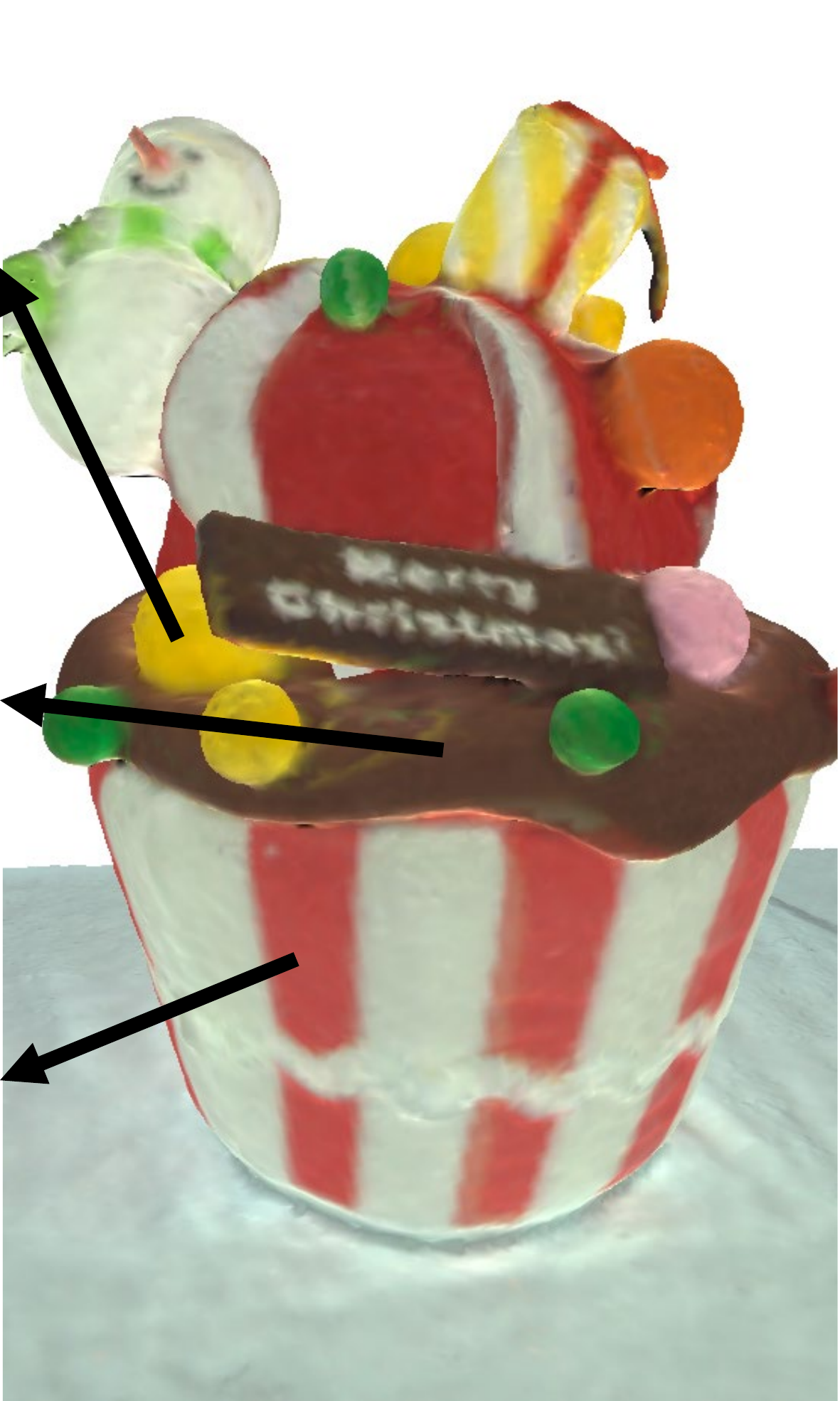}
        \end{minipage}\\
        \vspace{1mm}
	    \small{(a) sRGB and reflectance} \vspace{3mm}
    \end{minipage}
    \hspace{-2\fboxsep}
    \begin{minipage}{0.53\hsize}
        \centering
	    \includegraphics[width=0.48\hsize]{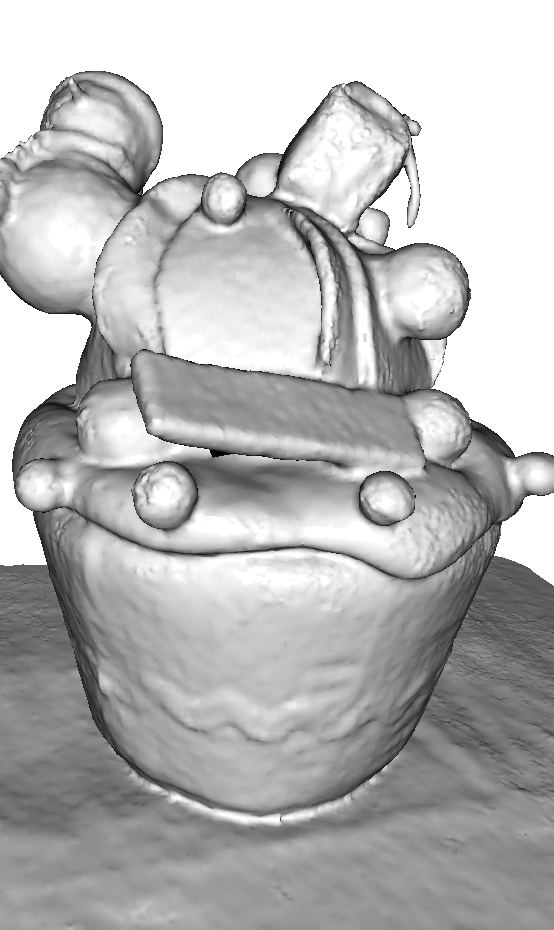}
	    \includegraphics[width=0.48\hsize]{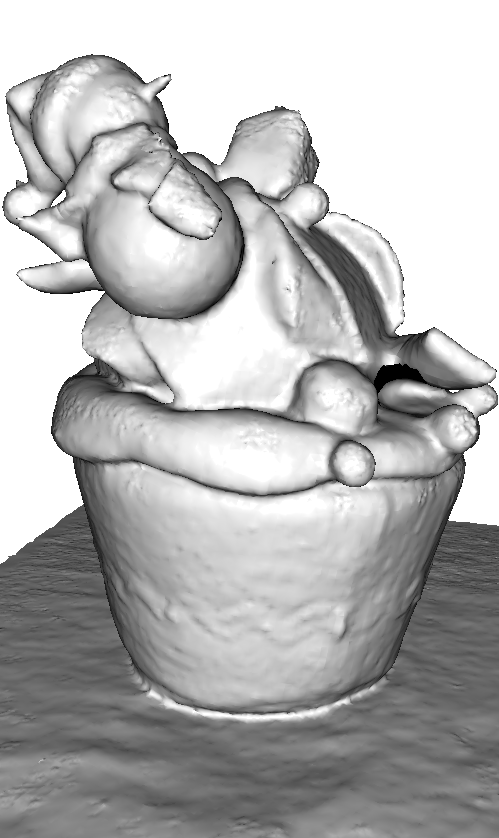}\\
	    \vspace{1mm}
	    \small{(b) 3D shape} \vspace{3mm}
    \end{minipage}
\end{minipage}
 \begin{minipage}{\hsize}
        \centering
          \begin{tabular}{@{\hskip 0pt}c@{\hskip 0pt}c@{\hskip 0pt}c@{\hskip 0pt}c@{\hskip 0pt}c@{\hskip 0pt}}
    \includegraphics[width=0.20\linewidth]{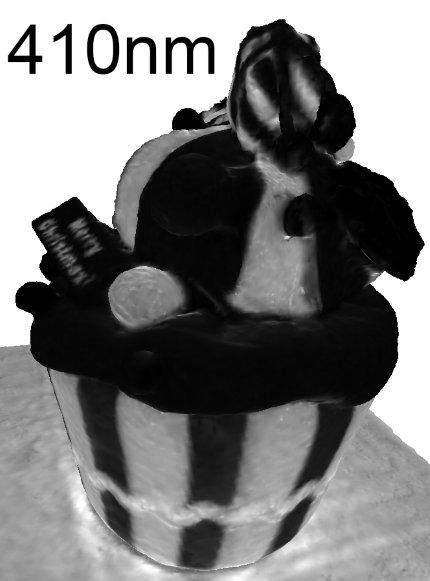} &
	\includegraphics[width=0.20\linewidth]{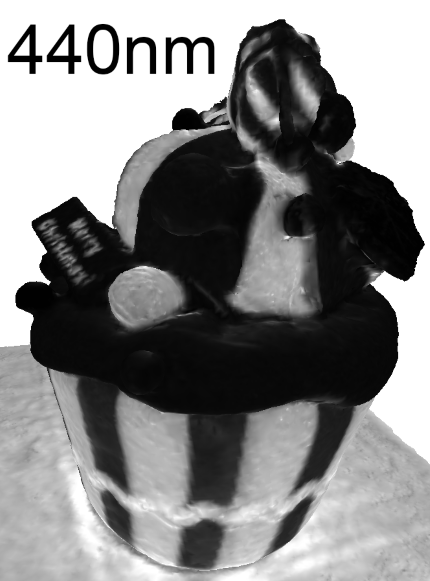} &
	\includegraphics[width=0.20\linewidth]{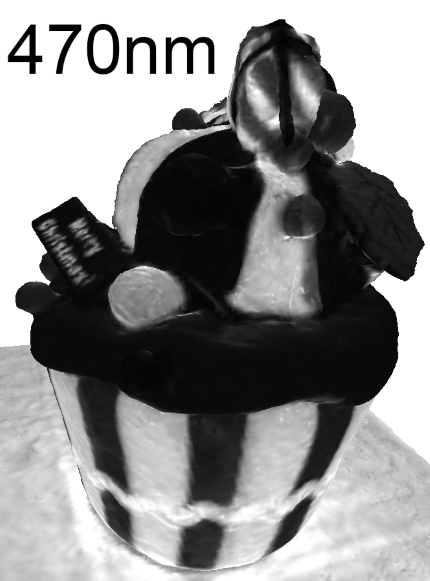} &
	\includegraphics[width=0.20\linewidth]{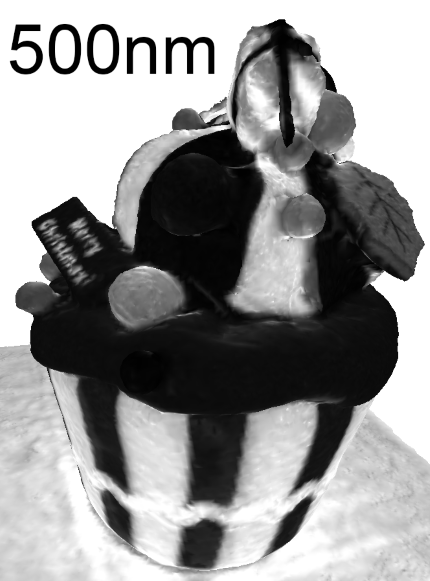} &
	\includegraphics[width=0.20\linewidth]{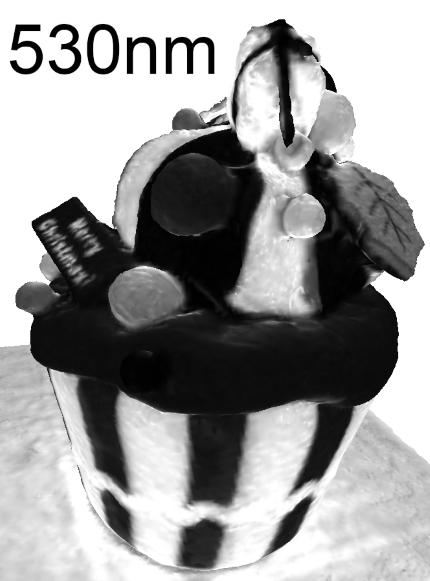} \\
	\includegraphics[width=0.20\linewidth]{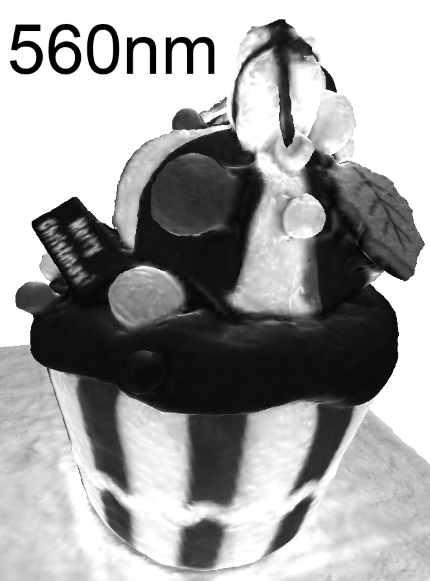} &
	\includegraphics[width=0.20\linewidth]{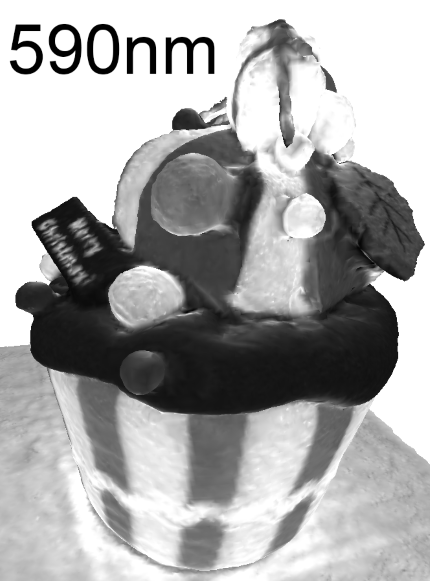} &
	\includegraphics[width=0.20\linewidth]{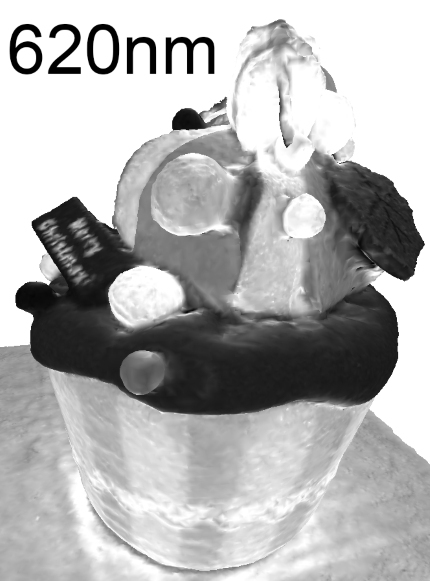} &
	\includegraphics[width=0.20\linewidth]{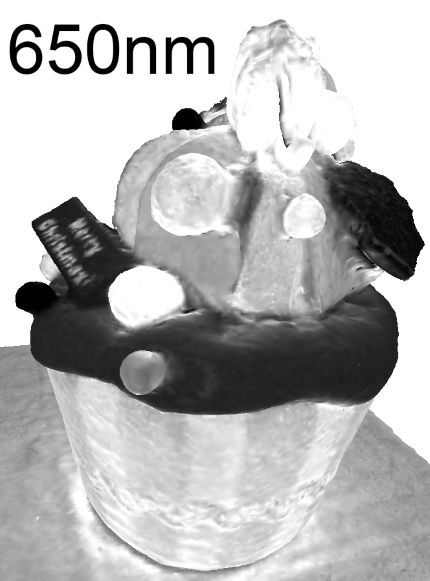} &
	\includegraphics[width=0.20\linewidth]{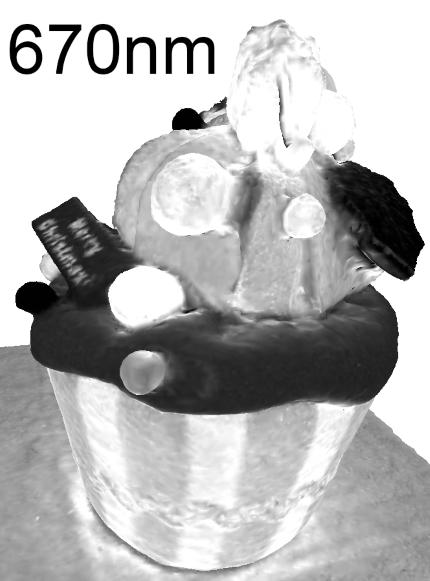} \\
  \end{tabular}\\
  \small{(c) Spectral patterns at each wavelength}
 \end{minipage}
  \vspace{1mm}
  \caption{Results of the spectral 3D acquisition on a stuffed toy.}
  \label{fig:realdatachristmas}
\end{figure}

\subsection{Spectral 3D acquisition results}

Figure~\ref{fig:realdatascarlet} shows the spectral 3D acquisition result on the clay sculpture. Figure~\ref{fig:realdatascarlet}(a) shows the spectral reflectance results for some 3D points. It is demonstrated that our method can accurately estimate the spectral reflectance compared with the ground truth measured by a spectrometer. Figure~\ref{fig:realdatascarlet}(a)--(c) compare the sRGB results converted from the estimated spectral reflectances by our method and the single-view method~\cite{Han2}. We can observe that our spectral reflectance estimation model considering the geometric information can effectively remove the baked-in effect of the shading and the shadow, which is apparent in the sRGB result of the single-view method. 

Since Pro-Cam SSfM can accurately estimate both the 3D points and the spectral reflectance, it is possible to perform the spectral 3D relighting of the object for synthesizing the appearance illuminated under an arbitrary light orientation and spectral distribution. Figure~\ref{fig:realdatascarlet}(d) shows the result of spectral relighting under the projector-cyan illumination, where we can confirm that our relighting result is close to the reference actual image taken in the same illumination orientation and spectral distribution. The differences at the concave skirt regions are due to the effect of interreflections, which is not considered in our current model.  
Figure~\ref{fig:realdatascarlet}(e) shows the more complex relighting result under two mixed light sources (projector cyan and halogen lamp), which are located at different sides of the object. Figure~\ref{fig:realdatascarlet}(f) shows the 3D relighting results under different illumination orientations. As shown in those results, we can effectively perform the spectral 3D relighting based on the estimated 3D model and spectral refletance. 

Figure~\ref{fig:realdatachristmas} shows the spectral 3D acquisition result on a stuffed toy. Figure~\ref{fig:realdatachristmas}(a)--(c) respectively show the spectral reflectance and the sRGB results, the 3D shape results, and the spectral patterns at each wavelength.
The presented results demonstrate the potential of Pro-Cam SSfM for accurate spectral 3D scanning and rendering. Additional results can be seen in the supplemental video.

\section{Concluding Remarks}

In this paper, we have proposed Pro-Cam SSfM, the fist spectral 3D acquisition system using an off-the-shelf projector and camera. By effectively exploiting the projector as active lighting for both the geometric and the photometric observations, Pro-Cam SSfM can accurately reconstruct a dense object 3D model with the spectral reflectance property. We have validated that our proposed spectral reflectance estimation model can effectively eliminate the shading effect by incorporating the geometric relationship between the 3D points and the projector positions into the cost optimization. We have experimentally demonstrated the potential of Pro-Cam SSfM through the spectral 3D acquisition results on several real objects.

Pro-Cam SSfM has several limitations. First, we currently assume that the illumination spectrum is known. Second, our spectral reflectance estimation model currently ignores interreflections. Possible future research directions are to address each limitation by simultaneously estimating the spectral reflectance and the illumination spectrum~\cite{Oh} or separating direct and global components by using projected high frequency illumination~\cite{Nayar}.

\vspace{3mm}
\noindent {\bf Acknowledgment} This work was partly supported by JSPS KAKENHI Grant Number 17H00744.

\clearpage

{\small
\bibliographystyle{ieee_fullname}
\bibliography{egbib}
}

\end{document}